The American University in Cairo

School of Sciences and Engineering

# Measuring Atmospheric Scattering from Digital Images of Urban Scenery using Temporal Polarization-Based Vision

A Thesis Submitted to

Computer Science Department

The American University in Cairo

in partial fulfillment of the requirements for

the degree of Masters of Science

BY

Tarek El-Gaaly

B. Sc. In Computer Science, 2005

The American University in Cairo

under the supervision of

Dr. Joshua Gluckman

2009

# Abstract

Suspended atmospheric particles (particulate matter) are a form of air pollution that visually degrades urban scenery and is hazardous to human health and the environment. Current environmental monitoring devices are limited in their capability of measuring average particulate matter (PM) over large areas. Quantifying the visual effects of haze in digital images of urban scenery and correlating these effects to PM levels is a vital step in more practically monitoring our environment. Current image haze extraction algorithms remove all the haze from the scene and hence produce unnatural scenes for the sole purpose of enhancing vision. We present two algorithms which bridge the gap between image haze extraction and environmental monitoring. We provide a means of measuring atmospheric scattering from images of urban scenery by incorporating temporal knowledge. In doing so, we also present a method of recovering an accurate depthmap of the scene and recovering the scene without the visual effects of haze. We compare our algorithm to three known haze removal methods from the perspective of measuring atmospheric scattering, measuring depth and dehazing. The algorithms are composed of an optimization over a model of haze formation in images and an optimization using the constraint of constant depth over a sequence of images taken over time. These algorithms not only measure atmospheric scattering, but also recover a more accurate depthmap and dehazed image. The measurements of atmospheric scattering this research produces, can be directly correlated to PM levels and therefore pave the way to monitoring the health of the environment by visual means. Accurate atmospheric sensing from digital images is a challenging and under-researched problem. This work provides an important step towards a more practical and accurate visual means of measuring PM from digital images.

# Table of Contents

# List of Figures

# Chapter 1 – Introduction

# 1.1 Problem Statement

How air pollution affects the visual appearance of scenes has been a research challenge for scientists. Changes in emissions, urban growth, and many other factors influence the amount and type of pollution suspended in the atmosphere. Suspended particles in the atmosphere are known as particulate matter (PM). PM visually degrades urban scenery and is hazardous to human health and the environment. Current measuring devices calculate the concentration of PM in the local vicinity of the device. These approaches do not measure the overall global concentration of PM over large urban areas and therefore lack accuracy in their measurements. Multiple measurements over a wide area are required to calculate the average PM level. Other means of measuring PM use laser devices and analyze satellite imagery. These methods are expensive, require skilled professionals to operate and are not suited to urban populated areas. Understanding and quantifying the visual effects of the atmosphere over large urban areas captured in digital images is a vital step in measuring global levels of PM more practically and accurately.

Understanding air pollution lies in the challenging process of measuring atmospheric scattering. Atmospheric scattering fluctuates proportionally to the levels of PM in the atmosphere and therefore offer a means of measuring PM. The challenge of measuring atmospheric scattering visually lies in finding the relationship between image pixel values and the atmospheric scattering taking place in the urban scene. The difficulty of finding this relationship lies in accounting for all the many factors that affect the pixel values. There are multiple factors which include: background molecular scattering of light, the position of the sun, the calibration of the camera's radiometric response curve, camera viewpoint, multiple-scattering, presence of non-spherical particles and the background scene reflectance which varies according to illumination.

Currently, haze extraction algorithms exist which remove the haze in a scene. This is ideal from the perspective of enhancing vision but finding the relationship between the extracted haze and overall global PM levels from ground-based images is an under researched problem. We compare existing dehazing algorithms in the context of monitoring atmospheric scattering and present a method of more accurately measuring atmospheric scattering from a sequence of images captured over time. This provides an important step in estimating PM much more practically using imaging devices. The most hazardous subcategory of PM is known as PM10 and consists of particles with diameters less than 10 micrometers. These particles make up most of the haze in urban scenery and greatly affect the visual perception of the scene. This research presents an approach to measure atmospheric scattering from image pixel values using simple off the shelf equipment consisting of a Single-Lens-Reflex (SLR) camera, polarizer filter and color filters.

# 1.2 Motivation

With the escalating concern over changes in our environment due to man-made pollution, a need for easier and simpler means of simulating and monitoring airborne pollution has risen. Air pollution that causes visual degradation in the appearance of distant urban scenery and is hazardous to humans and the environment is known as particulate matter (PM). PM consists of aerosol particles suspended in the atmosphere. These aerosol particles greatly affect the appearance of scenery by scattering light. Particulate matter is made up of a number of components, including acids (nitrates and sulfates), organic chemicals, metals, soil particles, dust particles and soot particles [6]. The size of aerosol particles is inversely proportional to the hazardous effect they cause. PM consisting of particles with diameter equal to or less than 10 micrometer is able to bypass the throat and lungs of humans and cause damage to the heart and lungs.

The motivation for this research lies in the fact that current dehazing algorithms remove all the haze present in a scene and hence produce unnatural images for the sole purpose of enhanced vision. We approach this problem from an environmental perspective. When comparing extracted knowledge of atmospheric scattering taking place in images taken over a period of time we are able to measure the fluctuations in PM. We offer a visual quantification of atmospheric scattering and hence man-made PM which fluctuates proportionally to the changes in atmospheric scattering. This is because the background scattering of molecules in the atmosphere is constant and the only change in the atmosphere is caused by the changes in PM levels. We want to visualize and simulate the effects of PM haze which is both valuable from a computer graphics perspective and an environmental monitoring standpoint. Most current inexpensive methods of measuring particulate matter (PM) do not measure the global concentration. Instead, these methods measure PM levels locally at the site of the device. In order to get an estimate of the average particle pollution level, they combine the readings of multiple devices over a large area. We also want to produce a more accurate depthmap of the scene by using our algorithm. More accuracy in depth measured from images is important as when dealing with vast distances, the smallest of errors can be quite inaccurate.

With the enormous increase in the number of cameras found throughout modern cities, this approach will fit right in. The visual data is lying around unutilized, when it could provide us with valuable knowledge. It will enable us to simulate and monitor PM. PM is a health hazard to humans as they are able to infiltrate the body and cause damage to the lungs and heart. Our approach will provide a more practical step to effectively determining the PM concentrations in the atmosphere to be able to monitor this hazardous air pollution more closely.

The work in this area of measuring, removing and visualizing air pollution in scenery captured from terrestrial viewpoints is very limited. We present an approach which will add a vital step into an unchartered research field.

# 1.3 Thesis Goal

The goal of this research is to more accurately measure atmospheric scattering from digital images of urban scenery in order to accurately estimate PM levels in the atmosphere. We intend to develop a process which takes in digital images, analyzes the pixel data and measures the amount of atmospheric scattering in the scene. This process will include radiometric calibration of the camera being used, extracting haze from images using a polarization-based algorithm, applying a temporal dehazing framework and applying a statistical prior of out-door haze filled scenes. Another goal of this thesis is to produce a more accurate depth map computation method which produces a more accurate depth map of scenes by using temporal knowledge. To sum up, the goals of this thesis are:

- Polarization-based haze extraction from image
- Temporal haze extraction from images
- Haze removal from images using statistical prior
- Measure atmospheric scattering from digital images of a scene captured over time
- Recover a more accurate depthmap of a scene using a sequence of digital images captured over time
- Compare our algorithms of measuring atmospheric scattering and depth with existing dehazing methods

This leads us to the thesis statement:

*Measuring atmospheric scattering from digital images of urban scenery and recovering more accurate depthmaps of the scene. We present two algorithms based on temporal knowledge which measure atmospheric scattering, recover a more accurate depth-map of the scene and remove haze from the images. As fluctuations in atmospheric scattering are directly proportional to the changes in levels of air pollution, we provide a valuable step in measuring air pollution from digital images. We use existing haze extraction algorithms along with a camera and polarizer filter to achieve this.*

The contribution of this thesis is an algorithm for measuring the visually degrading effects of PM from digital images of urban scenery. We do this by measuring the atmospheric scattering which is directly proportional to the levels of PM. Fluctuations in atmospheric PM results in correlated fluctuations in atmospheric scattering. This thesis bridges between previous researches in haze image extraction and environmental monitoring. We provide an important step in measuring PM from digital images. In addition to this, we recover more accurate depth estimation from a sequence of digital images of a haze-filled scene taken over time. This thesis provides a valuable contribution in the process of quantifying PM from pixel data in digital images. It will lay out a framework to more accurately and practically measure hazardous levels of PM through visual means.

# Chapter 2 – Background

In this chapter we describe the theoretical foundation of the proposed research. We start off by defining particulate matter (PM), airlight (haze) and current methods of measuring PM. Then we will describe the theory that this research is based upon, which is the physical interaction of light and the atmosphere. The physical models of how light changes and interacts with airborne molecules and particles is studied in order to understand the resultant irradiance captured by images of distant scenes. Light undergoes changes as it passes through the atmosphere. Its wave patterns, intensity and direction changes as the photons come into contact with molecules and aerosols in the atmosphere. We describe a phenomenon known as polarization which happens to light waves as they are scattered by molecules and particles. Then we will explain photographic methods of dealing with this phenomenon known as light polarization. We will then look at the physics of light scattering.

## 2.1 Particulate Matter (PM)

According to the US Environmental Protection Agency (EPA) [6],

*"Particulate matter, also known as particle pollution or PM, is a complex mixture of extremely small particles and liquid droplets. Particle pollution is made up of a number of components, including acids (such as nitrates and sulfates), organic chemicals, metals, and soil or dust particles. The size of particles is directly linked to their potential for causing health problems"*

*"EPA is concerned about particles that are 10 micrometers in diameter or smaller because those are the particles that generally pass through the throat and nose and enter the lungs. Once inhaled, these particles can affect the heart and lungs and cause serious health effects"*

The EPA classifies particles which make up PM into two categories:

1. **Inhalable coarse particles** – found close to roads and dusty industries. These particles are larger than 2.5 micrometers and smaller than 10 micrometers in diameter.
2. **Fine particles -** found in smoke and haze and are 2.5 micrometers in diameter and smaller. These particles can be directly emitted from sources such as forest fires, or they can form when gases emitted from power plants, industries and automobiles react in the air.

PM is not only hazardous to human health, but also affects visibility. Particulate matter is the main cause of atmospheric visibility reduction due to haze [7]. Haze is seen as a bluish or brownish obscuration in distant scenes.

## 2.2 Methods of Measuring PM

In this section we describe the current methods of measuring PM10 concentrations. According to NASTRO [9], which is a public and private partnership dedicated to improving the management of air quality in North America, the measurements of PM10 are mainly done using the following methods:

1. Gravimetric Methods – exchangeable filters (FRM)
2. Beta Attenuator Methods (FEM)
3. Tapered Element Oscillating Microbalance (TEOM)

Gravimetric Methods - exchangeable filters (FRM) methods use an air pump to draw in ambient air at a constant rate into a specifically shaped inlet where particulate matter is separated by size. Particulate matter is then collected on a filter. The filters are weighted before and after, to determine the net mass increase due to the collected matter. This net mass of particulate matter is measured relative to the volume of air intake which produces a measurement of micrograms per cubic meter of particle matter.

Beta Attenuator Methods (FEM) Beta particles are electrons with a certain energy range. These particles are attenuated according to an approximate exponential function when they pass through particulate deposits on a filter tape. Automated samplers use a filter tape. The unexposed tape is subjected to beta particles and the attenuation is measured and compared to the same measurement done on the same tape but after exposing it to ambient air flow. The particulate matter from the air deposits on the tape and causes a difference in the attenuations measurements. This difference is used to calculate the mass of particulate matter per volume of air.

Tapered Element Oscillating Microbalance (TEOM) consist of a tapered glass element with a filter attached which draws in air. The element oscillates according to a characteristic frequency that decreases as mass accumulate on the filter. This measurement of frequency is used to calculate the accumulated mass of particulate matter which then gives a measure of the PM concentration.

There are many other methods of measuring PM concentrations under research and development and not standard methods used. Some other methods are filter-based. Some apply light or lasers to samples of ambient air and calculating the PM concentration using the scattering and absorption of light [10 & 11]. LIDAR (Light Detection and Ranging) is an optical remote sensing technology that can be used to measure PM concentrations [11]. There is also a method of measuring, which measures the shock waves (acoustic energy) caused by the impact of particles with a probe inserted into the air flow [10].

# 2.3 Airlight (Haze)

Airlight or haze is the brownish or bluish color we see when we look at distant objects. This visible effect is due to the scattering of light by the atmosphere towards the viewer. The atmosphere consists of a mixture of molecules and particles of various sizes. The following figure illustrates what happens when we take a picture of distant objects through the atmosphere during the daytime.

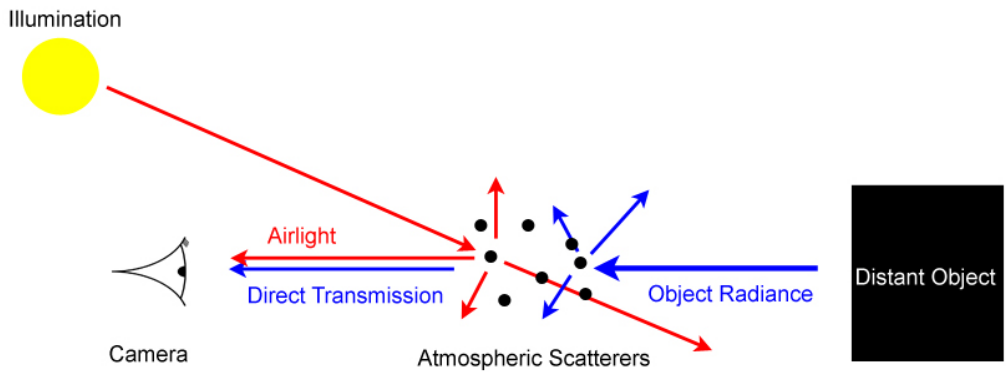

**Figure 1: This diagram shows what happens when we view a distant object through the atmosphere.**

The main illumination on a clear sunny day is the sun. The illumination from the sun is scattered by molecules and particles in the atmosphere in all different directions. Some of this scattered light is scattered towards the viewer (camera) which results in an additive radiance that increases the brightness of distant objects. This is known as airlight. The further away an object is, the brighter it appears. This is because the more atmospheric medium there is between a viewer and an object; the more scattering of light towards the viewer takes place. The object radiance is also attenuated due to scattering and absorption as it travels towards the viewer. The object radiance is partially scattered out of the line of sight and absorbed by the particles in the atmosphere. The direct transmission is the light radiance that remains after the object radiance is attenuated along the path towards the camera.

# 2.4 Light Polarization

Polarization is a property of transverse waves. It describes the orientation of vibration of the wave in the cross sectional plane perpendicular to the direction of motion. Light is a transverse electromagnetic wave. A light wave travelling forward can oscillate up and down, side to side or in any intermediate direction. Normally a light wave is composed of a collection of waves vibrating in all directions perpendicular to its line of propagation. If the light waves vibrate consistently in a particular orientation as it moves forward, the light is said to be polarized.

Natural light emitted from the sun is unpolarized. This means that the electromagnetic waves are oscillating in random directions about the axis of motion. When unpolarized light is reflected the resultant light is always polarized to a certain extent. Light can also be polarized by both double refraction and scattering in gases. The latter form of light polarization is of concern in this research and will be looked at in-depth.

As light rays pass through a medium it is scattered by the molecules and any suspended particles in that medium. The scattered light is partially polarized. A polarizer filter as we will discuss next can be used to control the capture of this scattered polarized light when taking pictures.

Polarizer filters (also known as Polaroid filters) are part of the arsenal of optical filters that professional photographers use. Polarizer filters are placed over the lens of a camera and can be rotated to different angles. As the filter is rotated it determines the amount of polarized light is filtered in and out of the lens. There are two types of polarizer filters: circular polarizer filters and linear polarizer filters. The former is used for cameras with through-the-lens metering systems or autofocus.

The basic concept behind these filters in our context is that scattered light is partially polarized perpendicular to the plane of incidence. The plane of incidence is created between the illumination source, the scattering particle and the camera. Referring to figure below, you can see that the polarizer filter when oriented to the vertical position will filter out any light polarized in any other direction. Only the light polarized in the vertical direction is let in through the lens. The axis of the filter can be rotated to any angle which will only let in the light polarized at that certain angle.

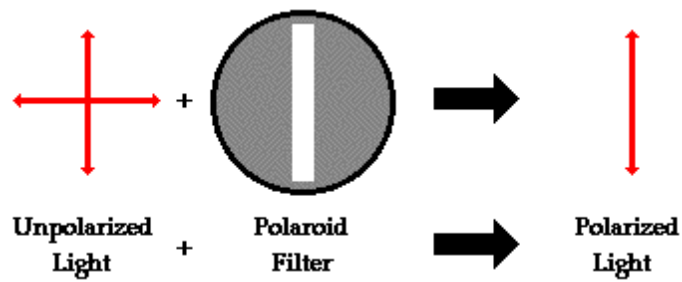

Figure 2: A Linear polarizer filter [19]

## 2.5 Light Scattering

When a light wave approaches a particle (molecule or suspended particle) it can reflect off it, diffract around it or refract into it. Once the wave has entered the particle, it can be absorbed, transmitted through, refracted out or reflected internally one or more times and then refracted out. All these processes except for absorption are responsible for light scattering when light waves pass through the atmosphere.

Light scattering can occur multiple times in translucent gaseous mediums like that of the atmosphere. This happens when light is scattered off particles and is then scattered again by other particles in the medium. This can go on recursively for a number of times until the effect is negligible. With the increase in optical thickness of any suspended particulate matter, the multiple-scattering of light increases. Since we are dealing with low optical thicknesses like that of the earth's atmosphere at ground-level, we can safely approximate by assuming only single-scattering.

The amount of light scattered by a particle depends on the angle between the incident beam and the scattered beam. This angle is known as the scattering angle and can be seen in figure 3. This angle is the angle between the forward direction of the incident beam and the scattered beam direction. The distribution function which maps the scattering angle to the scattered light intensity in the direction of the scattering angle is known as the phase function.

The phase function describes the distribution of the scattered light about a particle. The phase function varies with the scattering angle, shown as $\theta$ in the figure below. The scattering angle is the angle of deviation measured from the direction of the incident radiation to the scattered beam. Backward scattering occurs at an angle of 180 degrees and forward scattering occurs at 0 degrees.

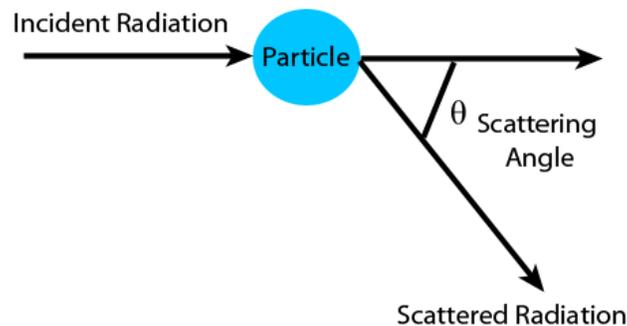

Figure 3: Scattering of incident radiation by a particle

The scattering of light off particles varies with the size and shape of the particles. For simplicity we will make the assumption that we are dealing with spherical particles. The diameter of spherical particles affects the distribution of light. The wavelength of light and the type of particles (e.g. soot, dust, and combustion particles) also affect the scattering. For simplicity again, we will assume a uniform mixture of atmosphere over the small time span during which we will do our measurements.

Airborne particles can be broadly classified into three regimes according to the size of the particles relative to the wavelength of light. The regimes are:

1. Rayleigh regime
2. Mie regime
3. Geometric regime

Rayleigh regime consists of particles with diameters much smaller than the wavelength of the incident visible light radiation (0.38 micrometers to 0.75 micrometers – figure 4). The condition for this regime is $d < 0.03w$, where $d$ is the particle diameter and $w$ is the wavelength of light. Mie regime consists of particles with diameters comparable in size to the wavelength of light which fit the following condition:

$0.03w < d < 32w$. The geometric regime consists of the particles larger in size than the two previous regimes ($d > 32w$). The three regimes, diameter ranges with respect to the wavelength of light and average particle sizes are shown in the table below.

| Rayleigh regime | $d < 0.03w$ | Particles in this regime have diameters approximately from 0.01 to 0.02 micrometers |
| Mie regime | $0.03w < d < 32w$ | Particles in this regime have diameters approximately from 0.02 to 20 micrometers |
| Geometric regime | $d > 32w$ | Particles in this regime have diameters approximately from 20 to 1000 micrometers |

**Table 1: The classification of particles into the three regimes according to the particle diameter with respect to the wavelength of light**

The visible light spectrum falls in a range of wavelengths from 380nm (0.38 micrometers) to 750nm (0.75 micrometers) [7]. The red color light range is from 600nm to 750nm. The Green color light range is from 500nm to 600nm. The blue color light range is from 380nm to 500nm. We will see at the end of this chapter that color filters can be attached to the lens of cameras and filter through a band of wavelengths along the visible spectrum. This will be useful to narrow down the wavelength of light captures in the images to get more precise estimations of particle sizes and concentration.

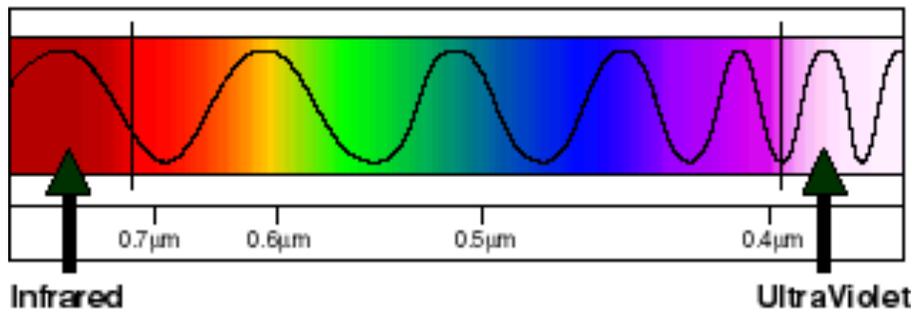

**Figure 4: The visible electromagnetic radiation spectrum [21]**

Light scattering is selective. This means that different sizes of particles scatter different wavelengths of light more than others. This can be explained by Rayleigh scattering and seen every day in nature as we observe the colors of the sun and sky. The atmosphere is mostly composed of molecules (small particles) and Rayleigh scattering is inversely proportional to the fourth power of the wavelength. This means that the shorter wavelengths of blue light will scatter more than the longer wavelengths of green and red light. This explains why the upper atmosphere of earth scatters more blue light down to us which makes the sky appear blue during the day. This is also why the setting sun appears orange or reddish. This is because the sun's rays are passing through much more atmosphere compared to when the sun is overhead. The increase in atmosphere means an increase in scattering out of the direction of the illumination. Since more blue and green light than red light is scattered away, the sun appears reddish. We will discuss the physics responsible for these natural phenomena in the following section.

As we are interested in PM10 concentrations in this research, we will focus on the Rayleigh and Mie regimes as the particles fall into these two categories. The scattering of light in the Rayleigh and Mie regimes are looked at in the following sections.

## 2.5.1 Rayleigh Scattering

Rayleigh scattering theory was derived by Lord Rayleigh in 1871. This law models the scattering of light by particles much smaller than the wavelength of light. This primarily occurs as light radiation travels through gases and is scattered by the molecules of that gas. The blue color of the sky has been explained and proved by the Rayleigh scattering law. The molecules present in the upper atmosphere of the earth are wavelength selective and scatter more blue light than the rest of the colors in the visible spectrum. Due to this, more blue light is scattered towards us which makes the sky appear blue. Rayleigh scattering also makes the setting sun appear reddish. This is because more blue and green light is scattered out of the line of sight as you look at the sun resulting in an orange or reddish appearance of the sun.

Rayleigh scattering is categorized by the following criteria:

$$2\pi r/\lambda \ll 1 \qquad \qquad 2.5.1.1$$

where $r$ is the radius of the particle and $\lambda$ is the wavelength of visible light. This means that the particles considered here are much smaller than the wavelength of light. This is the basic condition that causes Rayleigh scattering.

The model describes an approximated scattering phase function which describes the scattering of radiation by individual particles. By particle here we mean molecules and small particles. The scattering phase function for Rayleigh scattering (assuming spherical shaped particles) is shown below, according to [7]:

$$P(\theta) = \tfrac{3}{4}(1 + \cos^2(\theta)) \qquad \qquad 2.5.1.2$$

where $\theta$ is the scattering angle between the incident beam and the scattered ray of light. An angle of 0 degrees means forward scattering along the direction of the incident beam. An angle of 180 degrees constitutes backward scattering.

The phase function distribution looks like figure 5. The incident ray of light is coming from the left hand side and is scattered about in an ellipse-like distribution. The figure below is a 2D cross-sectional illustration. The actual distribution is a 3D ellipsoid-like shape symmetric about the horizontal axis in the illustration below.

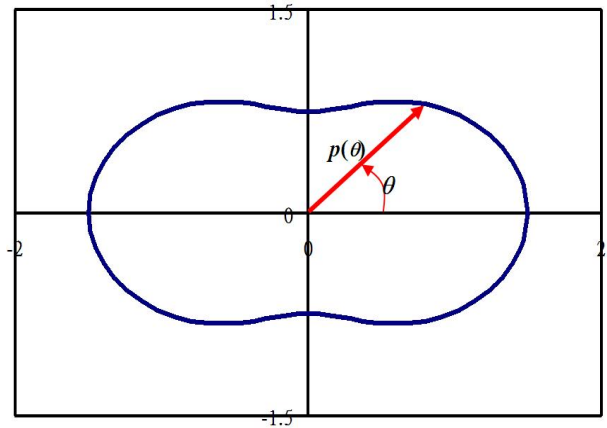

**Figure 5: Rayleigh phase distribution [20]. Figure 8 shows a comparison of Rayleigh scattering and Mie scattering.**

Notice from the figure above that the forward and backward scattering is more than the sideways scattering (almost double).

The degree of polarization defines how much light is polarized. The degree of polarization is a value from 0 to 1 where 0 means that the light is not polarized in any orientation and 1 means that the light is fully polarized in a certain orientation.

The phase function for the degree of polarization of the scattered light in terms of the scattering angle $\theta$ is also defined for Rayleigh scattering and is shown below, according to [1].

$$P_{pol}(\theta) = \sin^2(\theta)/(1+\cos^2(\theta))$$  2.5.1.3

Here, we are assuming that the particles are spherical in shape as the degree of polarization is considerably reduced by the presence of non-spherical particles. In reality aerosols come in all shapes and sizes. According to [7] & [30] most aerosol particles are spherical in shape or close to spherical in shape and therefore the assumption that aerosols are spherical gives a reliable approximation of reality.

Scattering in the Rayleigh regime is very selective according to the wavelength of the incident light. It is very wavelength-dependent. The intensity of light scattered by a Rayleigh scatterer is inversely proportional to the fourth power of the wavelength. As blue light has the smaller wavelength it is scattered much more than green and red light which have longer wavelengths.

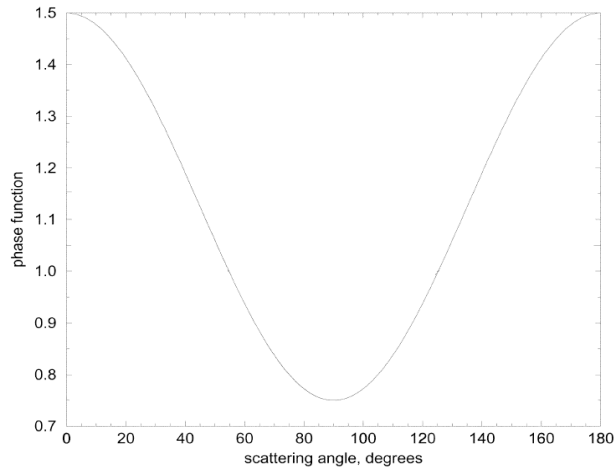

**Figure 6: Rayleigh scattering phase function plotted against the scattering angle [13]**

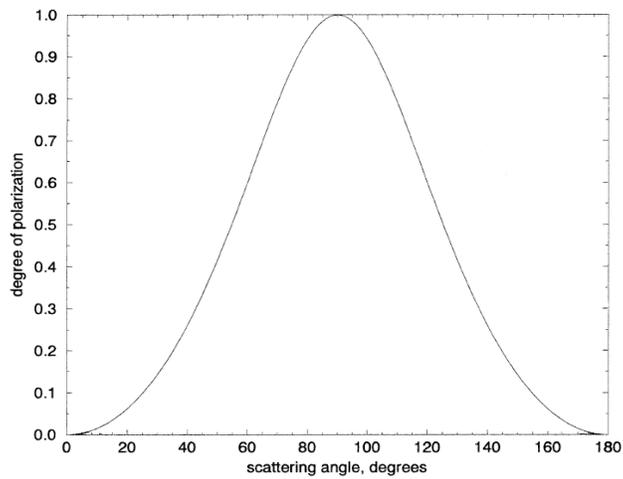

**Figure 7: Degree of Polarization against the scattering angle [13]**

As figure 6 shows, the intensity of scattered light is equal in the forward and backward scattering directions and almost halved in the directions perpendicular to the incident beam of light. The degree of polarization as shown by figure 7 is quite the opposite as the maximum degree of polarization is achieved perpendicular to the incident beam. In reality the maximum degree of polarization is less than the maximum shown in figure 7. According to Alexander Kokhanovsky's book called Light Scattering Media Optics [13], the degree of polarization shown in figure 7 can be considered an upper bound as in reality this degree of polarization is decreased by the presence of aerosols, non-spherical particles, ground reflection and multiple scattering (which increases with particle size and optical thickness).

## 2.5.2 Mie Scattering

Mie's theory, also known as Lorenz-Mie theory and Lorenz-Mie-Debye theory, is the generalization of Rayleigh theory and encompasses spherical particles similar or slightly larger in size than the wavelength of light. The diameter of particles is comparable to the wavelength of light (between 0.03 and 32 times the wavelength of visible light (0.38 to 0.75 micrometers)). Particles that satisfy this condition are known to be within the Mie regime and tend to scatter more light in the forward direction than in the backward direction and less light in the sideway directions. Mie scattering is valid for all possible ratios of diameter to wavelength, unlike Rayleigh scattering which is valid only for diameters that are much smaller than the wavelength [35]. Mie scattering, unlike Rayleigh scattering, is not strongly wavelength dependent.

The phase function which models the scattering of light by Mie regime particles is approximated by the Henyey-Greenstein function which is as follows, according to [7]:

$$P(\theta) = \frac{1-g^2}{(1+g^2-2g\cos\theta)^{\frac{3}{2}}}$$  2.5.2.1

The asymmetric parameter ($g$) is a parameter derived from the phase function and gives the relative direction of scattering by particles or gases. If the asymmetric parameter equals 0 the phase function models isotropic scattering which means that the radiation is scattered equally in all directions. As the value approaches +1 the scattering peaks strongly in the forward direction. As the value approaches -1 the scattering strongly peaks in the backward direction. According to [13] approximate values of the asymmetric parameter are 0.8 and 0.3 for dust aerosols and soot aerosols, respectively.

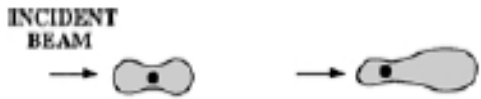

**Figure 8: Shows Rayleigh scattering (left) which scatters the same intensity symmetrically backwards and forwards. Mie scattering is predominantly in the forward direction as seen on the right [31].**

There is no simple equation or solution to model the degree of polarization caused by Mie scattering. As the aerosols are much larger than the background molecules and vary in shape and size, modeling the polarization is no easy task. There have been many models proposed to deal with different types of aerosols (depends on shape, size and chemical composition), different thicknesses of atmospheric layers and different refractive indices. A comprehensive look at environmental light polarization can be seen in [3].

# Chapter 3 – Previous Work

There has not been much research in the area of measuring particulate matter, particularly PM10, by visual means from digital images taken at ground-based viewpoints. The few researches relevant to our proposed research are described in this section. We describe two techniques for measuring PM10, both of which are not computer vision techniques as they rely on simply measuring the image brightness values and relating this to physics models. We also describe two computational techniques for extracting haze from digital images.

# 3.1 Image Processing Technique for Measuring Air Quality

One of the few relevant ground-based approaches is a paper called "Using Image Processing Technique for the Studies on Temporal Development of Air Quality" [14]. This paper describes a method of measuring PM10 concentrations using a video surveillance camera. The method uses distant green vegetation as a reference target with known spectral reflection to calculate the added reflection due to atmospheric scattering. It then uses atmospheric models borrowed from satellite imagery analysis to compute the PM10 concentrations.

Next, we rewrite the algorithm for PM10 concentration measurement presented in this paper. The total reflection measured by the camera is defined by,

$R_{total} = R_{ref} + R_{atm}$                       3.1.1

$R_{ref}$ is the reference target reflection (distant green vegetation) and $R_{atm}$ is the atmospheric reflection due to atmospheric scattering. The atmospheric reflection due to atmospheric scattering is composed of two parts; part corresponding to the added reflection due to scattering by molecules (Rayleigh scattering) and the other part corresponding to the added reflection due to aerosol scattering. The atmospheric reflection due to scattering is shown as,

$R_{atm} = R_a + R_r$                          3.1.2

$R_a$ is the reflection due to aerosol scattering and $R_r$ is the reflection due to the molecule Rayleigh scattering.

The algorithm presented in this paper uses an atmospheric reflection model which is described in section 4.2 (Based on Liu et al. [5]). The reflection model is based on research in measuring the atmospheric optical thickness from images taken by satellites. The reflection model for molecular scattering is,

$R_r = \frac{\tau_r P_r(\theta)}{4\mu_s \mu_v}$                            3.1.3

where $\tau_r$ is the optical thickness due to molecules in the atmosphere, $P_r(\theta)$ is the phase function for molecules (Rayleigh scattering phase function), $\mu_s$ is the cosine of the solar zenith angle and $\mu_v$ is the cosine of the viewing zenith angle.

Equation 3.1.3 shows that the reflection due to molecular scattering is linearly related to the optical thickness. Similarly the atmospheric reflectance due to aerosol particles is also linearly related to the optical thickness of aerosols. This is shown by King et al. and proved by the fact that Liu et al. [14] also found the linear relationship between both aerosol and molecular scattering.

$$R_a = \frac{\tau_a P_a(\theta)}{4\mu_s\mu_v} \qquad \qquad 3.1.4$$

where $\tau_a$ is the optical thickness due to aerosols, $P_a(\theta)$ is the phase functions for aerosol scattering respectively, $\mu_s$ is the cosine of the solar zenith angle and $\mu_v$ is the cosine of the viewing zenith angle.

Equations 3.1.3 and 3.1.4 can be substituted into 3.1.2 to form,

$$R_{atm} = \frac{1}{4\mu_s\mu_v}[\tau_a P_a(\theta) + \tau_r P_r(\theta)] \qquad \qquad 3.1.5$$

The optical thickness $\tau$ is expressed as,

$$\tau = \sigma\rho s$$

where $\sigma$ is the absorption parameter, $\rho$ is the density parameter and $s$ is the finite path.

The total optical thickness is equal to the addition of the aerosol optical thickness and the molecular optical thickness, which is:

$$\tau = \tau_a + \tau_r = \sigma_a\rho_a s + \sigma_r\rho_r s$$

Therefore, equation 3.1.5 becomes,

$$R_{atm} = \frac{s}{4\mu_s\mu_v}[\sigma_a\rho_a P_a(\theta) + \sigma_r\rho_r P_r(\theta)] \qquad \qquad 3.1.6$$

Since $\sigma_a, \sigma_r, P_a(\theta), P_r(\theta)$ are all dependent on the wavelength of light $\lambda$,

$$R_{atm}(\lambda) = \frac{s}{4\mu_s\mu_v}[\sigma_a(\lambda)\rho_a P_a(\theta,\lambda) + \sigma_r(\lambda)\rho_r P_r(\theta,\lambda)] \qquad \qquad 3.1.7$$

In the above equation $\rho_a$ is the particle concentration (PM10) and $\rho_r$ is the molecule concentration.

Two equations are creating by substituting two different wavelengths of light. These two equations are solved for $\rho_a$ simultaneously. This gives the particle concentration as,

$$\rho_a = a_0 R_{atm}(\lambda_1) + a_1 R_{atm}(\lambda_2) \qquad \qquad 3.1.8$$

where $a_0$ and $a_1$ are coefficients that are empirically determined through correlation with live data.

Equation 3.1.8 shows that the PM10 concentration is linearly related to the reflectance for wavelength 1 and wavelength 2. Retalis et Al. [14] found that PM10 was linearly related to the optical thickness and hence that PM10 is linearly related to the reflectance.

A spectroradiometer is then used to measure the solar radiation at ground level. The total reflection detected by the camera is the irradiance measured by the camera divided by the solar radiation measured. The reflectance of the known target is subtracted from this total reflectance according to equation 1 to reveal the reflectance due to the atmosphere. This is done for two bands. Actual data of particulate matter and the atmospheric reflectance for both bands are substituted into equation 8 to form a correlation between particle concentration and atmospheric reflectance. This was done by regression analysis. This approach is shown to correlate strongly with a Dusttrak meter measurements taken at the same time the images were taken.

This method of measuring PM10 concentrations is very restrictive. It is restrictive because it relies on a reference target in the scene which is the vegetation in the background. The reflectance of this vegetation has to be measured beforehand to be able to extract the reflectance due to atmospheric scattering. This restrictiveness makes it a non-passive approach which relies on previously known measurements. A spectroradiometer is also used to measure the solar irradiance which makes the approach less automated. The reflectance model used in this paper is based on satellite remote sensing and may not represent the actual physics of the terrestrial atmosphere accurately. The phase functions of Rayleigh scattering and Mie scattering used in this paper are not described. This is because they are treated as part of a black box approach in which the overall summation of the atmospheric reflectance due to both scatterings over two wavelength bands is correlated to the PM10 concentration. The results obtained can largely vary according to the phase functions used, specifically for Mie scattering which is not so straight forward. An important factor here is that the phase function is the distribution of light scattered by a particle and is a function of the scattering angle, therefore, unless the images were taken at the same time of day where the sun was approximately at the same position, the phase functions would give nonlinear results. This would affect the stability of the linear relation between PM10 concentration and atmospheric reflectance due to scattering which is proposed in this paper. This may also make this paper a restricted method of measuring PM10 levels and gives a very rough correlation between atmospheric scattering and PM10 concentration. We intend to build a more efficient model of measuring solar irradiance and its effect on the atmospheric reflectance due to scattering by including the Lambert-Beer law [12] which measures the solar irradiance after passing through the earth's atmosphere. We hope to build on this research and create a passive remote monitoring method of measuring PM10 concentrations. Our approach is passive in the sense that it does not require previous measurements or manual work in order to function. We will use a polarization-based approach which eliminates the need to use a reference target and we will not rely on a spectroradiometer to discover the atmospheric reflectance.

# 3.2 Satellite Image Remote Sensing

The research on remote sensing of the atmosphere from satellite imagery has been a very deeply research field. Among the many research papers written about the subject, we have found one certain research paper relevant. This paper describes a satellite image-based retrieval algorithm for aerosol characteristics and surface reflectance [15]. The algorithm assumes that the surface of the earth is non-uniform and Lambertian (diffuse reflection).

The paper focuses on modeling the reflectance of the atmosphere seen from satellites. The model of interest here is the model of the atmospheric reflectance due to scattering along the path of sight. This model simulates the scattering that happens as sunlight radiates down onto the earth's atmosphere and is scattered back towards the satellite. This reflectance due to scattering is the summation of molecular and aerosol scattered radiance. A single-scattering approximation is used to model the reflectance of molecular scattering. This model is simplified to the following model,

$$R = \frac{\tau P(\theta)}{4\mu_s \mu_v}$$
3.3.1

$\tau$ is the optical thickness, $P(\theta)$ is the scattering phase function of the scattering angle $\theta$, $\mu_s$ is the cosine of the solar zenith angle and $\mu_v$ is cosine of the viewing zenith angle. The phase function here can be substituted by the Rayleigh phase function. $\mu_s$ and $\mu_v$ increase as the viewing zenith angle (satellite zenith angle) and solar zenith angle increase respectively. The reasons for these angles affecting the atmospheric reflectance due to molecular scattering are due to the fact that more of the atmosphere contributes to the scattering as the angles increase. For a fixed point on the earth's surfaces and a fixed viewing direction; as the solar zenith angle increases more sunlight falls on the atmosphere along the line of sight. This causes an increase in reflectance due to scattering. For a fixed point on the earth's surface and a fixed solar zenith angle; as the viewing zenith angle increases the amount of atmosphere along the line of sight increases and therefore more scattering takes place.

As the solar zenith angle approaches 90 degrees the model increases to infinity. According to Alexander Kokhanovsky; when the sun is low on the horizon the plane parallel model of the earth's atmosphere no longer holds and a night-time illumination model should be used and hence the above model will not hold.

The research just described is for satellite remote sensing. The geometry of the reflectance model models satellite geometry where $\mu_s$ and $\mu_v$ describe the zenith angles of the sun and satellite, respectively. We can see that the reflectance due to scattering is proportional to the atmospheric optical thickness and the phase function of the scattering angle. This is the essence of the model if we ignore the denominator which is satellite geometry dependent. The proportionality here also applies to terrestrial sensing as the sunlight is also reflected by the terrestrial optical thickness of the atmosphere and the denser the optical thickness, the more light is reflected back to the ground-based observer. The reflectance due to scattering at the ground-level also varies greatly with the phase function as described by Rayleigh and Mie scattering.

# 3.3 Polarization-based Haze Extraction

Polarization-based vision is the use of polarizer filters to selectively control the light that enters the camera lens. Light is polarized due to scattering in the atmosphere and this is the basic premise of [1]. The basic principle behind this approach is the fact that airlight is partially polarized and dominates the light radiating towards a camera in distant scenes in daylight. The radiance coming from the object in the scene is minimally polarized relative to the airlight as it suffers attenuation due to scattering along the way to the camera while the airlight increases with distance. For this reason, the airlight can be assumed to be polarized to some degree and removed by the use of a polarizer filter which is oriented in such a way as to filter out the partially polarized airlight. Optics alone can remove most of the haze when the sun is perpendicular to the viewing direction (polarization is maximum when viewing perpendicular to the illumination direction). This is said to be the trivial case where the degree of polarization (DOP) of the airlight is closest to 1 (maximum). The general case is unlike this, as the sun can be located at any position therefore limiting the DOP. This makes optics alone inefficient in dehazing scenes. Based on this general case, the authors introduce an important image dehazing algorithm which further dehazes the captured scene. We will first look at the image formation model which is the foundation of this paper and then outline the algorithm.

The image formation model described in this paper explains the formation of the image irradiance $I$ which is finally captured by the camera. The objects in the scene reflect light towards the camera. This light is known as the object radiance. This radiance goes through the atmosphere as it travels towards the camera and therefore suffers attenuation mainly due to scattering. It decreases exponentially over distance. The remaining light which reaches the camera after attenuation is known as direct transmission. Airlight is the extra brightness that we see as objects get further away. This is due to the scattering of global illumination (mainly the sunlight irradiance) by the atmospheric medium. Part of this light is scattered towards the camera, therefore the further away an object is, the more the atmosphere there is scattering light towards the camera. The final image irradiance captured by the camera $I$ is composed of the direct transmission $D$ and the airlight radiance $A$ as follows,

$$I = D + A \qquad \qquad 3.3.1$$

$D$ decomposes further into the following,

$$D = Re^{-\beta z} \qquad \qquad 3.3.2$$

where $R$ is the actual scene object radiance, $\beta$ is the scattering coefficient of the atmosphere and $z$ is the depth of the scene object relative to the camera. $A$ decomposes to,

$$A = A_\infty (1 - e^{-\beta z}) \qquad \qquad 3.3.3$$

where $A_\infty$ is the airlight radiance at infinite distance. We will see shortly how $A_\infty$ is computed. The overall airlight component increases exponentially with the increase in depth. This gives us the overall equation of image formation,

$$I = Re^{-\beta z} + A_\infty(1 - e^{-\beta z}) \qquad 3.3.4$$

$I$ is the sum of the image irradiance of the image taken as the polarizer filter is set perpendicular to the plane of incidence $I^\perp$ (best polarization image) and the image irradiance of the image taken as the polarizer filter is parallel to the plane of incidence $I^{||}$ (worst polarization image). The plane of incidence is formed between camera, illumination (sun) and the scattering particle.

$$I = I^\perp + I^{||} \qquad 3.3.5$$

Now that we have seen the process of image formation based on modeling the scattering that takes place in the atmosphere, we can outline the dehazing algorithm which dehazes the scene. The algorithm can be done using two images taken at different orientations of the polarizer filter. The most stable results are obtained by taking the pictures with the polarizer filter oriented perpendicular and parallel to the plane of incidence. The algorithm estimates two global parameters: $A_\infty$ and $p$. $A_\infty$ is the airlight radiance corresponding to an object at infinite distance which is dominated by airlight and thus cannot be seen through this added brightness. $p$ is the degree of polarization at any given scene point and is computed by,

$$p = (A^\perp - A^{||})/(A^\perp + A^{||}) \qquad 3.3.6$$

where $A^\perp$ and $A^{||}$ are the airlight components at each pixel in the perpendicular and parallel directions, respectively, relative to the plane of incidence. $A_\infty$ is obtained by averaging the summation of $A^\perp$ and $A^{||}$ over patches of sky close to the horizon. $A_\infty$ used to discover the degree of polarization $p$ and transmittance. Since in the low sky region of the scene, all we see are pixels completely dominated by airlight as the direct transmission has been totally overcast. This gives us a pure airlight value from which we discover the DOP of airlight in the scene and the transmittance of the atmospheric medium in the scene.

For each pixel in the images the two irradiance values are extracted. These are $I^{||}$ and $I^\perp$ which correspond to the best and worst polarized images, respectively. The airlight component is calculated by,

$$A = (I^\perp - I^{||})/p \qquad 3.3.7$$

The direct transmission of the scene is the airlight component subtracted from the total irradiance,

$$D = (I^\perp + I^{||}) - A \qquad 3.3.8$$

which removes the additive effect of airlight. As the scene radiance moves through the atmosphere it is attenuated by the atmosphere. The transmittance of the atmosphere is measured by,

$$t = 1 - A/A_\infty \qquad \qquad 3.3.9$$

Therefore the final object radiance compensated for airlight radiance addition and attenuation due to haze is,

$$R = \frac{(I^\perp + I^{||}) - A}{t} \qquad \qquad 3.3.10$$

This produces the final dehazed image. This is the basic algorithm descried in the paper; although the authors go on to describe two byproducts of this process. A range map is the first. The paper describes the creation of a range map which shows the relative depth of the scene in grayscale. This is produced from the dependence of the transmittance $t$ on the distance. The second byproduct is rudimentary information about the aerosols. This information can be produced from the extinction coefficients of each color channel (RGB). The extinction coefficient is defined as,

$$\beta z = -\ln t \qquad \qquad 3.3.11$$

From this relationship the extinction coefficients can be established for each color channel (RGB). This is valuable because the extinction coefficient encompasses the scattering coefficients which are determined by the size of the scattering particles. Assuming that the scattering is the dominant cause of radiance extinction (attenuation) in the atmosphere, the ratios of extinction coefficients of each color channel to the average extinction coefficient show roughly how much scattering in each color channel has taken place. This was done empirically in the paper and it was shown that the blue light scattering was about 60% stronger than the red light scattering. If Rayleigh scattering was dominant (i.e. atmosphere was dominated by small Rayleigh particles) then the blue light scattering would have been much stronger (about 300%) than the red light scattering. This shows that the atmosphere was not dominated by small particles, but rather larger aerosols which were causing the visible haze.

Other research papers have added onto the above algorithm by incorporating "blind" ways of estimating the airlight at infinity $A_\infty$. The paper called "Blind Haze Separation" [33] discovers that in the high spatial regions of the images, the direct transmission and airlight components are independent of each other. Wavelet transformation is performed to obtain these high spatial frequency regions and then Independent Component Analysis is used to separate the components from each other in the most optimized way possible. In this way the degree of polarization is computed. $A_\infty$ is then estimated using two scene points which have similar unknown radiance but lie at different depths.

The polarization-based dehazing research papers display very interesting results. These papers are not concerned with measuring the atmospheric particulate matter (PM). They are concerned with increasing scene visibility and only propose an extension to the research which can estimate aerosol size. The results display a marked increase in scene visibility. The problems of this paper lie in the fact that airlight is not fully polarized and therefore a complete dehazing of the scene is not possible. A process of depolarization takes place which affects the overall degree of polarization of the airlight. Depolarization is caused by the presence of aerosols, ground reflection and optical thickness [3]. It is our aim to utilize this polarization-based technique by including temporal knowledge gained over images taken under different haze conditions and other techniques that we will introduce shortly to more accurately estimate atmospheric optical thickness (which is correlated to PM10 levels). We will also propose methods to enhance the dehazing process, again for the purpose of laying down a foundation to measure PM10 from digital images.

## 3.4 Dichromatic Atmospheric Scattering Framework

The dichromatic atmospheric scattering framework is another research approach for dehazing digital images [31]. The framework allows the temporal analysis of images taken under different weather conditions (e.g. dense and mild fog and haze). It uses multiple images, each taken under different weather conditions. The framework produces a dehazed image and range map similar to the polarization-based approach described in the previous section. The framework rests on two assumptions. The first is that the airlight hue over the whole scene does not vary with changes in the weather. The second assumption is that the direct transmission of the same scene points under different weather conditions is constant. Only the intensities of direct transmission and airlight changes but the color remains the same. This framework represents the colors of scene points as vectors in a RGB 3D coordinate system. The vector directions define the color and the vector magnitudes define the intensity.

The color of the direct transmission and airlight are defined as unit vectors in a RGB 3D coordinate system (see figure 9). The magnitudes of the direct transmission and the airlight for each scene point vary along these unit vector directions. The color of a scene point is a linear combination of the direction of the direct transmission unit vector and the direction of the airlight unit vector. In this way the scene point color varies anywhere within the plane defined by the direct transmission and airlight unit vectors, as shown below. This plane is known as the dichromatic plane.

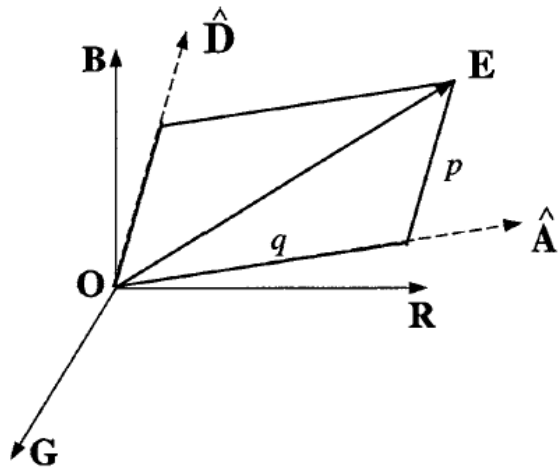

**Figure 9: The RGB coordinate system showing the airlight and direct transmission unit vectors. The scene point color is a linear combination of $p$ and $q$ along these unit vectors [31]**

The dichromatic framework states that the color of any scene point $E$ is equal to the summation of the two vectors with magnitude $p$ and $q$ in the RGB color space (see figure 9). According to this, $E$ is equal to the vector summation:

$$E = p\hat{D} + q\hat{A} \qquad 3.4.1$$

where $E$ is the color of a scene point and is composed of the addition of two vectors, each of which has a magnitude and direction. $p$ and $q$ are the magnitudes of the direct transmission and airlight, respectively. $\hat{D}$ and $\hat{A}$ are the color unit vectors which define the directions of direct transmission and airlight, respectively. $\hat{A}$ is globally constant under different weather conditions. $\hat{D}$ is constant over the same scene points under different weather conditions. $p$ and $q$ vary for each scene point under different weather conditions.

We will now decompose each component in equation 3.4.1 and explain how they are computed. Then we will put everything back together to show how this forms the dehazed image and range map.

The magnitudes $p$ and $q$ are further decomposed using two atmospheric scattering models: attenuation over distance due to scattering (Allard's law, see [31]) and airlight which increases exponentially with distance (derived in [31]).

$$p = \frac{A_\infty r e^{-\beta z}}{z^2} \qquad 3.4.2$$

$$q = A_\infty (1 - e^{-\beta z}) \qquad 3.4.3$$

where $\beta$ is the scattering coefficient and $z$ is the depth of the scene point. $A_\infty$ is a global parameter which defines the airlight radiance at infinite distance. This global parameter signifies the column of atmosphere which is composed of pure airlight. This model is relative to $A_\infty$ as it varies directly with illumination changes, such as changes in the intensity of illumination (sun, diffuse sky and ground-reflection) over time. In this research, $A_\infty$ is not estimated by sky regions of the image like the polarization-based approach of the previous section. This is because the sky regions are not trustworthy as they are prone to intensity saturation and color clipping. $A_\infty$ represents the scene point color of a pure airlight column just above the horizon at infinite distance. It is estimated by obtaining the ratio of direct transmissions for scene points under two weather conditions and transforming this into a line fitting problem. We will describe how this is done now.

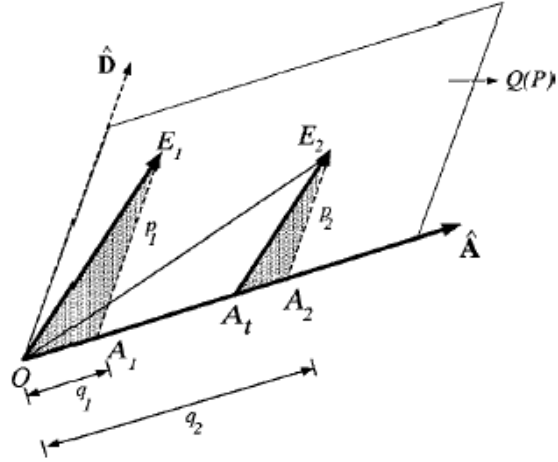

**Figure 10:** $E_1$ and $E_2$ are the colors of the same scene point under two different weather conditions. The shaded triangles are similar and therefore the direct transmission ratio $p_2/p_1$ is equal to the ratio $|E_2 A_t|/|E_1 O|$.

The ratios of direct transmission magnitudes for scene points with the same depth under two different weather conditions are equal. $p_1$ and $p_2$ shown in figure 10 are the direct transmission magnitudes of the same scene point under different weather conditions. Because the shaded triangles in figure 10 are similar, $p_2/p_1$ is equal to $|E_2 A_t|/|E_1 O|$ which can be computed directly from the pixel colors of the images. We will now use this ratio to compute $A_\infty$. Through deriving the direct transmission ratio in terms of $A_\infty$ and airlight magnitudes for both weather conditions we get a straight line equation of the form:

$$c = A_{\infty 2} - (p_2/p_1) A_{\infty 1} \qquad 3.4.4$$

where $c$ is equal to $|OA_t|$ in figure 10. Equation 3.4.4 is a straight line equation with two unknowns; $A_{\infty 1}$ and $A_{\infty 2}$ which are the airlight radiance at infinite distance of the two weather conditions. With several pairs of values of $c$ and the direct transmission ratio $p_2/p_1$ computed from different scene points at different depths, we calculate $A_{\infty 1}$ and $A_{\infty 2}$ by solving this line fitting problem. Here we have seen how the airlight at infinite depth global parameter for both weather conditions can be computed using non-sky scene points.

The direction of airlight hue $\hat{A}$ shown in equation 3.4.1 is computed by plane intersection. Each pair of scene point colors from two different weather conditions constructs a plane called the dichromatic plane. This plane can be constructed for all the scene points under two different conditions. All these different planes intersect at one line, which is $\hat{A}$ and is global over all pixels over both weather conditions.

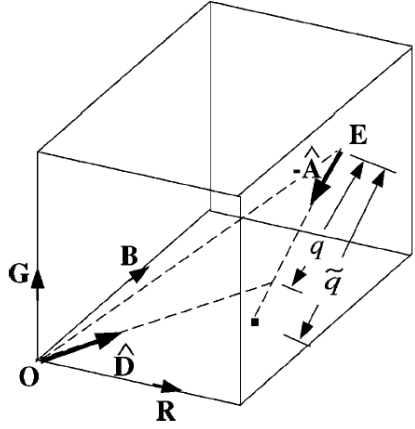

**Figure 11: The cube shown above is defined in the RGB coordinate system and contains all the color vectors of scene points under two different weather conditions.**

The cube shown above is constructed so that color vectors either lie on the surface of the cube or inside the cube. If the color vector lies on the surface of the cube, it has no direct transmission color to it and is fully composed of airlight. This color cube is used to compute the direction of direct transmission color $\hat{D}$ shown in equation 3.4.1. This cube defines the RGB color space within which all the color vectors of scene points lie. A cube with dimensions 10 was used in this research. An assumption is made that states that there exists at least one scene point whose direct transmission color vector lies on the surface of the cube. As explained before, each scene point color is defined by the addition of its vectors for direct transmission and airlight color (see figure 9) which all lie within this cube. As figure 11 illustrates scene point $E$ is defined by the direct transmission vector along $\hat{D}$ and the airlight vector of magnitude $q$ along $\hat{A}$. $\tilde{q}$ is the estimated airlight magnitude which is measured using geometry seen in figure 11. $\tilde{q}$ is the distance along negative $\hat{A}$ to the surface of the cube. For scene points with direct transmission color vector on the surface of the cube, $q$ and $\tilde{q}$ are equal. The rest of the scene points will have $\tilde{q}$ greater than $q$. For each scene point we measure $\tilde{q}$ geometrically from figure 11 and use this to estimate the optical thickness $\tilde{\beta_1} z_i$ using equation 3.4.3. The optical thickness is normalized to the scaled depth to produce $\tilde{\alpha}_i$ as follows:

$$\tilde{\alpha}_i = \frac{\tilde{\beta_1} z_i}{(\beta_2 - \beta_1) z_i} \qquad 3.4.5$$

The scene points with minimum $\tilde{\alpha}_i$ are the points where the estimated airlight magnitude $\tilde{q}$ and real airlight magnitude $q$ are equal. This is because these points have direct transmission colors which lie on the surface of the cube. Because of this, these points will have the minimum $\tilde{q}$ and therefore according to equation 3.4.3 minimum $\tilde{\beta_1} z_i$ and hence minimum $\tilde{\alpha}_i$. The real airlight magnitude of these scene points are then used to dehaze the rest of the scene.

In this research windows on buildings in the urban scenes represent scene points which are dominated by airlight and have no direct transmission color. These points have direct transmission color equal to black and therefore their direct transmission unit vector $\hat{D}$ lies on the cube at $O$. These points were automatically discovered using the above approach.

Since we have already computed $\hat{A}$, $\hat{D}$, $q$ and $A_\infty$ for the window points in the scene we can compute the optical thickness $\beta z$ of these points from equation 3.4.3. Based on these points, the rest of the scene can be dehazed using the relative depth equation 3.4.7. We will now see how this is done.

A scaled depth map of the scene can be computed from equation below which consists of already computed parameters.

$$(\beta_2 - \beta_1)z = \ln\left(\frac{A_{\infty 2}}{A_{\infty 1}}\right) - \ln\left(\frac{p_2}{p_1}\right) \qquad 3.4.6$$

Using the above equation we can compute the relative depth of any point with respect to another point by dividing their scaled depth. This would give us the depth ratio $z_2/z_1$ which is used to compute the relative optical thickness according to:

$$\beta z_j = \beta z_i (z_j/z_i) \qquad 3.4.7$$

where $\beta$ is the scattering coefficient for the scene, $z_i$ and $z_j$ are the depths of two scene points. Since we know the optical depth of the window scene points, we can use equation 3.4.7 to compute the relative optical depths of all other points. Now that we have the optical thickness $\beta z$ for the other scene points and have computed $A_\infty$ previously, we can use equation 3.4.3 to compute the airlight magnitudes of all the scene points.

The final dehazed color vector is the airlight color vector $q\hat{A}$ (magnitude and direction) subtracted from the color vector of the scene point $E$ as shown in equation 3.4.8. This equation forms the final dehazed scene.

$$p\hat{D} = E - q\hat{A} \qquad 3.4.8$$

The scaled depth map of the scene is computed by using equation 3.4.6. This map is created from two images taken under different weather conditions. The depth map is scaled to the difference in scattering coefficients $\beta$ of the two images. This difference in optical thickness is known as DOT. Another relative depth is recovered from equation 3.4.7 and is scaled to the global scattering coefficient of that image.

The downside of this framework is that it obtains knowledge over time by depending on variations in the atmospheric conditions in order to function. The dichromatic framework is restricted in its dehazing approach as it relies on the assumption that at least one scene point direct transmission is known (points representing windows on buildings in the scene). Therefore the dehazing is only as good as the choice of scene points with known direct transmission. The framework also relies on the assumption that the global illumination does not vary and therefore the hue of the airlight is constant. The framework also assumes overcast skies. We will use this temporal framework to our advantage by aiding it by the fact that airlight is partially polarized (according to [1]) to more accurately extract the haze from the scene and hence better estimate optical thickness. In our approach we will not exactly follow equation 3.4.2 for the magnitude of the direct transmission component of a scene point. This is described further in section 4.3.1.3.

# 3.5 Image Dehazing Using a Statistical Prior

Through the course of this research another dehazing algorithm has been developed and published [42]. This algorithm is based on statistics of haze-free outdoor images. It is based on a key observation that most local patches of haze-free outdoor images contain some pixels which have very low intensities in at least one color channel. These pixels are known as the dark channel which is a subliminal channel in captured images that have the lowest intensities of colors. The reason behind the existence of dark channels is due to the overwhelming presence of shadows, colorful objects/surfaces and dark objects/surfaces in outdoor images. The dark channel prior allows the estimation of haze thickness and hence when removed recovers a high quality haze-free image of the captured scene. A scaled depthmap is also produced as a by-product.

The formation of haze-filled images, like the previous approaches, is described by the following equation:

$$I = Re^{-\beta z} + A_\infty(1 - e^{-\beta z}) \qquad 4.5.1$$

where $I$ is the image radiance, $R$ is the scene radiance, $\beta$ is the scattering coefficient of the atmosphere, $z$ is the depth of the scene object relative to the camera and $A_\infty$ is the airlight radiance at infinite distance.

Given an input image, the dark channel algorithm analyzes patches of the image and computes the dark channel $R^{dark}$:

$$R^{dark}(x) = \min_{c \in \{r,g,b\}} \left( \min_{y \in \Omega(x)} (R^c(y)) \right) \qquad 4.5.2$$

where $R^c$ is a color channel of the scene radiance $R$ and $\Omega(x)$ is a local patch of the image centered at $x$. This dark channel takes the darkest of the three color channels over a patch of the image and assigns this to the pixel at the center.

The dark channel approaches zero in haze-free outdoor images as these images are filled with shadows, colorful surfaces and dark colored objects. The dark channel prior is statistically proven by analyzing a set of 5,000 haze-free outdoor images.

Due to the additive effect of airlight, a haze image is brighter than its haze-free version in parts of the image where the transmission $t$ is low. Because of this, the dark channel of the haze image will have higher intensity in regions with denser haze.

If we take the minimum of the haze model of equation 4.5.1 we form the following equation:

$$\min_{y \in \Omega(x)} (I^c(y)) = \tilde{t}(x) \min_{y \in \Omega(x)} (R^c(y)) + (1 - \tilde{t}(x)) A^c \qquad 4.5.3$$

where $\tilde{t}(x)$ is the transmission per patch and is considered constant over each patch. $A^c$ is the atmospheric airlight at infinite distance over each of the color channels.

If we divide all the terms of equation 4.5.3 by $A^c$ and take the minimum over the three colors, we form the following equation:

$$\min_c (\min_{y \in \Omega(x)} (\frac{I^c(y)}{A^c})) = \tilde{t}(x) \min_c (\min_{y \in \Omega(x)} (\frac{R^c(y)}{A^c})) + (1 - \tilde{t}(x)) \qquad 4.5.4$$

According to the dark channel prior, the first term on the right hand side of equation 4.5.4 tends to zero as the minimum color of the scene radiance over a patch of the image is very low in outdoor haze-free scenes. Therefore, the transmission of the patch $\tilde{t}(x)$ becomes:

$$\tilde{t}(x) = 1 - \min_c (\min_{y \in \Omega(x)} (\frac{I^c(y)}{A^c})) \qquad 4.5.5$$

$A^c$ is computed from the dark channel as the maximum pixel in the dark channel with the maximum imge irradiance value. This is quite different to taking the maximum pixels in the image irradiance values. This is because the sky may not be the brightest part of the image. A simple white car in the scene could appear brighter. Because of this the maximum pixels of the dark channel are taken which ensure that they are taken from areas of dense haze. The algorithm goes on to perform a Soft Matting algorithm to smoothen out the patches of the computed transmission map. Using the transmission map of the scene, they can recover the scene radiance $R$ which is essentially the dehazed image of the scene.

The transmission computed in equation 4.5.5 is equal to:

$$\tilde{t}(x) = e^{-\beta z}$$

where $\beta$ is the scattering coefficient of the atmosphere and $z$ is the depth of the scene. This allows the computation of a depthmap $\beta z$ of the scene scaled to the unknown scalar $\beta$.

Using the computed transmittance and estimated atmospheric airlight they then substitute these into equation 4.5.1 and recover the scene radiance.

This algorithm is remarkably simple but dehazes very effectively. It has been compared to other methods of dehazing and has shown to be uniquely easy to implement and effective in dehazing. It does have a documented flaw in dehazing surfaces in scenes which have similar color to airlight. It mistakes these surfaces as dense haze and hence distorts the recovered scene radiance and estimated depth.

From the perspective of analyzing atmospheric scattering, this algorithm may lack accuracy due to its founding assumption. In reality, the dark channel may not be absolute zero and may be above this by quite a bit. This error causes inaccuracy in measuring the atmospheric scattering coefficients and extracting knowledge about the atmosphere.

# Chapter 4 – Methodology

The first step in this research was a procedure known as radiometric camera calibration. This is done to linearize the captured images according to the light response curve of the imaging device. We also describe the image processing techniques that we incorporated in this thesis to be able to extract the knowledge efficiently from the images. We will then reconstruct the research undertaken in the image dehazing algorithms (sections 4.3-4.5) After achieving we describe our models of atmospheric scattering measurement and depth estimation.

# 4.1 Radiometric Camera Calibration

Radiometric camera calibration is a standard pre-processing step in computer vision. An image of a scene is a two-dimensional array of brightness values. Each pixel of an image consists of red, green and blue color components whose values range from 0-255 for 8-bit precision. The brightness values of each pixel are not a measure of the actual physical light radiance emitting from the scene being photographed. There is a nonlinear mapping unique to each image-capturing device that dictates the relationship between scene radiance and pixel values. Digital cameras introduce nonlinearities in order to mimic the response characteristics of film and to compose the high dynamic range of light prior to quantization. For this reason the relation between the actual physical light radiance coming from a scene is nonlinearly related to the brightness of the corresponding image pixel.

The graph that maps the pixel brightness values to the physical measure of light irradiance is known as the radiometric response curve. The following subsection describes the approach we took to calibrate our camera. We calibrated a Panasonic Lumix DMC-FZ8 and a Canon 20D DSLR camera. The latter camera is a semi-professional Single Lens Reflex (SLR) camera and will allow us to obtain more accurate results due to its larger imaging sensor with deeper bit-depth per pixel.

## 4.1.1 Obtaining the Response Curve

In order to calibrate the radiometric response of a camera we used the method of Debevec and Malik [4]. The following steps outline the process of obtaining the response curve:

1. Set the camera to full manual mode. Adjust the camera so that it is fixed in place and does not move.
2. Take pictures at equal increments or decrements of exposure. Shutter speed variation is the recommended approach over aperture change. Ideally the more pictures taken, the more accurate the response curves. Around five pictures are needed to produce a decent curve. Make sure the pictures are being captured in RAW format to prevent any extra nonlinear interference from the camera's image processing pipeline. Raw format images will hold the unprocessed data we need in order to get accurate measurements of light radiance. We took a sequence of images at varied shutter speed as shown below. Scenes with high ranges of light and dark regions produced smoother curves. Ideally a Macbeth chart should be used to calibrate the camera. The Macbeth chart is an industry standard used for calibration. It consists of a board with 24 color patches arranged on it. These color patches have known reflectance that enables it to be used for calibration. We did obtain a Macbeth chart to be able to perform this procedure but before we obtained this chart we approximated the procedure by printing our own Macbeth chart with the approximated equivalent color patches. We did this by printing out the approximate RGB colors of the Macbeth chart. The chart we created is shown below.

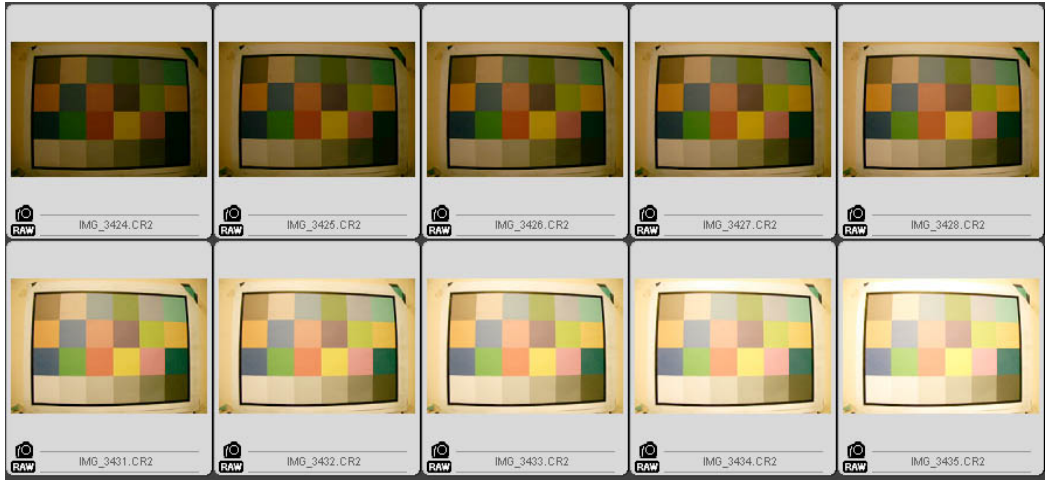

**Figure 12: The images of the Macbeth chart we created based on hexadecimal color representations of a real Macbeth chart. The chart is taken at different exposures by varying the shutter-speed**

3. Convert the RAW images to TIF uncompressed format to be able to process them in Matlab. TIF uncompressed format is compatible with Matlab and retains the per pixel information that we need in order to undertake our experiments. We used uncompressed TIFF images with 48 bits per pixel (16 bits for each color).
4. We loaded these images into Matlab and ran the Matlab script explained in [4].

The response curve looks like the following:

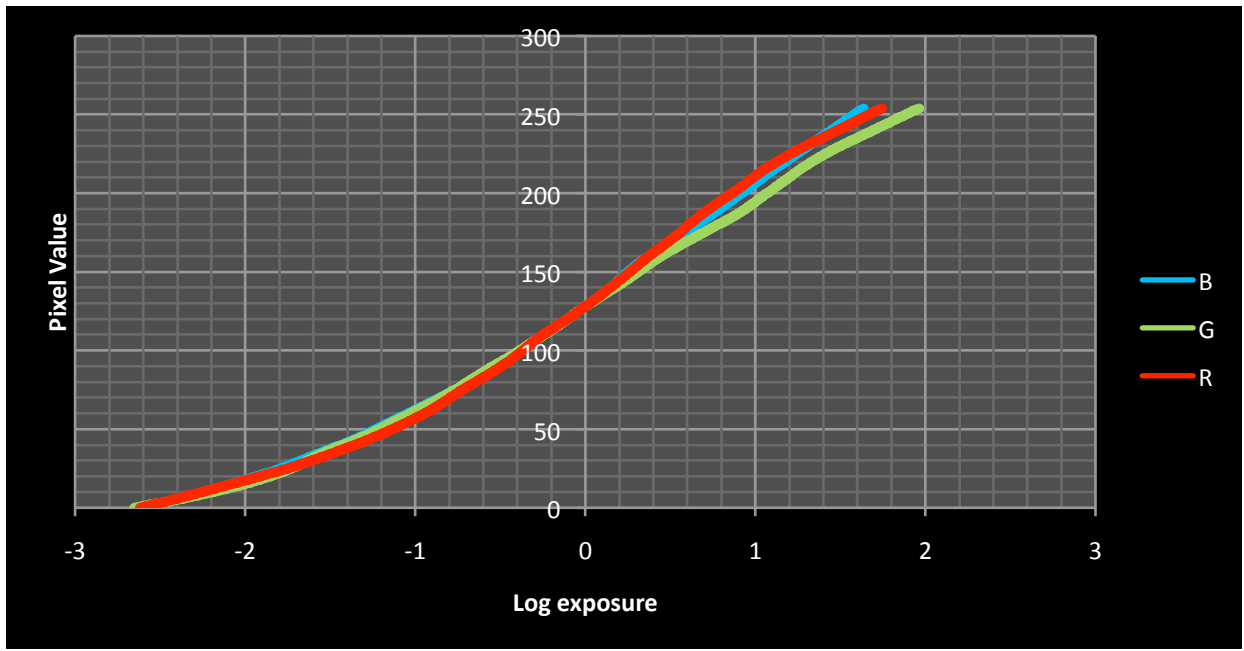

**Figure 13: Radiometric response curve measured for the Canon 20D DSLR**

This response curve is for a Canon 20D DSLR camera. The curve is composed of three different plots for each color (R, G and B). The X-axis is the scale for the natural logarithm of the exposure value which under the reciprocity equation is directly proportional to the irradiance [4]. The Y-axis shows the 0-255 scale of the image pixel brightness values.

In order to check the data in the response curve, we took sample pixels in each of the images used in the calibration and recorded their brightness values in each of the three colors. We plugged these brightness values into the response curve and obtained values for the exposures. From this data we plotted graphs showing the increase in exposure for each of the pixels over the images taken. The plotted graphs below (figure 14) show a linear increase in the brightness of one such pixel as the exposure increases. The graphs show that the response curve has indeed linearized the relationship between the irradiance and the pixel brightness value measured from the image.

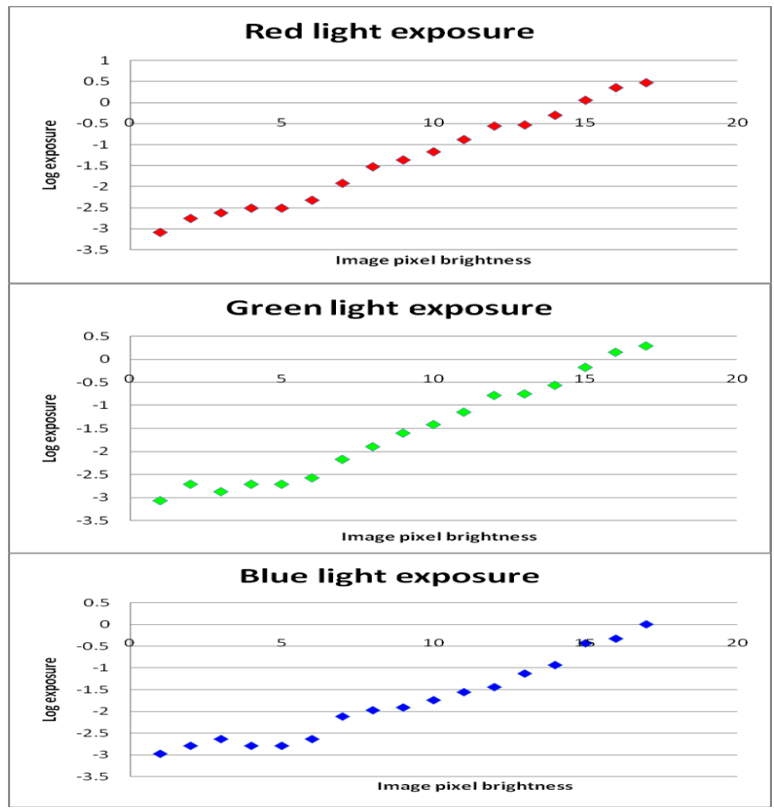

**Figure 14: The linear relationship between the log exposure and the image pixel values based on the radiometric response curve. This is plotted for one pixel in the sequence of images.**

## 4.1.2 Obtaining the Radiance

In order to retrieve the irradiance measurements from the exposure values; we divide the exposure by the exposure time ($\delta t$) according to the following equation as according to [4]:

$$\ln E = z + \ln \delta t$$

$\delta t$ is the shutter speed of the image, $E$ is the scene point irradiance and $z$ is the value of the radiometric response curve as a function applied to the image pixel brightness (i.e. the value of the image pixel brightness on the radiometric response curve). From this equation we are able to get an accurate measurement of the irradiance at each scene point in our scene.

## 4.2 Image Registration

In computer vision, sets of data acquired by sampling the same scene at different times, or from different perspectives, will be in different coordinate systems. Image registration is the process of transforming the different sets of data into one coordinate system. Registration is necessary in order to be able to compare or integrate the data obtained from different measurements [41].

This research is based on capturing images of scenery over time and comparing them based on the assumption that the depth of the scene objects is constant. The scene is kept constant to measure the other variations in radiance due to haze. Even though the camera is positioned at the same location and orientation before capturing every image there is still a minute error as the camera is repositioned manually. This causes the discrepancy in coordinate system between the images. This is a problem and requires image registration to align all the images together into one single coordinate system. As this research is based on pixel-wise computations this problem is a significant one and is addressed as follows.

In order to solve this problem we register the images by performing image transformation according to selected points that correspond to the same features in the images. In Figure 15, you can see a pair of points corresponding to the same feature (the corner of a building in the scene). We chose affine transformation to do this as this type of transformation includes translation and rotation. It also includes scaling and shearing which is not needed in registering our images as the only expected difference between the images of the same scene (resulting from movement of the camera) is rotation and movement in the horizontal and vertical directions. Affine transformation requires 3 pairs of selected points in order to align a pair of images.

We used the image processing toolbox of MATLAB. We use a command called *cpselect* and pass it two grayscale versions of the captured images. The *cpselect* command opens up a graphical user-interface (shown in Figure 15) which allows you to select pairs of points to use to align the images.

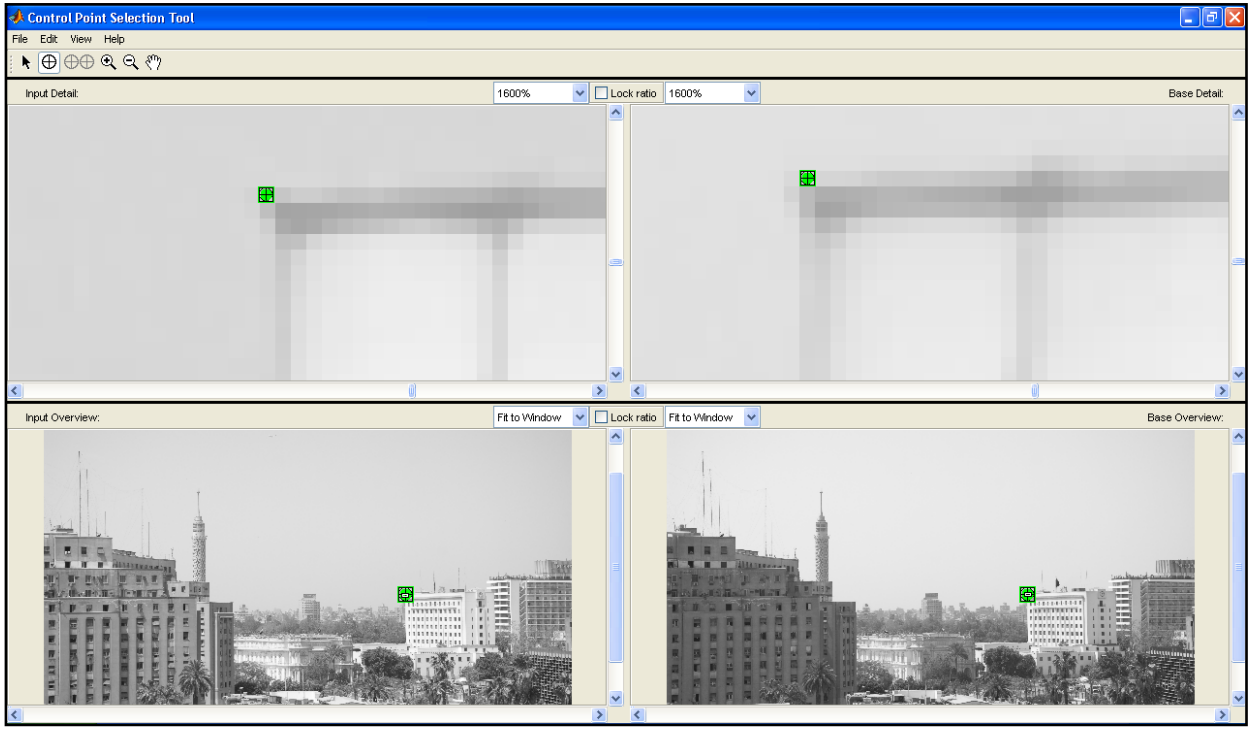

**Figure 15: MATLAB's *cpselect* user interface for selecting control points in order to register multiple images**

Once the pairs of control points are selected we perform affine image transformation. We perform it on one of the images to rotate it and translate it in order to align with the other image. The transformation describes what happens to the perceived positions of observed objects when the point of view of the observer (camera) changes. Refer to appendix for the MATLAB script to do this.

This process of image registration can be automated by using common algorithms such as the Lucas-Kanade method [44] or the Irani and Anandan algorithm [45].

# 4.3 Polarization-based Image Dehazing

In order to extract the airlight (refer to background section 2.3) we follow two approaches. The first is the approach undertaken by the paper: "Polarization-based Vision through Haze" [1]. We used the dehazing algorithm described in this paper to separate out the airlight in the scene. The second approach is the dichromatic framework which also separates out the airlight but depends on images taken under different atmospheric conditions (e.g. dense haze, mild haze, clear day).

We will describe the procedure we have undertaken to reconstruct the polarization-based dehazing algorithm and display results of this approach.

# 4.3.1 Dehazing Algorithm

Two images are taken of a scene of an urban horizon with haze limiting visibility. The two images taken are captured with different orientations of a polarizer filter. In our case we took the pictures from an elevated area in the center of Cairo. The sun was situated approximately perpendicular (90 degrees) to the viewing direction. One of the images was taken with the linear polarizer filter positioned parallel to the viewing direction in order to limit the polarized light reflecting off the air particles in our scene. The other image was taken with the polarizer filter positioned normal to the viewing direction to allow the partially polarized airlight to pass through the lens. The airlight is the dominating polarized light detected by the camera. Scattered light is always partially or fully polarized perpendicular to the plane of incidence. In our case the plane of incidence is defined by the plane between the illumination source (sun), the scattering particle and the camera. As the sun is roughly perpendicular to our viewing direction, the airlight is polarized in the vertical axis (i.e. the waves are oscillating up and down). When we rotate the polarizer filter to be perpendicular to the plane of incidence (in this case vertically), the polarized airlight is allowed through the lens. When we rotate the polarizer filter parallel to the plane of incidence (in this case horizontally), the polarized airlight is filtered out of the lens.

The total irradiance captured by the camera is the sum of the direct transmission and airlight (refer to background section 2.3 for a description of this). The equation for the total irradiance is:

$$I = D + A \qquad \qquad 4.4.1.1$$

where is $I$ the total irradiance, $D$ is the direct transmission and $A$ is the airlight. These are all per pixel quantities.

We assume that the light reaching the camera is dominated by the polarized airlight and therefore the direct transmission is minimally polarized [1]. Therefore the variation due to the polarizer rotation is from the airlight in the scene.

**The Dehazing Algorithm:**

This algorithm refers to the work described in section 3.3. Two images are taken at approximately parallel and perpendicular orientations of the polarizer filter provide us with two maps of scene point irradiance values. We will call the image taken with the parallel polarizer, the best polarization image and the perpendicular polarizer, the worst polarization image.

We denote the best polarization irradiance by $I_\perp$ and the worst polarization irradiance $I_\parallel$. The total irradiance $I_{total}$ of best and worst polarization states is:

$$I_{total} = I_\perp + I_\parallel \qquad \qquad 4.4.1.2$$

The first step is to estimate two global parameters: the airlight at distance tending to infinity $A_\infty$ shown in equation 4.4.1.4 below and the degree of polarization of airlight $P$ shown in 4.4.1.3 below. In order to estimate we extract a patch of sky low on the horizon almost totally obscured by haze. This ensures that the irradiance emitted from the region is totally dominated by airlight. We need this patch of sky dominated by airlight because it is the only part of the image that does not contain background object radiance. This means that it is fully composed of airlight and can therefore tell us the degree of polarization of the airlight without any interference from background object radiance. We therefore use this region to calculate the degree of polarization of airlight $P$. This is defined as to what extent the airlight is polarized. If $P$ is equal to 1, then it is fully polarized. Here we divide the difference between the worst polarized sky irradiance and the best polarized sky irradiance by the total sky irradiance. Again we average over the whole patch of the low sky region. This estimates the extent of polarization of the airlight in the image. $P$ and $A_\infty$ are defined as follows:

$$p = \frac{\Delta I(sky)}{I_{total}(sky)} \qquad 4.4.1.3$$

$$A_\infty = I_{total}(sky) \qquad 4.4.1.4$$

where $\Delta I(sky)$ and $I_{total}(sky)$ are the difference and summation of the perpendicular and parallel image irradiance components, respectively. Based on an image formation calculations the value of the airlight at any pixel $A$ is defined as:

$$A = (I_\perp - I_\parallel)/p \qquad 4.4.1.5$$

From the calculations above we can now estimate the direct transmission $D$.

$$D = I_{total} - A \qquad 4.4.1.6$$

This produces an image which is composed of direct transmission alone. What remains is approximately the direct transmission radiance from the objects in our scene. In order to take away the degradation of light due to attenuation over distance we calculate what is known as the transmittance $t$. This defines how much the light is transmitted through the atmosphere.

$$t = 1 - A/A_\infty \qquad 4.4.1.7$$

From this, the radiance of any object in the scene equals:

$$L_{object} = \frac{D}{t} \qquad 4.4.1.8$$

This removes the effect of airlight from the scene. By this we have regained the radiance of the objects without airlight and compensated for attenuation due to distance.

## 4.3.2 Results

The dehazing algorithm results can be viewed in the figure below. Despite the sky region of the dehazed image, there is an increase in contrast and restoration of color especially for distant objects. The best and worst polarized images are very similar. This shows that optics alone in this case is unable to dehaze the scene. The dehazed image on the other hand shows a marked increase in visibility as the layer of atmospheric scattering has been removed. The green hue of the vegetation has been clearly restored. The dehazed image also has increased contrast. The dehazed image shows a dark sky because the skylight which is caused by atmospheric scattering is removed leaving the dark deep-space sky. This is consistent with [1].

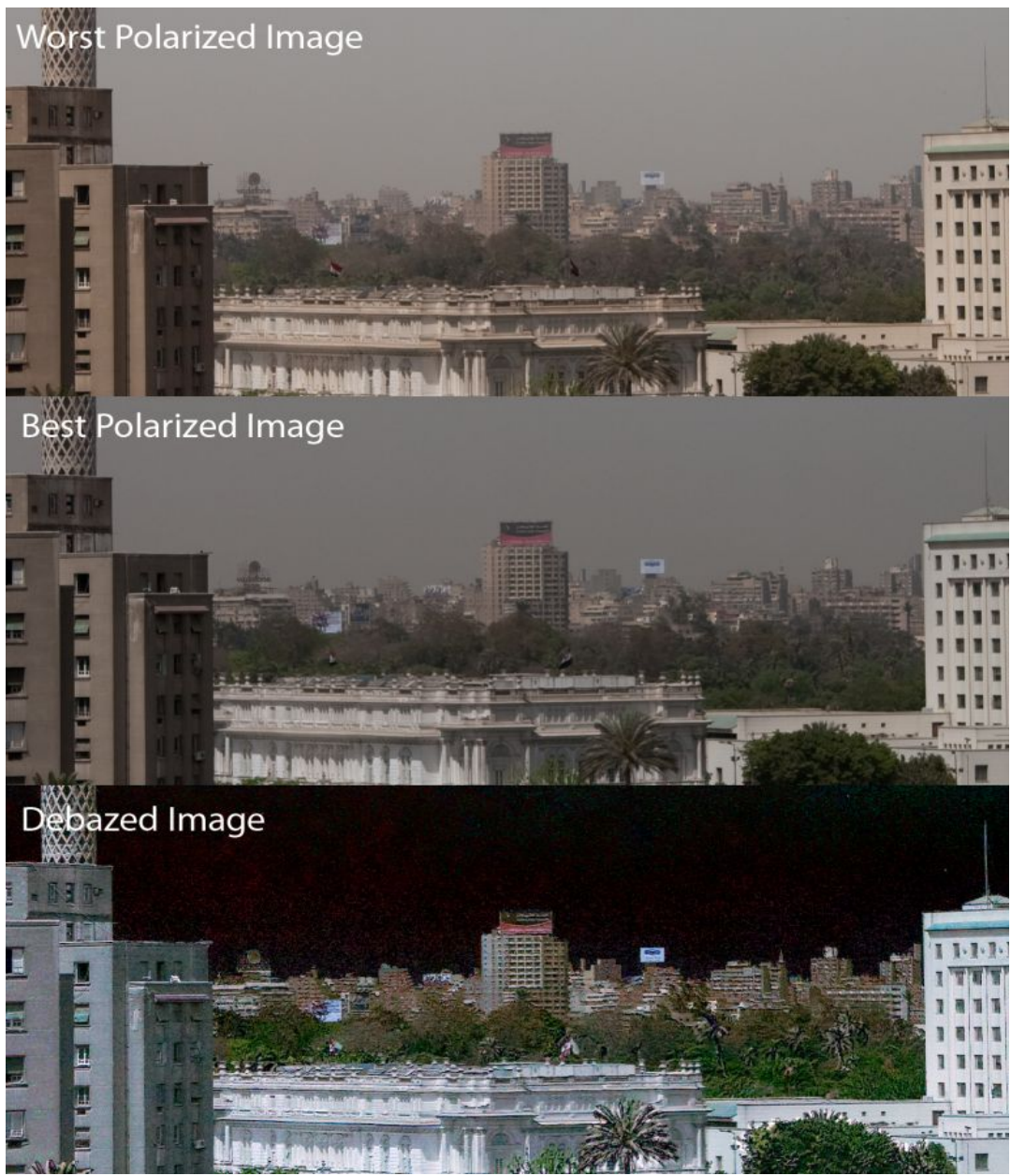

**Figure 16: The worst polarized image, best polarized image and the dehazed image produced by the polarization-based dehazing algorithm.**

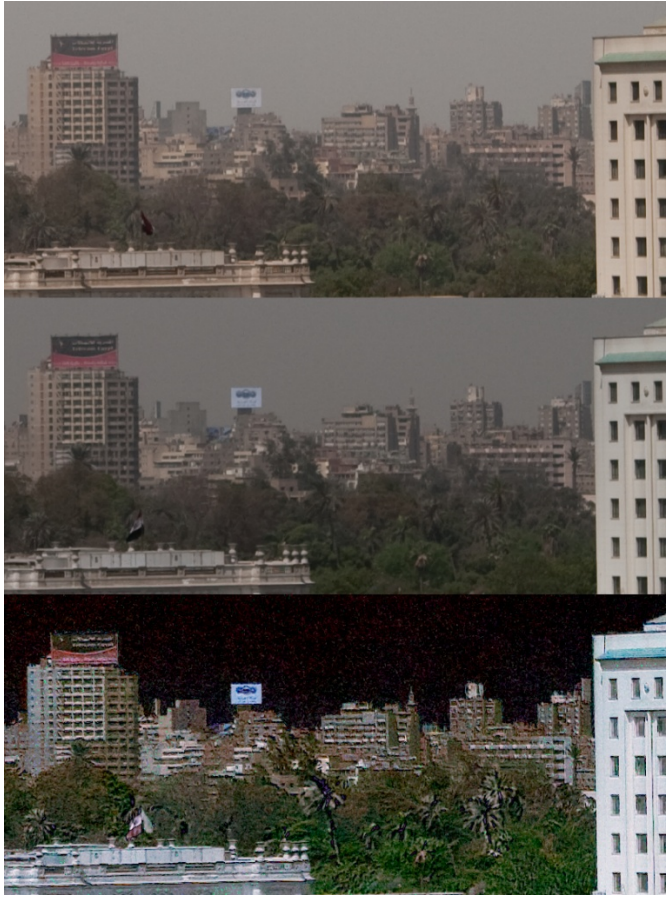

**Figure 17: This diagram shows a zoomed in section of the images in the previous figure. The first image is the worst polarization image, the second is the best polarization image and the third is the dehazed image.**

A byproduct of the dehazing algorithm is a range map. A range map is a grayscale image which shows the depth of the scene in shades of gray. The darker black regions in the foreground signify close objects. Objects further away get lighter and lighter. Figure 18 shows the range map for the scenes above. There are of course some exceptions in the grayscale depth map which are objects directly lit by sunlight which causes them to reflect highly polarized light (this can be seen by the light patches in the center bottom of the image in figure 18 & 19). Because of this, the algorithm mistakes them for far away objects dominated by polarized airlight. This is consistent with the dehazing algorithm [1].

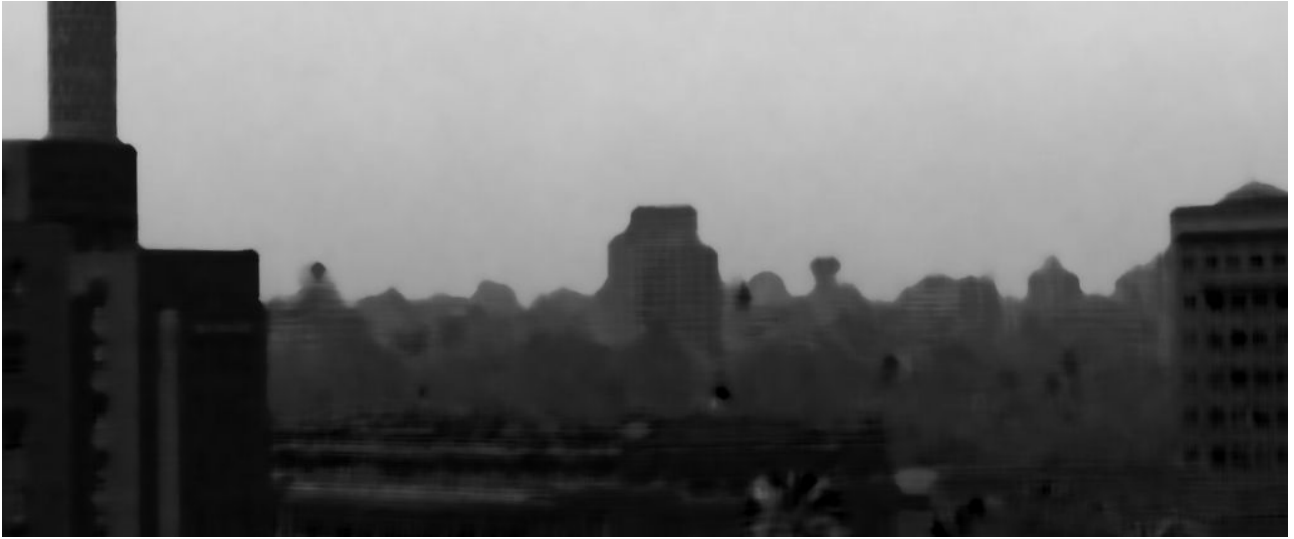

**Figure 18: The range map byproduct of the dehazing process shows the relative depth of objects in the scene in the form of a grayscale image (lighter shaded objects have more depth). This image shows the same scene as in figure 16.**

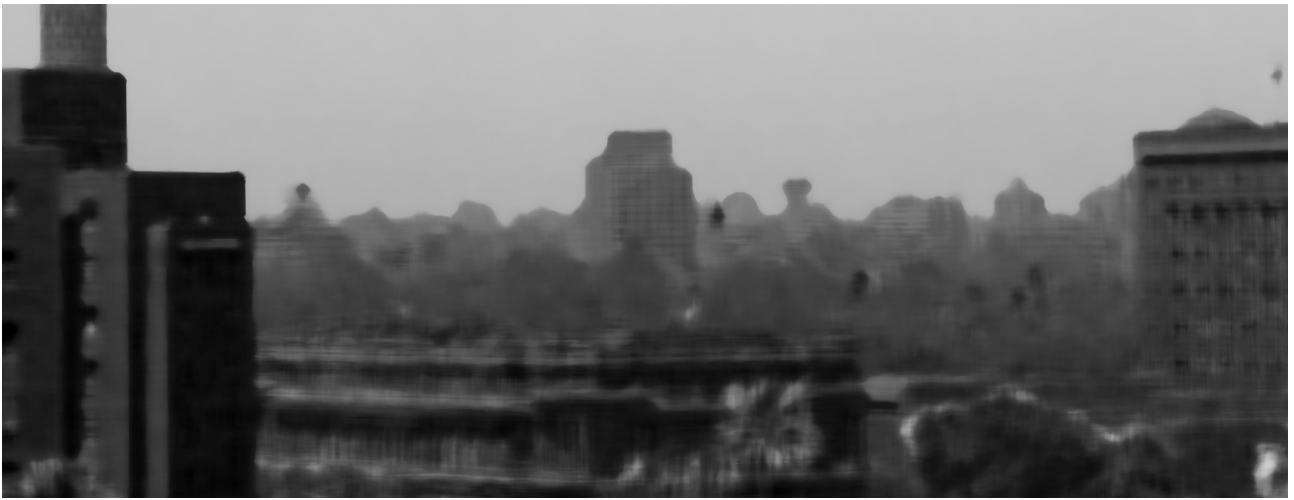

**Figure 19: Similar to the previous figure, the haze extracted from the scene can be displayed as a grayscale map.**

In the figure above, we can see the increase in brightness of objects as their depth increase. This figure and the previous figure are correlated to each other as an increase in depth means an increase in haze radiance. Some exceptions can be seen due to scene objects contributing to the polarization of the light reaching the camera. Despite this, the background of the above figure clearly displays the increase in the brightness of the shade of gray as the depth increases rapidly.

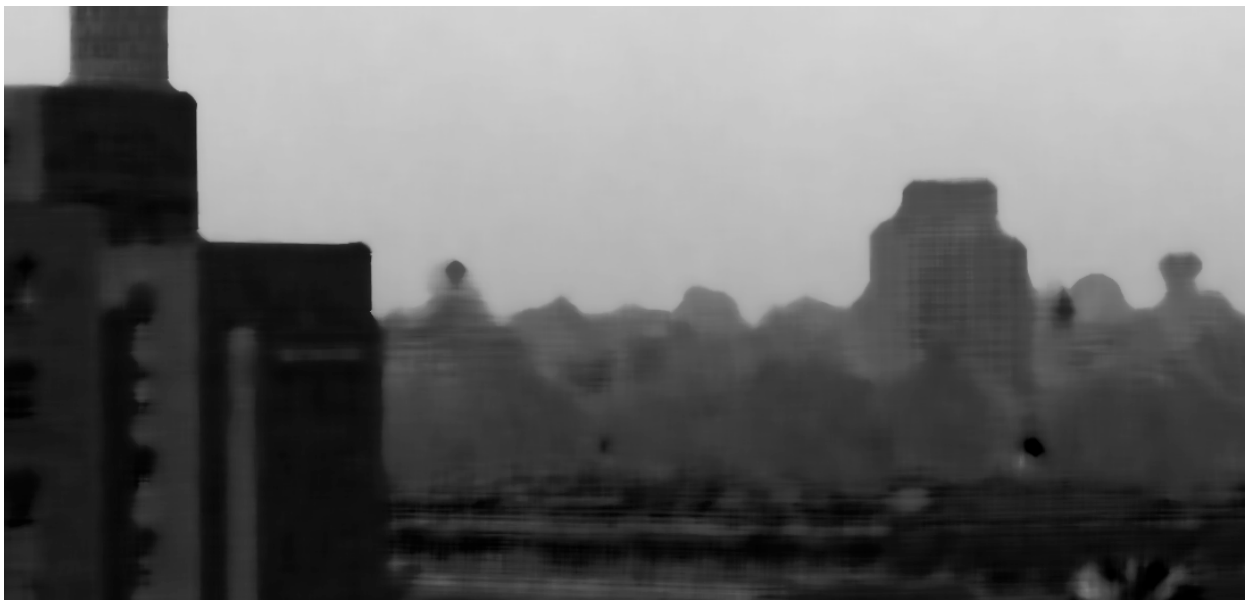

**Figure 20: A cropped and enlarged region of the depth map produced by the algorithm. This figure shows the increase in brightness with depth.**

# 4.4 Temporal Image Dehazing

The temporal image dehazing is based on the dichromatic atmospheric scattering framework described in section 3.4. The framework takes two images taken under different haze conditions and produces a dehazed scene and scaled depth map of the scene. The steps of the algorithm are described in detail in section 3.4.

We reconstructed this research and tested the algorithm on synthetic scenes. The reason why we tested this on synthetic scenes and not real scenes lies in the impracticality of this temporal dehazing algorithm. This algorithm relies on considerable fluctuations in the haze conditions of the images in order to function properly. In addition to this, this algorithm is designed for dehazing scenes with overcast illumination (i.e. cloudy skies). The images we captured did not display considerable variations in haze and were taken under sunny skies. As a result of this, we were unable to apply this approach to real images. Another reason that makes this approach less accurate than other dehazing algorithms in the context of measuring atmospheric scattering is that it assumes that the color of airlight does not change. This is a false assumption as the color of airlight does change with variations in the optical thickness and concentration of atmospheric pollutants

A scene consisting of 16 color patches was created (see Figure 21: The leftmost image shows the scene in moderate haze conditions. The middle image shows the scene in dense haze conditions. The rightmost image shows the dehazed scene. Each colored patch in the scene has a different relative depth value.). Each color patch was assigned a different relative depth value. The two atmospheric models of the dichromatic framework (see section 3.4) were applied to the scenes under two different haze conditions. These two models add airlight radiance and account for radiance attenuation due to scattering. The haze conditions were simulated by varying the scattering coefficient $\beta$ and the radiance at the horizon (at infinite distance) $A_\infty$.

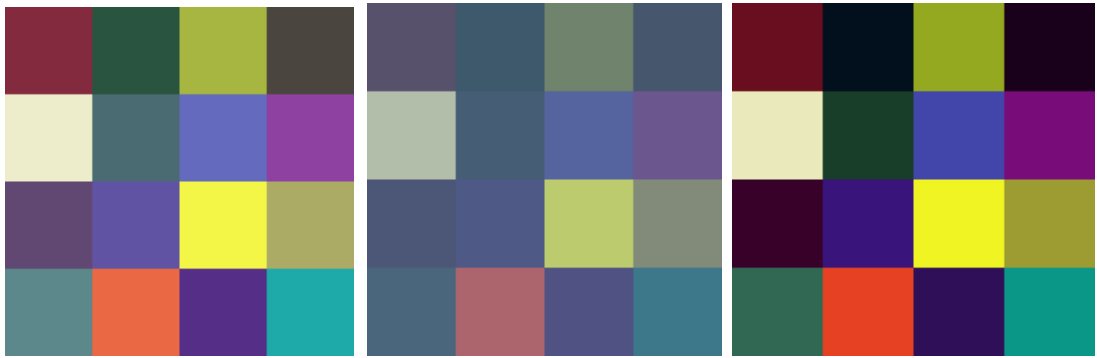

Figure 21: The leftmost image shows the scene in moderate haze conditions. The middle image shows the scene in dense haze conditions. The rightmost image shows the dehazed scene. Each colored patch in the scene has a different relative depth value.

The temporal image dehazing algorithm of the dichromatic framework was applied to these synthetic simulated scenes. Each colored patch in the scene shown above has a different relative depth value. You can see that the temporal dehazing algorithm has restored the color and contrast to most of the scene, shown in the rightmost image. The leftmost image shows the scene under moderate haze and the middle image shows the scene under simulated high levels of haze.

The temporal image dehazing algorithm first computes a plane for each scene point in the images. The colors of each scene point changes under the two simulated haze conditions. The color vectors created by the two colors of each scene point lie in the same plane (refer to Figure 9 and Figure 10). This plane is called the dichromatic plane and is computed for each scene point. In actual fact, the normal of this plane is computed by the cross product of the two color vectors, as follows:

$$N_i = E_1^i \times E_2^i \qquad 4.5.1$$

where $N_i$ is the normal of the dichromatic plane, $E_1^i$ and $E_2^i$ are the color vectors of the scene point under the two simulated haze conditions. The airlight color unit vector is given by computing the cross product of two different dichromatic plane normal vectors, as follows:

$$\hat{A} = \frac{N_i \times N_j}{||N_i \times N_j||} \qquad 4.5.2$$

where $N_i$ and $N_j$ are two different dichromatic plane normal vectors. A robust airlight unit vector can be computed by the intersections of several dichromatic planes. For a robust intersection, we minimized the following objective function:

$$\epsilon = \sum_i (N_i \cdot \hat{A}) \qquad 4.5.3$$

This is a linear least-squares minimization problem. We minimize equation 4.5.3 by expanding the dot product according to the red, green and blue colors of the normal vector $N_i$ and the airlight vector $\hat{A}$. Then we partially differentiate with respect to the red, green and blue airlight colors. You can see this below:

$$\frac{\partial \epsilon}{\partial A_r} = 2 \sum_i (N_{ir} A_r + N_{ig} A_g + N_{ib} A_b) N_{ir} \qquad 4.5.4$$

$$\frac{\partial \epsilon}{\partial A_g} = 2 \sum_i (N_{ir} A_r + N_{ig} A_g + N_{ib} A_b) N_{ig} \qquad 4.5.5$$

$$\frac{\partial \epsilon}{\partial A_b} = 2 \sum_i (N_{ir} A_r + N_{ig} A_g + N_{ib} A_b) N_{ib} \qquad 4.5.6$$

The three linear equations above are assigned to zero as we are looking for the minimum value of $\epsilon$ where the rate of change (gradient) is zero. The three linear equations above all equal zero and can be written in the form of matrix multiplication, as follows:

$$Na = 0 \qquad 4.5.7$$

where $N$ is a 3x3 diagonal matrix with values taken from coefficients of the airlight color components in the three equations above. $a$ is the matrix formed by the three color components of the airlight vector. Equation 4.5.7 is known as the homogeneous least-squares problem. This problem can simply be solved by assigning $a$ to zero. We are not looking for this trivial solution. We are looking for the value of $a$ which provides a non-trivial (non-zero) solution to the problem. This means that $a$ is a vector constrained to a unit length of 1 (so that it is a non-zero vector) and satisfies equation 4.5.7. This homogeneous least-squares problem can be solved by using either Eigen Decomposition [37] and/or Singular Value Decomposition [38]. As a result, we were able to compute the best intersection vector of all the dichromatic planes and therefore the best airlight color unit vector of the simulated scenes. We verified this by comparing the estimated airlight unit vector with the actual airlight unit vector we used to simulate the scenes. They were approximately equal.

Another problem we faced in applying the dichromatic framework to the synthetic simulated haze scenes was computing the value of $A_t$ shown in Figure 10. In order to find $A_t$, we minimized equation 4.5.8. $A_t$ is equal to the airlight vector multiplied by a scalar $t$. The essence of the equation below is that we are trying to find the value of the scalar $t$ which makes the vectors $E_2 - A_t$ and $E_1$ parallel. The reason behind this is that if we add a certain magnitude of airlight to $E_2$, it will become parallel to $E_1$. This is because $E_1$ and $E_2$ are two vectors each of which is equal to a linear addition of a direct transmission component and an airlight component. They lie in the dichromatic plane created by $\hat{D}$ and $\hat{A}$ (refer to figure 10). Therefore adding a certain amount of airlight $t$ to $E_2$ will shift it along the $\hat{A}$ vector and will therefore make it parallel to $E_1$. The cross product of any two parallel vectors is zero and so we derived the following equation to solve for $t$:

$$\epsilon = [(E_2 - A_t) \times E_1]^2 \qquad 4.5.8$$

We expanded equation 4.5.8 into a quadratic equation in $t$:

$$(E_2 \times E_1)(E_2 \times E_1) - 2(A \times tE_1)(E_2 \times E_1)t + (A \times E_1)(A \times E_1)t^2 \qquad 4.5.9$$

We differentiated equation 4.5.9 with respect to $t$ and assigned this to zero as to find the minimum value of $t$. This gave us the following equation:

$$t = \frac{(A \times E_1) \cdot (E_2 \times E_1)}{(A \times E_1) \cdot (A \times E_1)}$$

which allows us to find the value of $t$.

Once the airlight color unit vector and $t$ are computed, we follow the dichromatic framework approach to estimating $A_{\infty 1}$ and $A_{\infty 2}$ which are the airlight radiance at the horizon for both simulated haze scenes. These are estimated by solving a line-fitting problem as described in section 3.4. The rest of the dichromatic framework is straightforward and was applied as is (refer to section 3.4). The results of running the algorithm on the simulated scenes can be seen in Figure 21. The dehazing algorithm greatly restores the color and contrast of the scene.

The estimated relative scaled depths were computed using equation 3.4.6 of the dichromatic framework. The computed scaled depths were compared with the actual scaled depths used in simulating the synthetic haze scenes. This comparison can be seen in Figure 22.

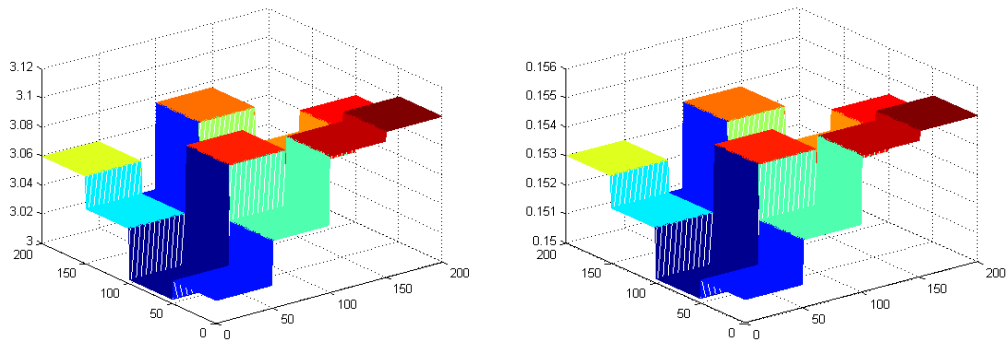

**Figure 22: The depth maps for the actual relative depth (left) and the estimated relative depth (right). As you can see the relative depths are estimated correctly.**

As you can see above, the actual relative depth of the synthetic scene is identical to the estimated relative depth recovered by the temporal dehazing algorithm. This proves that the algorithm is functioning correctly and not only able to dehaze the scene, as shown before, but also able to recover an accurate relative depthmap of the synthetic scene.

# 4.5 Dark Channel Dehazing

Through the course of this research another algorithm of dehazing has been developed and published [42]. This algorithm is based on statistics of haze-free outdoor images. It is based on a key observation that most local patches of haze-free outdoor images contain some pixels which have very low intensities in at least one color channel. This is due to the overwhelming presence of shadows, colorful objects/surfaces and dark objects/surfaces in outdoor images. They call this prior, the dark channel prior and use it with the haze imaging model described in the polarization-based dehazing algorithm and the temporal dehazing algorithm described in the previous sections. This dark channel prior enables them to estimate the thickness of haze and recover a high quality haze-free image of a scene. They also produce a by-product scaled depthmap.

The simulation of haze formation in images is described by equation 3.3.4. We simulated a sequence of images of a scene (shown in Figure 23) under varying scattering coefficient. The depths of the scene vary spatially but are constant over time. Refer to section 6.1 for details on simulating haze.

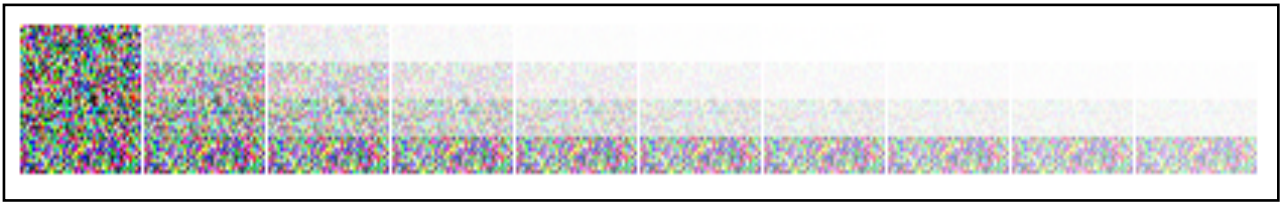

**Figure 23: Simulated scenes with increasing scattering coefficients. As a result of this, the simulated haze increases and decreases the contrast of the scene radiance, eventually saturating all colors in airlight.**

The dark channel approaches zero in haze-free outdoor images as these images are filled with shadows, colorful surfaces and dark colored objects. The dark channel prior has been statistically proven by analyzing a set of 5,000 haze-free outdoor images [42].

Due to the additive airlight, a haze image is brighter than its haze-free version in where the transmission $t$ is low. Because of this, the dark channel of the haze image will have higher intensity in regions with denser haze.

We can see the resultant dehazed image of a scene and the depthmap produced using this dehazing algorithm, in Figure 24 and Figure 25.

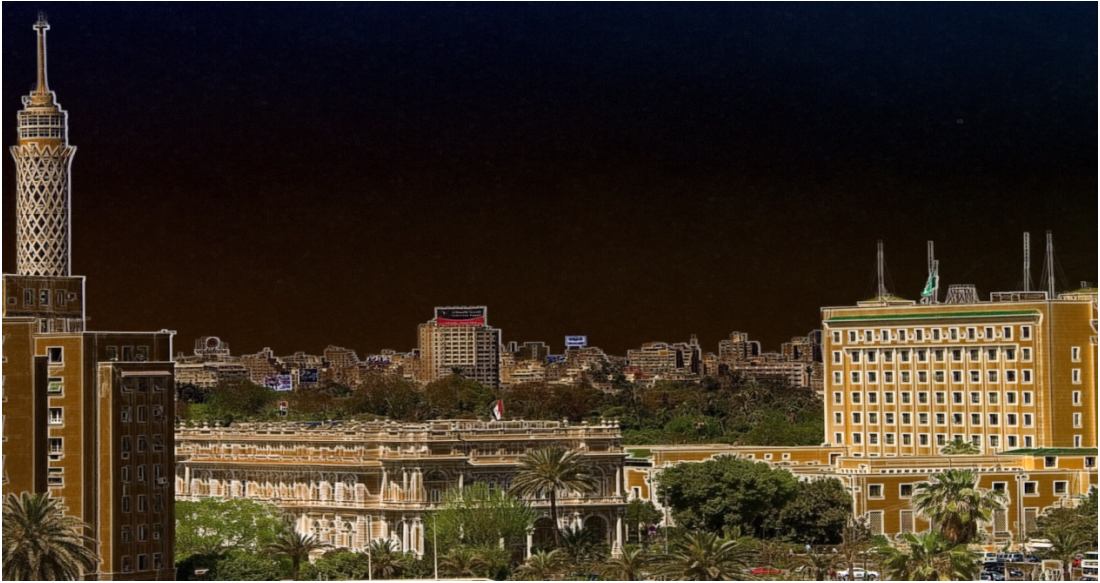

**Figure 24: A dehazed image of a scene using the dark channel dehazing algorithm. Some of the colors are distorted due to their colors being close to the colors of airlight.**

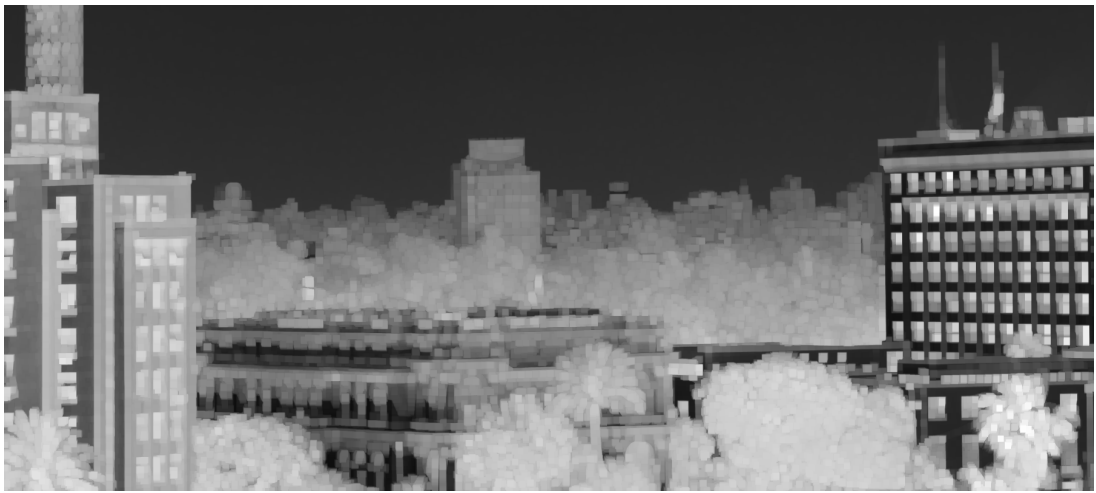

**Figure 25: The depthmap of the scene produced by the dark channel dehazing algorithm. The Darker the objects, the further away they are. Some surfaces are confused to be further away as the color of these surfaces is very close to the color of atmospheric airlight.**

The results of this algorithm are remarkable considering the simplicity of its model. As you can see form the figures above the scenes are dehazed effectively and the depthmap is accurate, despite being some distortions in the recovered colors and depths due to an error in the prior. The distortions take place because some surfaces have colors very close to the color of atmospheric airlight. Because of this, the algorithm mistakes these surfaces to be regions of dense haze. It can be seen in the depthmap in the figure above, as the dark surface of the building to the right of the image. The surface of the building has been mistaken for a far densely haze filled region, hence showing up as almost black similar to the sky region. The airlight composing the sky region has completely been removed leaving the deep space appearance similar to the polarization-based dehazing algorithm. The blockiness effect we see in the depthmap in the above figure is due to processing the dark channel algorithm per patch of the image. In this experiment we used patches of dimensions 10x10 pixels, which leaves this blockiness effect.

## 4.6 Comparison of Dehazing Algorithms

Once the three known dehazing algorithm described in this chapter were reconstructed and tested on simulations and real life images we compared each these algorithms along with the algorithms described in the next chapter, from the perspective of measuring atmospheric scattering accurately and recovering depth maps.

Each of the three known dehazing algorithms (polarization-based, temporal and dark channel) perform haze removal on images of hazy scenes and produce as byproducts depthmaps of the scene scaled to an unknown scalar, known as the atmospheric scattering coefficient and denoted by $\beta$. Our goal in this research is to measure $\beta$. This allows us to quantify atmospheric scattering which can be strongly correlated to PM levels and recover a more accurate depthmap of the scene as we remove the scale factor of the scattering coefficient from the depthmap. With this goal in mind we compared each dehazing algorithm extensively along with the algorithms introduced in the next chapter which are the contributions of this thesis. The results of the comparison are displayed in chapter 6. The next chapter describes the algorithms presented in this thesis for atmospheric scattering and a more accurate depthmap recovery.

# Chapter 5 – Measuring Atmospheric Scattering and Unscaled Depthmap

The transmittance of a medium is a measure of how much the medium transfers light through it. A fully opaque medium has a transmittance equal to 0 and a fully transparent medium has a transmittance of 1. In order to recover the transmittance of a scene and measure the atmospheric scattering coefficient, denoted by $\beta$ we implemented two optimization algorithms.

The first optimization is a standalone optimization as it is not based on any previous dehazing algorithm. We refer to it as the *Color Optimization*. It takes a sequence of images captured over time and solves for the set of transmittances and scene radiance pixels that best fit the model of haze formation in images. We will refer to this model of haze formation as the *haze image formation model*. This model is used in the dehazing algorithms and can be seen in equation 3.3.4. We will reintroduce it in the following section.

The second optimization algorithm is an extension of the dehazing algorithms. We will refer to this optimization as the *Constant Depth Constraint* (CDC). This optimization decomposes the scaled depthmaps produced by each of the three dehazing algorithms (polarization-based, temporal and dark channel) into a single depthmap and set of scattering coefficients that best fits the sequence of images according to a transmittance constraint.

Both these optimizations are performed using nonlinear optimization. We will describe them in the following sections.

# 5.1 Color Optimization

As part of the goal of this thesis to analyze the atmospheric scattering from digital images in order to more accurately measure particulate matter, we designed and implemented the Color Optimization algorithm. This algorithm is based on a simple ratio derived from the haze image formation model which accounts for the addition of airlight radiance as distances increase and attenuation of scene radiance due to scattering. This model is described by the following equation. We use a sequence of multiple images of the same scene captured over time and this model to find two unknowns: the scene radiance $R$ and the scaled depth $\beta z$. In other words we are trying to optimize to find the scene radiance and scaled depth of the scene that best fit the haze image formation model applied to a sequence of images of the same scene captured over time. This is an optimization problem that applies this model in a temporal fashion to find the two unknowns.

$$I = Re^{-\beta z} + A_\infty(1 - e^{-\beta z}) \qquad 5.1.1.1$$

where $I$ is the image irradiance per pixel, $R$ is the scene radiance, $\beta$ is the scattering coefficient of the atmosphere and $z$ is the depth of the scene object relative to the camera. $A_\infty$ is the atmospheric airlight at infinite distance which we can obtain from the sky region similar to the method the polarization-based dehazing uses. We rewrite the above equation by simply replacing and simplifying the variables into a form more intuitive to our optimization:

$$I_i^c(x) = R^c(x)T_i(x) + A^c(1 - T_i(x)) \qquad 5.1.1.2$$

The superscript signifies the value is over the three color channels of red, green and blue. The subscript signifies that this variable varies with time. $T_i(x)$ is the global transmittance of each image captured at time $i$. This defines the transmittance at time $i$ for each patch of the scene represented over $x$, where $x$ is the spatial index over the pixels. $T_i(x)$ is defined by the transmittance equation:

$$T_i(x) = e^{-\beta z}$$

The scene radiance $R$ (over the three color channels) in equation 5.1.1.2 varies over time as the illumination position changes (e.g. the sun moving with time). This is according to Lambert's law of diffuse reflection [34]. We assume that the scene radiance does not vary much over the limited period of time that our sequence of images is captured over.

We reshuffle the variables of the equation into the following:

$$I_i^c(x) - A^c = (R^c(x) - A^c)T_i(x) \qquad 5.1.1.3$$

We use least squares regression to optimize the above constraining equation over a sequence of images captured over time. This forms the following error function that represents the squared difference between both sides of equation 5.1.1.3 over all color channels, pixels and time. We minimize this error function in order to find the best-fit values for the transmittance for each patch of the input images at each time $i$ and one 2-dimensional matrix of scene radiance $R^c(x)$:

$$\varepsilon = \sum_i \sum_x \sum_c [I_i^c(x) - A^c - (R^c(x) - A^c)T_i(x)]^2 \qquad 5.1.1.4$$

We minimized the above error function over each 10x10 pixel patch of a sequence of digital images of the same scene captured over time. We tried multiple approaches of solving this minimization problem. First we used a gradient descent search for the minimum solution. We tried to enhance the minimization by performing a constrained search that constrained the transmittances and radiance between 0 and 1. After still being very slow to converge we enhanced the optimization by providing the partial derivatives equations instead of relying on MATLAB's internal ways of computing them automatically. MATLAB's automatic partial derivative computer was slow in our case as we were dealing with large matrices of color pixels. The partial derivatives are as follows:

$$\frac{\partial \varepsilon}{\partial T_i(x)} = \sum_x \sum_c [I_i^c(x) - R^c(x)T_i(x) - (1 - T_i(x))A^c](A^c - R^c(x)) \qquad 5.1.1.5$$

$$\frac{\partial \varepsilon}{\partial R^c(x)} = \sum_i \sum_c [I_i^c(x) - R^c(x)T_i(x) - (1 - T_i(x))A^c](-T_i(x)) \qquad 5.1.1.6$$

These partial derivatives are the gradients of the error function and equal 0 at the minimum solution. After still taking a long time to converge we decided to use our own iterative method.

We solved both partial derivative equations analytically for the minimum solution for $R^c(x)$ and $T_i(x)$ where these partial derivatives reach zero. The atmospheric airlight intensity in each color channel and the image irradiance at specific times and color channels and space overthe image are all known variables. The two partial derivative equations shown above each have one unknown: $R^c(x)$ in the first partial derivative and $T_i(x)$ in the second equation. We reshuffle around the variables and summations in each of these partial derivative equations so that each computes $R^c(x)$ and $T_i(x)$ directly. We then iterate back and forth between solving for $R^c(x)$ and then solving for $T_i(x)$.

The equations for $R^c(x)$ and $T_i(x)$ after reshuffling equations 5.1.1.5 and 5.1.1.6 are:

$$T_i(x) = \frac{-\sum_x \sum_c (I_i^c(x) - A^c)(A^c - R^c(x))}{\sum_x \sum_c (A^c - R^c(x))^2} \qquad 5.1.1.7$$

$$R^c(x) = \frac{\sum_i T_i(x)(I_i^c(x) - A^c + A^c T_i(x))}{\sum_i (T_i(x))^2} \qquad 5.1.1.8$$

There are many variables in this non-linear system of two equations shown above. The transmittance is defined over time and space (over each image taken at time $i$), the image irradiance is defined over space, time and color channel, the atmospheric airlight is defined over the color channels and the scene radiance is defined over space and the color channels. Due to the large number of unknown variables there comes a high possibility of multiple local minima solutions to this problem. In order to try and remedy this we clamped the maximum transmittance to the value of one so that we ensured that exactly one solution scaled to an unknown scalar was returned. When comparing the estimated transmittances to the real transmittances, we also applied this scaling by dividing by the largest value so that the largest transmittance in the set had a value of one. By this, we ensured that both transmittance sets were scaled by the same amount and could be compared. The atmospheric airlight $A^c$ in equations 5.1.1.7 and 5.1.1.8 is set to the largest irradiance value found in the image (i.e. the intensity of the brightest pixel in the sequence of digital images). This is to ensure that one minimum solution is found and scaled by the value of the atmospheric airlight. Unfortunately this does not guarantee that no other local minima will be found. Through visualizing the values of transmittance and scene radiance by plotting them we could see that the algorithm converged to a solution but it was hard to tell if this was the global minima of the problem. We also experimented with trying different starting points for solving the problem. These multiple trials showed the same convergence. This did not add much significant information about whether the algorithm was converging to local minima or global minima.

The basic algorithm of our final iterative approach is as follows:

```
For each patch
      While delta of T > 0
            R <- min(R,1)
            R <- max(R,0)
            T <- min(T,1)
            T <- max(T,0)
            Compute T using equation 5.1.1.7
            Clamp T of darkest image (image with least atmospheric airlight) to 1
            Compute R using equation 5.1.1.8
      End
End
```

The stopping condition of the while loop in the above algorithm is when the change in transmittance values is zero. We used the maximum difference in values of the transmittances as the change in transmittance. Absolute zero could not be reached, so we substituted this with a low value, such as 1e-5. This was sufficient to estimate the transmittances and scene radiance.

In optimizations like these the initial starting points are important in determining how fast the algorithm will converge. In some cases where the problem space is filled with multiple local minima the initial starting values become vital to the success of the optimization. We experimented with values of 0, 0.5 and 1 for the transmittances. The results of these experiment runs are shown in chapter 6.

In order to test the efficiency of this algorithm we ran it on image patches of 10x10 pixel dimensions with random colors (as described in chapter 6). This approach achieved a large speedup over using MATLAB's built in functions and converged in a fraction of the time. Our iterative minimization was able to compute the transmittance and scene radiance for a patch in less than 1e-6 seconds as opposed to the approximately 60 seconds or longer it took MATLAB to. We tested the robustness of this approach by subjecting the input images to noise and saw that it was quite stable under a decent amount of noise. We show the detailed results of this algorithm in the next chapter.

This optimization method is less prone to the errors in atmospheric airlight that are possible in the three dehazing methods. Errors can be introduced in the three dehazing algorithms in the form of errors in the degree of polarization (which is computed from the difference in best polarized image and the worst polarized image) and the dark channel. The measurement of the degree of polarization is a manual process of rotating the polarizer filter and hence is naturally prone to errors. The dark channel's assumption that the subliminal dark channel is equal to zero for out-door images is not always true for some patches of the scene which resulted in the errors shown in Figure 24 and Figure 25. Estimating the intensity of atmospheric airlight from sky regions of the images can be biased as the sky regions are prone to saturation and color clipping. The color optimization algorithm borrows an approach proposed in the dark channel algorithm to measure the atmospheric airlight. This approach is known as the "brightest pixel" approach. This approach estimates the atmospheric airlight by taking the maximum radiance in the sequence of images. This method is still prone to errors in the sky region but avoids the errors due to errors in the degree of polarization and dark channel. We will describe the details of this in the next chapter.

In order to avoid any problems due to the fact that there may be multiple possible solutions to the equations described above, we clamp one of the transmittances to the value of 1 and all other transmittances are therefore scaled with respect to this clamped one. The transmittance chosen to be clamped to 1 is the transmittance of the darkest image (i.e. the image with the minimum atmospheric airlight intensity). This is because the transmittance of this image is bound to have the maximum transmittance value. As all our transmittance values are clamped between 0 and 1, we clamp the maximum value corresponding to the darkest image to 1. This allows the optimization to find only one solution in the space of the problem. Through simulation we verified the soundness of this clamping method (described in the next chapter).

The recovered image of the scene was dehazed well. Refer to chapter 6 for the results. We attempted to enhance the resulting dehazed image by mapping a relation between the pixel values of the sequence of input images and the set of transmittance across them. Each pixel has color values for red, green and blue and a transmittance value (which comes from the patch in which this pixel resides). We plot the graph of the color values for the same pixel throughout the sequence of images against the transmittance. This gives us a graph which we can use to extrapolate pixel values for higher transmittances. By this, we introduce an extrapolation method which estimates values for pixels above the maximum transmittance in the sequence of images. These estimated pixel values make up the dehazed scene and prove to be much better than the original scene radiance recovered by the color optimization. We will display the results and further conclusion of this method in the next chapter.

The color optimization takes into consideration the temporal factor of a sequence of images of the same scene taken over time and solving under the constraint of the haze image formation model. By this, it recovers the transmittance (which can then be further decomposed using the constant depth optimization we will describe in the next section) and scene radiance. This optimization algorithm also provides a standalone dehazing algorithm as shown in the next chapter.

## 5.2 Constant Depth Constraint (CDC)

This second optimization algorithm is a simple optimization based on the relation between the transmittance, atmospheric scattering and depth (described in equation 5.1.2 below). This optimization method extends the already existing dehazing algorithms (polarization-based, temporal and dark channel) and the color optimization described above.

This optimization is based on the simple model of transmittance described by the following equation:

$$T = e^{-\beta z} \qquad \text{5.1.2.1}$$

where $T$ is the transmittance, $\beta$ is the atmospheric scattering coefficient and $z$ is the depth.

Given the transmission map produced by any one of the dehazing algorithms or the color optimization, we can solve for a set of scattering coefficients and a single depth map that best fit equation 5.1.2.1.

We implement the standard linear sum of squares regression to find the best fit variables to the transmittance model. Subtracting one side of equation 5.1.2.1 from the other yields the following error function to minimize:

$$\varepsilon = \sum_i \sum_x [\beta_i z(x) - \log(T_i(x))]^2 \qquad \text{5.1.2.2}$$

We attempted to minimize the error function above using the same steps of the color optimization. These steps included fist trying the built in functions of MATLAB. These are *fminunc* for unconstrained minimization and *fmincon* for constrained minimization. When using both these equations, the converging was extremely slow and therefore needed to be speedup. It was extremely slow even on small patches of dimensions 10x10. We followed the successful approach of the color optimization in speeding up the convergence. We calculated the partial derivatives of equation 5.1.2.2 by hand. These are:

$$\frac{\partial \varepsilon}{\partial \beta_i} = \sum_x [\beta_i z(x) - \log(T_i(x))] z(x) \qquad \text{5.1.2.3}$$

$$\frac{\partial \varepsilon}{\partial z(x)} = \sum_i [\beta_i z(x) - \log(T_i(x))] \beta_i \qquad \text{5.1.2.4}$$

Again, we solved for these partial derivatives (gradients) assigned to zero. We reshuffled these equations so that the two unknowns could be calculated directly. These equations become:

$$\beta_i = \frac{\sum_x T_i(x) z(x)}{\sum_x (\beta_i)^2} \qquad 5.1.2.5$$

$$z(x) = \frac{\sum_i T_i(x) \beta_i}{\sum_i (\beta_i)^2} \qquad 5.1.2.6$$

Using the above equations we are able to iteratively solve for the scattering coefficients and then solve for the depthmap and then loop again. We clamped the maximum scattering coefficient to a value of 1 to ensure that only one possible solution would be found. We compared the estimated set of scattering coefficients with the real set but again scaled so that the maximum of the set is 1. This allowed us to properly compare the estimated and the real values.

The algorithm for this optimization is as follows:

```
while delta of Beta > 0
      Beta <- min(Beta,1)
      Beta <- max(Beta,0)
      Compute Beta using equation 5.1.2.5
      Clamp Beta of brightest image (image with maximum atmospheric airlight) to 1
      Compute z using equation 5.1.2.6
end
```

In order to avoid any problems due to the fact that there may be multiple possible solutions to the equations described above, we clamp one of the atmospheric scattering coefficients to the value of 1 and all other scattering coefficients are therefore scaled with respect to this clamped one. The coefficient chosen to be clamped to 1 is the atmospheric scattering coefficient of the brightest image (i.e. the image with the maximum atmospheric airlight intensity). This is because the atmospheric scattering coefficient of this image is bound to have the maximum value. As all the scattering coefficients are clamped between 0 and 1, we clamp the maximum value corresponding to the brightest image to 1. This allows the optimization to find only one solution in the problem space. Through simulation we verified the soundness of this clamping method (described in the next chapter). The atmospheric airlight is measured using the "brightest pixel" approach described in the previous section.

The stopping condition of the while loop in the above algorithm is when the change in scattering coefficients is zero. We used the maximum difference in values of the scattering coefficients as the change. Absolute zero could not be reached, so we substituted this with a low value, such as 1e-5. This was sufficient to estimate the scattering coefficients and depth scaled so that the maximum scattering coefficient is 1.

In a matter of a few iterations it converges to a solution for small image. On larger resolution images, it was reasonably fast in converging to a sequence of scattering coefficients for each time and a single depthmap. We compared the accuracy of the scattering coefficients and the depthmap with the accuracy of these produced by each individual dehazing algorithm. The results of this comparison are detailed in chapter 6.

In optimizations like these the initial starting points are important in determining how fast the algorithm will converge. In some cases where the problem space is filled with multiple local minima the initial starting values become vital to the success of the optimization. We experimented with values of 0, 0.5 and 1 for the scattering coefficients. The results of these experiment runs are shown in chapter 6.

The idea behind this optimization is to recover the scattering coefficients of a sequence of images of the same scene with the intent of at a later stage correlating them to pollution levels in the atmosphere. In addition to this benefit, we also recover a depthmap with increased accuracy as it loses the scale factor of the scattering coefficient.

# Chapter 6 – Experimentation, Results & Evaluation

# 6.1 Simulating Haze

In order to compare the different dehazing algorithms and constraint optimizations, we created synthetic haze-filled scenes. We used the standard haze formation model used in the three known dehazing algorithms to simulate the effects of haze. The model is described by equation 3.3.4.

To represent out scene radiance without the effects of haze, we created a 100x100 pixel image with randomly generated colors seen in Figure 26.

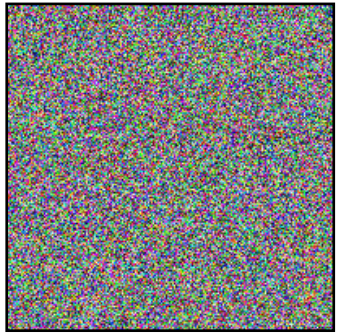

**Figure 26: A simulation of the radiance of a scene using a random value generator**

Haze is a depth-dependent effect in real-life scenes, so in order to simulate this here we divided the image shown in Figure 26 up into patches of 10x10 pixels. Each of these patches was assigned a depth value from 1 to 20.

We simulated five images of the same scene (shown in Figure 26) captured over time. As haze varies with time, we simulated different haze intensities by assigning each of the five images, an atmospheric scattering coefficient and atmospheric airlight intensity. The former represents the scattering of light by the haze in the scene per unit volume of atmosphere. The latter describes the airlight intensity through an infinite depth dominated by airlight scattering. This airlight intensity can typically be seen at regions of low horizon in real-life image. Both these coefficients increase with each other. We assigned the scattering coefficients for each time values of 0.1, 0.15, 0.2, 0.25 and 0.3. We chose this set of values as the most realistic estimates of real-life haze scattering. We gave the images atmospheric airlight intensities of 0.5, 0.6, 0.7, 0.8 and 0.9 as these are realistic intensities of airlight from a scale of 0 to 1. We referred to the previous work in order to obtain these realistic values.

The resulting five images of our scene under simulated different haze conditions are shown in the figure below.

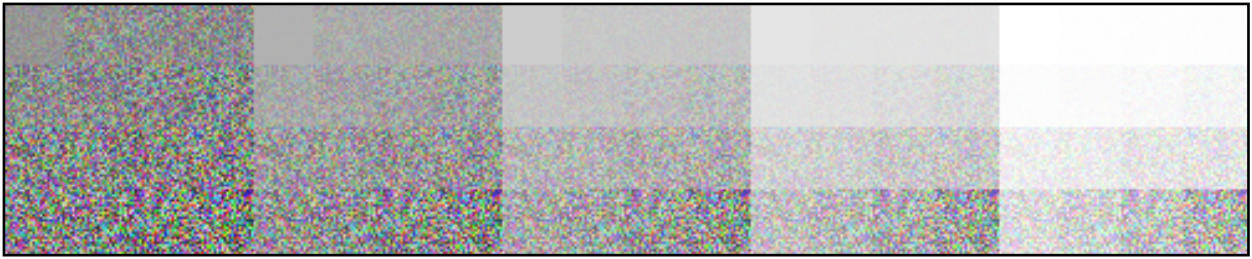

**Figure 27: A sequence of five images showing the simulated haze increasing as the scattering increases. The scene has square patches of different depths (increasing from right to left and bottom up). As the patches at the top are the furthest away they get washed out with the atmospheric airlight.**

These images were the base of the comparisons that are described in the remainder of this chapter. The simulations are accurate as they are based on the haze image formation model used in the known dehazing algorithms and established on known physics models of atmospheric radiative transfer. It has proven to be an effective model in simulating atmospheric scattering and attenuation of radiance over distance. As we synthesize these haze images and know the ground-truth values we put into these simulations, we are able to test the accuracy of the algorithms and hence compare the algorithms presented in this research.

## 6.2 Atmospheric Scattering Accuracy

In order to be able to measure the concentration of airborne particles in the atmosphere from images, it is vital to be able to measure the atmospheric scattering coefficient accurately from visual information in the form of digital images. The three known dehazing algorithms described in the previous work chapter and our presented Color Optimization (CO) dehazing algorithm do not produce direct estimates of the atmospheric scattering coefficient. They only produce transmission maps/depth maps scaled to the atmospheric scattering coefficients. We extend the existing dehazing algorithms and our presented CO algorithm with the Constant Depth Constraint optimization (CDC) to decompose the transmission maps of each of these algorithms into a set of atmospheric scattering coefficients over time and a single unscaled depthmap. By unscaled, we mean a depthmap that is not scaled to the atmospheric scattering coefficients. There may be other unknown scale factors in the depthmap which we do not account for in this research. We compare the following algorithms extended with our CDC algorithm with respect to their accuracy in recovering the atmospheric scattering coefficient. Our goal here is to identify which approach is more accurate in measuring atmospheric scattering from a sequence of images.

The four algorithms under comparison are the following:

1. Color Optimization with Constant Depth Constraint (CO-CDC)

2. Polarization-based dehazing with Constant Depth Constraint (POL-CDC)

3. Temporal dehazing (Dichromatic Framework) with Constant Depth Constraint (DICH-CDC)

4. Dark Channel dehazing with Constant Depth Constraint (DC-CDC)

The accuracy of the scattering coefficient estimation by each algorithm is measured by the following error function:

$$\varepsilon = \sqrt{\sum_i (B_{gt}^i - B_{est}^i)^2}$$

where $B_{gt}^i$ is the ground-truth scattering coefficient at time $i$ and $B_{est}^i$ is the estimated scattering coefficient at time $i$. This error function represents the deviation between the estimated and ground-truth values for the sequence of scattering coefficients over time.

To show the accuracy of the estimated atmospheric scattering coefficients we ran the simulation of haze images described in section 6.1 but each time increased the image noise by a linear amount. We added image noise to the simulated images by increasing the standard deviation of the Gaussian distributed noise from 0 to 0.2. As the image pixel values in our experiments were values from 0 to 1, a standard deviation of 0.2 is a significant amount of noise. We then ran each dehazing algorithm on the simulated images for each noise level. The results show how accurate and stable each approach is in the presence of increasing image noise.

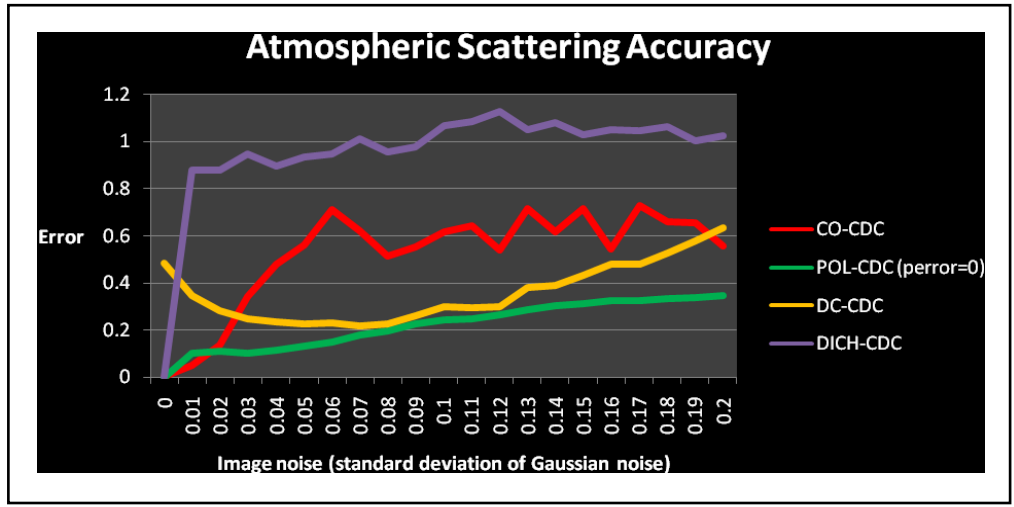

**Figure 28: The accuracy of the estimated atmospheric scattering of the four algorithms.**

In the figure above, the only algorithm that does not start from close to zero error for no image noise at all is the dark channel. This may be explained by the fact that the dark channel overestimates haze by assuming that the subliminal dark channel of out-door haze images is zero when in reality it could be higher than this. The above graph displays the POL-CDC algorithm with degree of polarization equal to 1. A value of 1 is almost impossible in atmospheric scattering. Plots for more natural values of the degree of polarization are shown in the graph below. So although the POL-CDC algorithm looks like the most stable and most accurate plot of the graph above, this is the perfect world scenario which is almost impossible. The dark channel algorithm (DC-CDC) is not accurate for no image noise. This can be explained by the dark channel prior which over estimates haze by assuming that the dark channel of an out-door haze-filled image is zero when in reality it may be more than this. This error can take place, for example, because of the presence of some surfaces/objects in the image which have similar colors to haze. The dichromatic temporal algorithm (DICH-CDC) shows a huge increase in error for low image noise then steadies off. The CO-CDC algorithm on the other hand starts from zero for no image noise at all and increases linearly before leveling off but fluctuating a little. The fluctuations in the graphs above and below are due to the random factor of the haze image simulator. For each run the scene radiance is randomly generated. We ran each simulation multiple times at each noise level (20 times) and computed the average error over these times. This was done to avoid any large fluctuations due to the randomness of the generated scene radiance. The fact that CO-CDC has lower errors for low levels of image noise proves it to be a more accurate approach to the parabolic-like curve of the DC-CDC which does not have zero-error at any noise level. We can also see from the above graph that the CO-CDC is much more accurate than the DICH-CDC algorithm. As for the POL-CDC algorithm, we see how CO-CDC performs with more natural values of degree of polarization. We obtained these more natural values by looking at [33] and running the polarization-based dehazing algorithm on real images of urban scenery. The range obtained is between 0.25 and 0.45. For this reason we simulated degrees of polarization within this range, as can be seen in the figure below.

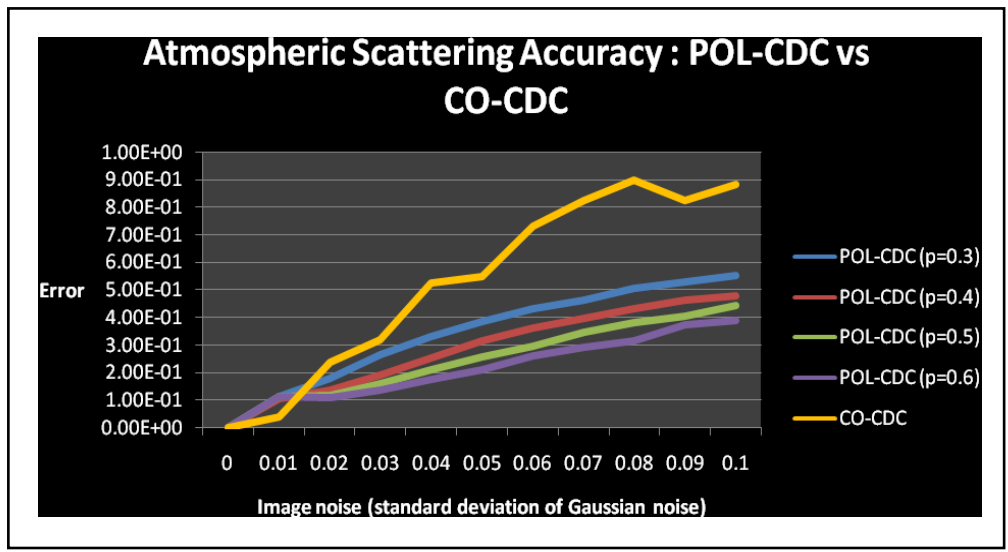

**Figure 29: The accuracy of the atmospheric scattering for CO-CDC and POL-CDC under natural values for the actual degree of polarization (0.3, 0.4, 0.5 and 0.6).**

The POL-CDC with actual degree of polarization equal to 0.6 in the above figure is the most accurate under most of the noise levels. The CO-CDC shows more accuracy at noise levels less than a standard deviation of 0.02. As the degree of polarization is a value between 0 and 1 (but never in reality equal to 1), the range of standard deviation of image noise shown on the horizontal axis (0 - 0.1) is quite significant. You can see that the accuracy of atmospheric scattering estimation decreases with a decrease in actual degree of polarization of the light in a scene. When compared with the CO-CDC algorithm, the POL-CDC is more accurate for noise levels with standard deviation above about 0.02. An assumption was made here that the POL-CDC algorithm estimates the actual degree of polarization perfectly. This is seldom the case as the estimation of the degree of polarization is dependent on a manual process of rotating the polarizer filters. This subjects the estimated degree of polarization to significant errors which would affect the measured atmospheric scattering greatly. We will describe shortly, the effect of errors in the estimated degree of polarization on the atmospheric scattering estimation. In the figure above we can see that the degree of polarization is the parameter that the POL-CDC algorithm uses to extract the haze from the scene and from this haze estimates the atmospheric scattering. The amount of haze measured all depends on the amount extracted by means of the difference between the best and worst polarized images. As the degree of polarization decreases, less haze is extracted and therefore less atmospheric scattering is estimated. The CO-CDC algorithm does not change with variations in the polarization of the scattered light. CO-CDC is based on temporal knowledge gathered over time.

All the known algorithms have a parameter which is estimated from the image. This parameter is the atmospheric airlight intensity. In the CO-CDC approach we assumed the highest intensity of ths parameter which was taken from the pixel with the highest irradiance intensity in the sequence of input images. Since our images are of mostly diffuse far away urban scenery, the brightest pixels in the images are that of the sky region. To avoid errors in this approach due to objects in the scene reflecting more light than the sky, we can simply average over a percentage of the highest intensity pixels. This is a technique presented in the dark channel prior dehazing algorithm [42]. The atmospheric airlight intensity, also known as the airlight at infinite distance is an exceedingly important parameter and there are many proven and proposed ways of estimating it in previous work. Each dehazing algorithm has a main technique and proposed solutions to finding the most accurate atmospheric airlight intensity. The main method in the polarization algorithm is from low sky regions which can be prone to color saturation and clipping. This sky estimate may also misrepresent the airlight color and intensity in local regions of the scene as it is biased by an infinite column of Rayleigh scattering. The dichromatic dehazing method estimates two atmospheric airlight intensities for each of its input images (taken at different times). It does this by solving a line fitting problem (refer to the previous work chapter). The dark channel estimates the atmospheric airlight by finding the maximum pixels in the dark channel and then finding the pixels in this set of pixels with the brightest image irradiance value. As you can see the ways of estimating this parameter which is common to all the known dehazing algorithms are numerous and various.

We injected errors into the atmospheric airlight estimation of each of the four algorithms (known dehazing algorithms and color optimization) to see how accurate they would estimate the scattering coefficients. The figure below shows the results of this comparison.

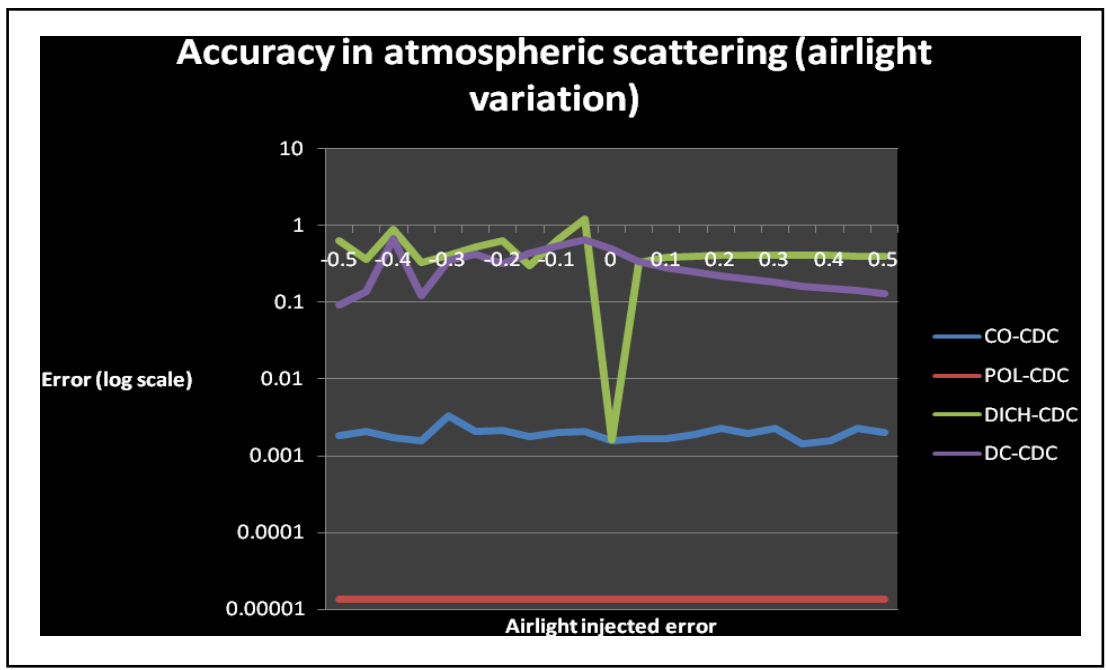

**Figure 30: This graph shows the accuracy of the atmospheric scattering estimation with variations in the atmospheric airlight intensity.**

In the figure above we can see the changes in accuracy of the four algorithms as errors are injected into the atmospheric airlight intensity. As different errors are injected into the estimated atmospheric airlight we see that both the POL-CDC and CO-CDC are almost constant (notice the log scale on the vertical axis). The DICH-CDC and DC-CDC are both very dependent on the atmospheric airlight and therefore fluctuate quite significantly. This is intuitive as the DICH-CDC bases most of its computations on the estimated airlight intensity of two different haze images. Refer to chapter 5 to see these airlight dependent computations. An error in the atmospheric airlight in the two input images of the dichromatic algorithm affects the outcome two-fold. The dichromatic framework is founded upon changes in haze intensities over time. This makes it an exceedingly inefficient algorithm in measuring atmospheric scattering properties. This is because it relies on noticeable fluctuations in haze. The dichromatic framework faces another problem. This is the assumption of constant spectral illumination under overcast skies and the assumption that the color of airlight is constant. In order to combat changes in illumination a more recent dehazing algorithm has emerged but unfortunately not covered in this thesis due to its recent date of publication. This dehazing algorithm is called the Single Image Dehazing [43]. The dichromatic dehazing method faces the problem of illumination changes which can drastically affect the shading and colors of scenes. In the simplest case of the movement of the sun across the sky during the day, this made us unable to apply this dichromatic framework method to real images. In addition to the problems of illumination changes, the haze was not varying much during the period of a few hours we captured the images in (images shown in **Figure 32**). The dark channel is also dependent on the atmospheric airlight intensity as it uses this to normalize the input image and compute the scene radiance. In the case of the polarization based dehazing algorithm, the atmospheric airlight is just used as a scale factor that does not alter the end result of the estimated scattering. The way this algorithm dehazes is using the polarization information of the two images.

For the polarization dehazing algorithm, the accuracy of measuring the atmospheric scattering coefficients is dependent on the accuracy of the estimated degree of polarization. We simulate the best and worst polarization images over time with a value of 1 for the actual degree of polarization. This means that the light is fully polarized in the scene. We inject errors into the estimated degree of polarization that decreases it from its actual value. The injected error represents the error caused by the manual process of rotating the polarizer filter parallel and perpendicular to the plane of incidence (refer to chapter 2) in order to capture the best and worst polarized images. The figure below shows the effects of errors in the estimated degree of polarization on the accuracy of the measured atmospheric scattering.

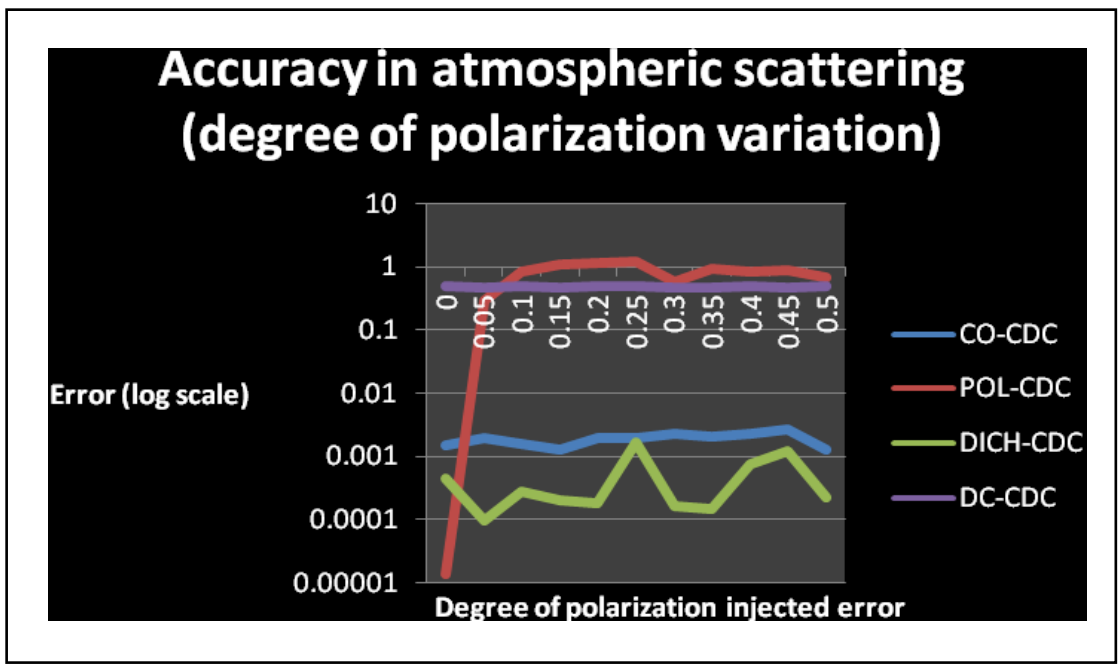

**Figure 31: This graph shows the accuracy of the atmospheric scattering estimation of the four algorithms with variations in the degree of polarization (degree of polarization = 1 - injected error).**

In the figure above, it can be observed that as different errors are injected into the degree of polarization, all the algorithms except the POL-CDC are constant in accuracy (the degree of polarization is critical to the POL-CDC algorithm). The fluctuations in the plots for DICH-CDC and CO-CDC can be explained by the random generation of the simulated haze images for each run. The POL-CDC algorithm is theoretically very accurate for degrees of polarization equal to 1, which means the haze is fully polarized and therefore perfectly measured. This is far from the reality due to factors such as multiple scattering, ground reflection, illumination and manually setting the polarizer filters. What we also notice is that the dark channel is overestimating the atmospheric scattering by quite a significant amount. This also can be seen in **Figure 30**. An explanation for this lies in the fact that the dark channel algorithm assumes that the subliminal dark channel present in out-door haze-filled images is zero. In reality, this is not the case and so by this, we get an overestimate of haze. We corroborated this by inspecting measurements of atmospheric scattering using the dark channel method for real images. The sequence of five real haze images can be seen in the figure below.

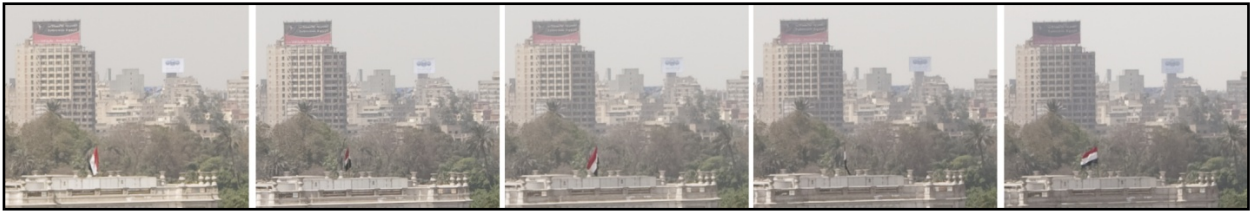

**Figure 32: A sequence of five images taken over time. Each picture was taken at half an hour intervals. This is the sequence of images that is used by the CDC extended algorithms to measure the sequence of atmospheric scattering coefficients.**

The atmospheric scattering coefficients were estimated by the color optimization algorithm and the dark channel algorithm. The dichromatic dehazing failed to dehaze the sequence of images due to changes in the position of the sun resulting in changes in the color shades due to Lambert's law of diffuse reflection [34]. The spectral illumination also changes slightly as the sun moved lower in the sky. Another factor that prohibited the dichromatic temporal dehazing algorithm from working here is the ever so slight variations in haze. The polarization dehazing algorithm gave erroneous results and was unable to converge to a proper solution. This could be due to the extra error prone parameter of the degree of polarization that is estimated based on two images taken under different orientations of the polarizer filter. This so far is a manual process of rotating the filter by hand. If this process can be automated given calibration information such as the position of the sun and the direction the camera is pointing, it would enhance this approach as an atmospheric monitoring method.

In addition to recovering accurate estimates of atmospheric scattering from the images, the standalone Color Optimization algorithm (without CDC), returns a dehazed image of the scene. We shall see this in the next section.

# 6.3 Color Optimization Dehazing

The color optimization is an algorithm that takes in a sequence of images of a constant scene taken over time and recovers the scene radiance in the absence of the visual effect of haze and a transmittance map which can be easily converted into a depth map. In this section we will see the results of this algorithm from the perspective of dehazing.

The color optimization algorithm was tested using the simulated synthetic haze images described in section 6.1. In this section we display and discuss the results of this Color Optimization method and how it compares to the other dehazing algorithms in its performance in dehazing and depthmap estimate accuracy.

An assumption is made in the color optimization that the scene radiance is constant over time. This is an inaccurate assumption as movement of the direction of illumination (e.g. movement of the sun across the sky) causes a change in the intensity of the scene radiance. This is described by the Lambertian law of diffuse reflection [34]. To account for this we normalized each image by dividing by the average irradiance over a patch of the scene in the foreground. The reason we chose a patch of the scene in the foreground is that foreground objects are least affected by haze and therefore give us more accurate information on how the scene radiance intensity varies with changes in illumination. Further away objects would be affected greatly by haze and would therefore give inaccurate variations of scene radiance intensity. We chose a flat surface at the bottom left corner of the scene that did not fall into shadows (see Figure 33). The radiance over this patch varies from image to image as the sun moves across the sky. By dividing by the average radiance over this patch, we factor out the variation in scene radiance taking place behind the layer of haze in the images. The patch can be seen in the figure below.

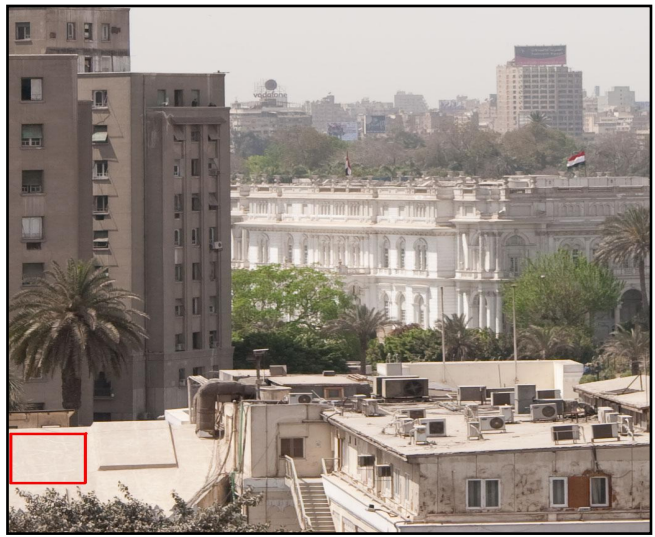

**Figure 33: The patch of foreground chosen to account for the variation in scene radiance due to the change in illumination. The patch is displayed as the red rectangle in the bottom left corner of the image above.**

By simply applying the color optimization algorithm described in section 5.1 and accounting for the change in scene radiance due to illumination as described above, we are able to dehaze a scene captured by a sequence of haze images taken over time. In Figure 34, we can see the sequence of four images taken over time and the final resulting dehazed image recovered by the color optimization. Here we estimated the atmospheric airlight intensity as the brightest pixel in the sequence of input images.

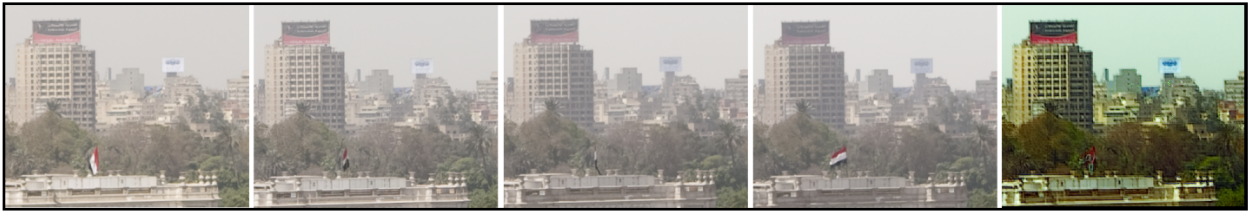

**Figure 34: A sequence of real images captured over time and the resulting dehazed image (rightmost). The layer of haze has been removed. The color and contrast is restored. The sky is returned to its clear-day light blue color.**

We compare our dehazing algorithm with two other known dehazing algorithms: polarization-based dehazing and the dark channel dehazing algorithm. Figure 35 shows the dehazed scene produced by these two dehazing algorithms and our dehazed scene. Both the polarization and dark channel algorithms removed the airlight of the sky, creating a black void similar to the vacuum of space. Our dehazing algorithm is more sensitive by incorporating temporal knowledge. The temporal knowledge is sensitive to the fluctuations in haze from image to image. This makes it optimize for the changes in haze and not blindly remove all the haze from the scene. This results in a depolluted scene. The thick layer of haze seen in Figure 34 (due to suspended particles) has been removed. What remains in the sky region is the natural color of the sky. The white outlines in the dark channel image in the figure below are due to the blockiness effect of the patch processing of this algorithm.

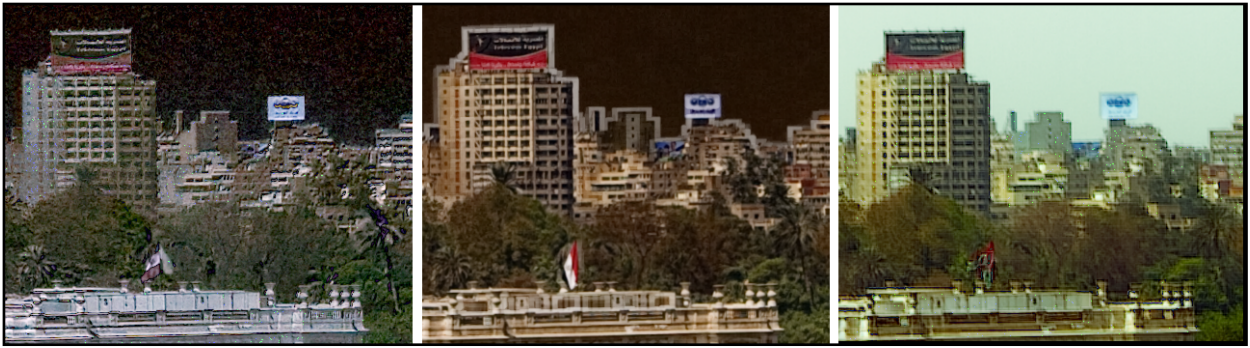

**Figure 35: The results of the polarization-based dehazing (left), dark channel dehazing (middle) and our Color Optimization dehazing (right). These images show that the color optimization is a novel way of dehazing scenes.**

From a vision perspective of dehazing, the color optimization recovers the color and contrast of the scene similar to the polarization and dark channel dehazing methods. In the above image we used patches of size 4x4 pixels to give reasonably sharp dehazed images. The white outline on the horizon, in the dark channel image in the middle of the above figure, where the buildings meet the sky is due to the patch processing.

The output of the color optimization algorithm can be used with the Constant Depth Constraint (CDC) to recover the atmospheric scattering coefficients and a depthmap of the scene. The CDC decomposes the transmittance measured by the color optimization algorithm into atmospheric scattering coefficients for each of the taken over time and a single depthmap. The experiments for combining the color optimization and CDC algorithms together are described in the next section. From the point of view of dehazing, the CDC does not add to the dehazing capabilities of the color optimization algorithm (shown in Figure 34 and Figure 36). If we multiply the depthmap by each of the scattering coefficients we should obtain a scaled depthmap for each image (scaled by a single scalar representing the atmospheric scattering at each time). This scaled depth can be substituted into equation 5.1.1.1 to obtain the scene radiance (dehazed image). This does not produce better dehazed images. This is because the scattering coefficients are global measurements across each image. The scattering coefficients are average measures of the atmospheric scattering across the whole images measured for the purpose of monitoring the overall atmosphere in the scene. These coefficients do not vary spatially across the images (i.e they are global across each image). The color optimization on the other hand optimizes for the best transmittance per pixel. This means that the color optimization is spatially sensitive in its dehazing capability while the CDC is not. This makes a difference from a dehazing perspective as haze can be non-homogeneous in density across urban scenes. Different regions of the images may fall under slightly different atmospheric scattering. As a result, applying the CDC and substituting the scaled depth into equation 5.1.1.1 produces dehazed images with color distortion in parts of the image. Because of this, the recovered dehazed image shown in Figure 34 and Figure 35 is not improved in anyway by using the CDC.

# 6.4 Depthmap Recovery

The color optimization is an algorithm that takes in a sequence of images of a constant scene taken over time and recovers the scene radiance in the absence of the visual effect of haze and a transmittance map which can be easily converted into a depth map. When extended with the CDC constraint optimization, the color optimization retrieves the atmospheric scattering coefficients as we saw in section 6.2 and an unscaled depthmap of the scene. By unscaled here, we mean that the depthmap is no longer scaled by the atmospheric scattering coefficients. This is the case in all the known dehazing algorithm described in this research. In this section we will see the results of two comparison experiments:

1. Scaled depthmap recovery: Comparison of the depthmap produced by the color optimization (without the CDC constraint) algorithm and the depthmaps produced by the known dehazing algorithms.

2. Unscaled depthmap recovery: Comparison of the depthmap produced by the CO-CDC algorithm compared with the known dehazing algorithms also extended with the CDC constraint optimization.

We measure the error of the recovered depthmap by computing the standard deviation of the pixels of the depthmap from the ground-truth values for these pixel determined by the simulations (refer to section 6.1). This error is equivalent to a root mean squared error between the ground truth depth and estimated depth summed over the pixels of the depthmaps. We sum the square of the differences over all the pixels and take the square root of this as follows:

$$\varepsilon = \sqrt{\sum_x (D_{gt}(x) - D_{est}(x))^2}$$

6.4.1

where $D_{gt}$ is the ground-truth depthmap, $D_{est}$ is the depthmap estimated by the dehazing algorithm under analysis. We sum this square of the differences over all pixels $x$. This is the standard way we measure the accuracy of the depthmap. The lesser this error is, the more accurate the estimation of depth. In the experiments in this chapter we simulate a sequence of five haze images (refer to section 6.1). These five simulated images produce five depthmaps when input into the dehazing algorithms. In the case of the dichromatic temporal algorithm, we simulated 6 images in order for it to produce five depthmaps for the five different simulated times. The overall trend is that the more haze present in the simulated scene, the more accurate the depthmap.

Our Color Optimization algorithm is a temporal approach in the fact that it combines temporal knowledge and recovers one single depthmap in the end. In order to compare the depthmaps produced by the known dehazing methods with our Color Optimization algorithm, we picked the best depthmap produced for the sequence of images by the known dehazing methods and compared this to our single produced depthmap.

# 6.4.1 Scaled Depthmap Comparison

Here we compare the scaled depthmaps (scaled by the atmospheric scattering coefficients) of the color optimization and the three known dehazing methods: polarization, temporal (dichromatic) and the dark channel dehazing.

The transmission maps produced by the color optimization algorithm on the simulated haze images can be seen in Figure 38. As the haze increases with time, the transmittance decreases, therefore become darker in the grayscale images. Depthmaps scaled to the scattering coefficients can be retrieved simply by applying a logarithmic function to the transmittance maps shown below and negating the result. This is in accordance with the model described by the transmittance equation 5.1.2.1.

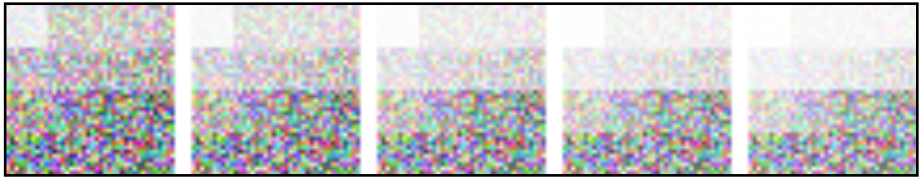

Figure 36: Simulated haze scenes each containing 16 patches of various depths

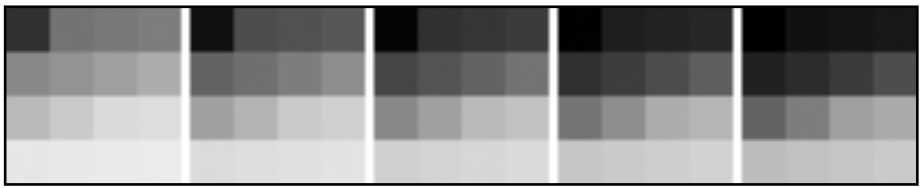

Figure 37: This figure shows the transmission maps produced by the CO algorithm using the images shown in Figure 37. The figure shows a decrease in transmittance as haze increases. The darker the patch appears, the less the transmittance.

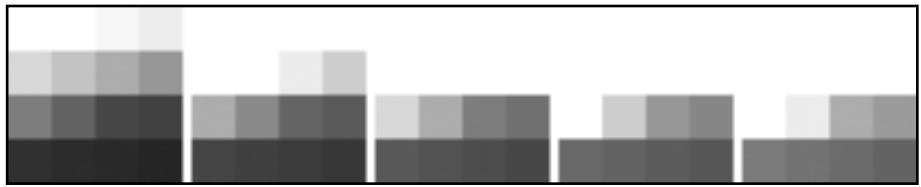

Figure 38: This figure shows the scaled depthmaps computed from the transmission maps of Figure 38 using equation 5.1.2.1.

The transmission maps recovered by this method are compared to the ground-truth transmittance maps in the following figures. Under no image noise, the estimation is exact as seen by Figure 40.

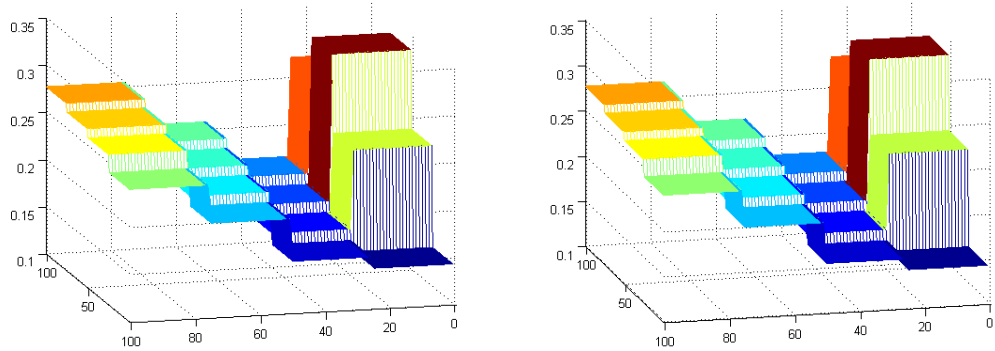

**Figure 39: The actual transmission map (left) and the estimated transmission map (right) of the scene. This shows the accuracy of the method in the absence of noise**

As noise is added to the sequence of input images, errors start to appear in the estimated transmissions. Once the image noise, modeled by a Gaussian distribution, exceeds a standard deviation of 0.03, errors start appearing in the relative depths of the estimated transmission map.

We used equation 6.4.1 to compute the errors of the estimated transmission map. We can see in Figure 45 the accuracy of the color optimization algorithm compared to the other dehazing algorithms under variations in image noise.

The following table shows the error in accuracy of the estimated transmittance map as the image noise varies and the actual transmittance in the scene changes. As less and less haze is present (i.e. transmittance increases), the estimated transmittance becomes less accurate. This is consistent with the known dehazing algorithms.

| Actual transmittance | Standard deviation of noise | | | |
|---|---|---|---|---|
| | 0 | 0.01 | 0.05 | 0.1 |
| 0.1 | 0.042413 | 0.69682 | 1.1652 | 1.6013 |
| 0.2 | 0.058594 | 1.0308 | 1.8596 | 2.5736 |
| 0.3 | 0.07703 | 1.2686 | 2.6939 | 3.0337 |
| 0.4 | 0.10345 | 1.54 | 3.0234 | 4.3963 |
| 0.5 | 0.18084 | 1.3533 | 3.4955 | 4.2075 |
| 0.6 | 0.13892 | 1.7523 | 3.6701 | 5.5214 |
| 0.7 | 0.10241 | 1.671 | 4.3966 | 5.9173 |
| 0.8 | 0.13017 | 1.9694 | 4.249 | 6.0566 |
| 0.9 | 0.18003 | 2.1915 | 4.8276 | 6.5583 |

**Figure 40: This table shows the error in accuracy of the estimated transmittance map as the image noise varies and the actual transmittance in the scene changes.**

We ran the color optimization algorithm on a sequence of real images (not simulated) captured in the downtown Cairo area. Figure 42 shows the transmission map recovered of the scene. Further away objects in the scene appear darker. This is intuitive as the transmission of objects further away decreases as more haze falls in front of them. The transmission map is essentially a depthmap of the scene as the overall trend is that the darker the object appears – the further away it is. A 3D reconstruction of the scene is shown in Figure 43. This depthmap is scaled to the atmospheric scattering coefficient and computed using 5.1.2.1. The displayed depthmap was computed from the transmission map of returned for the first image of the sequence. The color optimization method produces one of these for each image in the sequence.

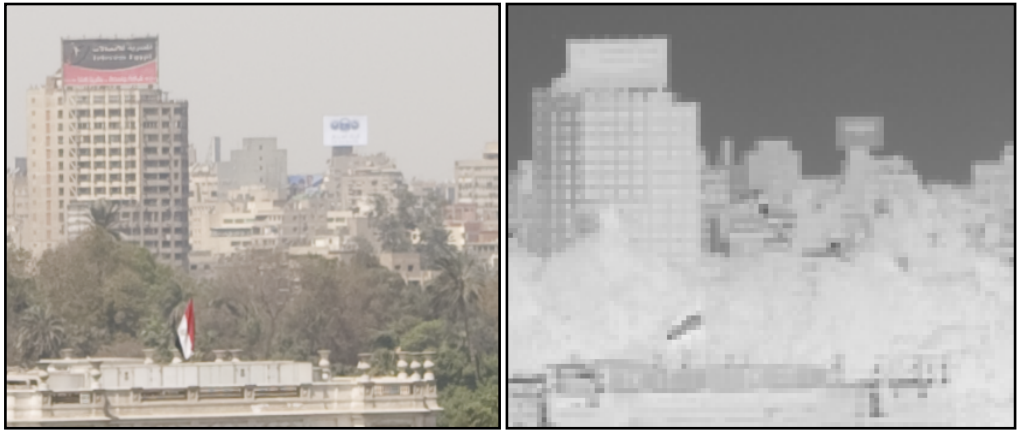

**Figure 41: The actual scene and the transmission map recovered. There is a slight decrease in grayscale intensity for further objects in the scene.**

The transmission map shown above is erroneous as can be seen by the darkish lower left corner of the map. This maybe the case for individual patches, but the overall trend of lighter intensity for closer objects is met. The blockiness here is due to processing the image patch by patch (4x4 pixels).

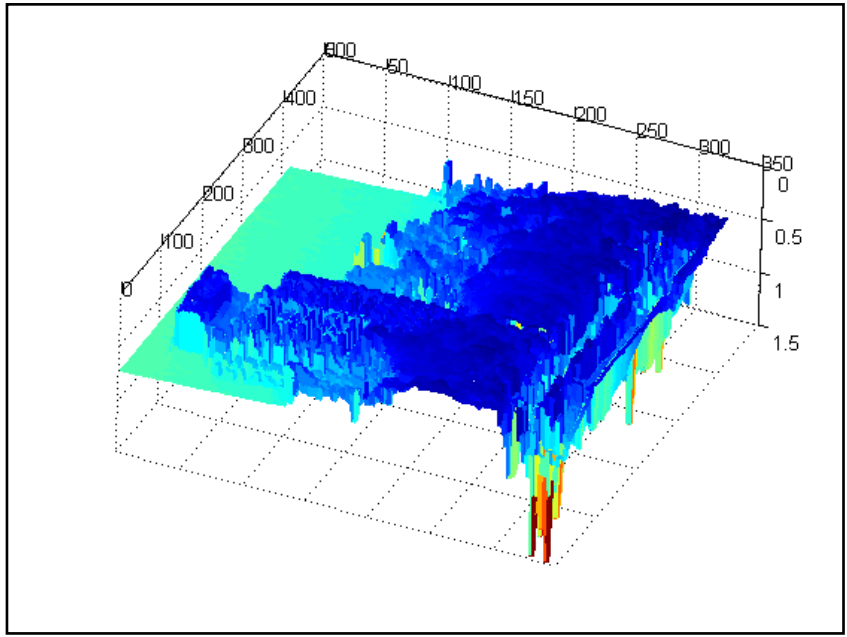

**Figure 42: The scaled depthmap of the scene shown in figure 42 in 3D (rotated and tilted to show the 3D depth effect) produced by the color optimization algorithm.**

The scaled depthmap shown above is scaled to the scattering coefficient of the atmosphere. It is consistent with the scene shown in figure 42. The depthmap above is rotated in order to show the scene in 3D. The building, sky and trees are all consistent with the actual scene shown in figure 42. There are some errors shown in the lower left corner of figure 43 which are scene points in the very foreground of the image.

Figure 42 shows the original image which is the first in the sequence of five real images taken over time and its corresponding transmission map computed by the color optimization algorithm. Figure 43 shows the 3D reconstruction of the scaled depthmap computed directly from the transmission map using the Constant Depth Constraint optimization. As you can see the 3D depthmap is consistent with the original image of the scene showing the successful recovering of the relative scaled depth of the scene.

We extend this algorithm by feeding the sequence of transmission maps produced into the constant depth optimization algorithm (CDC) to retrieve the scattering coefficients and a more accurate depthmap of the scene. We will see the results of this enhancement in estimating the depthmap in the next section (scaled depthmap comparison).

The recovered scaled depthmaps of the four dehazing algorithms were compared as noise was increased. The figure below shows the results of this.

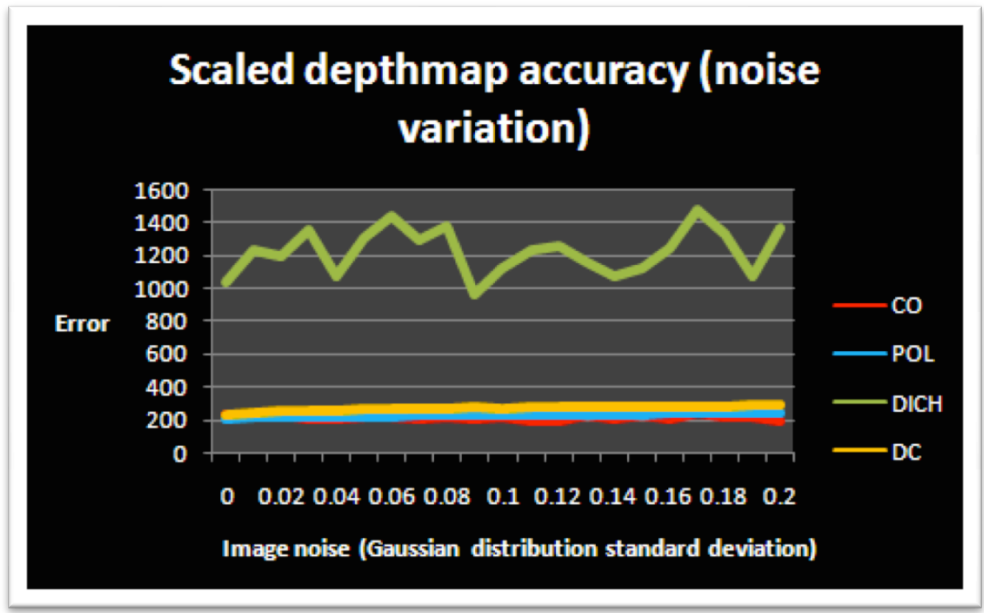

**Figure 43: The scaled depthmap accuracy for the four algorithms.**

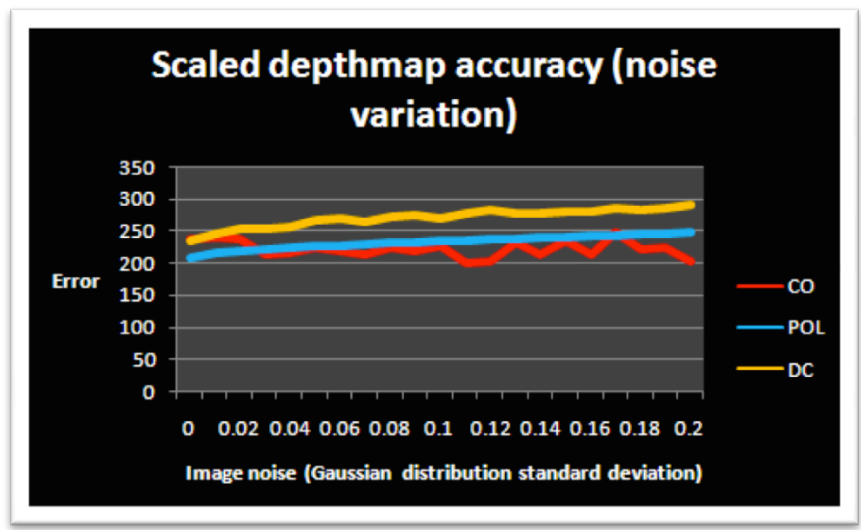

**Figure 44: A close up of the above graph showing the slightly improved accuracy of the color optimization algorithm**

The figures above show that the color optimization algorithm has very slightly increased accuracy over the other three dehazing algorithms. The dichromatic temporal dehazing algorithm has the worst accuracy. This can be explained by the fact that it relies on the color pixels of two different images which doubles the effect noise has on its estimations. The polarization dehazing algorithm is also based on two images but uses them for polarization information which is averaged over many pixels. This averages out the noise and allows it to recover an accurate scaled depthmap of the scene. The minimum errors at 0 image noise are not equal to zero because here we are experimenting with the scaled depthmap of the scene and therefore they are scaled up by the atmospheric scattering coefficients.

We have shown the capability of the color optimization algorithm to recover the transmission map and hence the scaled depthmap of the scene being captured by the sequence of images taken over time. We showed the accuracy of this in simulation under different levels of image noise and for real images and showed that the depthmap recovered is consistent with the imaged scene.

# 6.4.2 Unscaled Depthmap Comparison

The CDC constraint algorithm, when applied to any of the three known dehazing algorithms and the color optimization (CO) algorithm presented in this research, recovers an unscaled depthmap. The depthmap is unscaled as it is no longer scaled to the atmospheric scattering coefficients. It may be scaled to some other unknown scalars but finding these is not within the scope of this research. The CDC algorithm decomposes the transmittance maps produced by the dehazing algorithms (including CO) into a set of atmospheric scattering coefficients and a single unscaled depthmap of the scene. We focus on the accuracy of the unscaled depthmap in this section.

We used the sequence of five simulated haze images to compare the unscaled depthmap produced by the four dehazing algorithms: polarization, temporal (dichromatic), dark channel and our CO algorithm. Each of these algorithms was extended with the CDC constraint optimization to recover the unscaled depthmap. We evaluate the accuracy of these algorithms by comparing each of them to the ground-truth values of the simulation.

The figure below shows the accuracy of the four algorithms when run on the simulated haze images. The error was computed using function 6.4.1. The image noise axis is in units of standard deviation of Gaussian distribution. The errors of the algorithms under no image noise are not perfect zeros because the recovered depth is scaled differently to the ground-truth depth. This is because to avoid multiple solutions, we clamped the largest scattering coefficient to 1 and therefore set a different scale factor for both the recovered scattering coefficients and the depths. The estimated unscaled depths for all the dehazing algorithms except the dark channel are perfectly accurate at image noise of zero but scaled up by the largest scattering coefficient. This is the explanation why the minimum error is approximately not equal to zero when computed relative to the real depth.

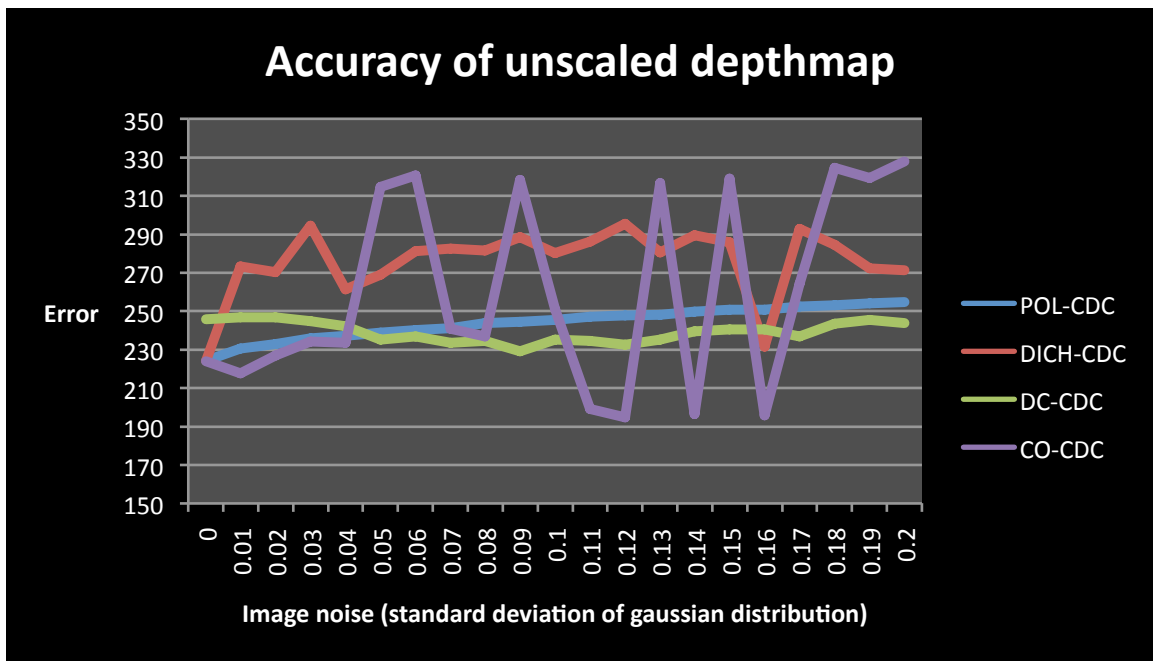

**Figure 45: The accuracy of the unscaled depthmap for the four algorithms extended by the CDC.**

In the figure above we see that the most accurate unscaled depthmap produced from image noise ranging from standard deviation 0 to 0.05 (5% image noise standard deviation) is recovered by the color optimization method. This method uses temporal information gathered over a sequence of images of the same scene captured over time. The CO-CDC displays a very large fluctuation in its error measurements after 0.05 image noise standard deviation. There are multiple possible explanations for this. The first lies in the possibility of multiple local minima in the problem space of the nonlinear image haze model which the first part of the CO-CDC algorithm is trying to solve. Multiple local minima may be found for every iteration of the CO-CDC algorithm. There may be multiple local minima in the color optimization part and/or the constant depth constraint optimization part. Another reason maybe due to the random synthesis of the simulated haze scenes. The scene radiance is randomly generated each time before the haze model is applied to it. This random generation may cause the instability in error measurements. Despite the large fluctuations, the error of the CO-CDC, if averaged out to a smooth graph, would be comparable to the POL-CDC and DC-CDC error measurements. The DICH-CDC algorithm takes two different images into consideration at a time and is dependent on quite a few external parameters such as the spectral illumination. Changes in color and shades throw this algorithm off balance. This is because this algorithm is designed for overcast foggy illumination, which conserves the spectral qualities of the image. This also makes this algorithm prone to errors due to image noise that may affect the colors of the images. This is an explanation for why the DICH-CDC is quite higher in error than the POL-CDC and DC-CDC. Our CO-CDC algorithm does show increased accuracy in the range of image noise from 0.1 to 0.16 but again suffers from large fluctuations.

Due to the odd nature of the large fluctuations observed in Figure 46 we performed statistical analysis to see if the fluctuations were consistently being produced and if multiple trial experiments displayed confident results (i.e. trials produce similar means and variances). Each trial runs the CO-CDC algorithm on a synthetic random scene and increases the image noise one step at a time. At each step the error of the recovered unscaled depthmap is computed. We performed analysis of variance on the multiple trials. We ran a t-test (assuming unequal variance) on the trial results. The P-value returned showed confidence in the results. The results are displayed in the table below.

| t-Test: Two-Sample Assuming Unequal Variances | | |
|---|---|---|
| | *Variable 1* | *Variable 2* |
| Mean | 281.0014286 | 277.9033333 |
| Variance | 2235.042803 | 2313.855113 |
| Observations | 21 | 21 |
| Hypothesized Mean Difference | 0 | |
| Degrees of freedom | 40 | |
| t Stat | 0.21049962 | |
| P(T<=t) one-tail | 0.417173198 | |
| t Critical one-tail | 1.683851014 | |
| P(T<=t) two-tail | ==0.834346396== | |
| t Critical two-tail | 2.02107537 | |

The P-value to be considered here is highlighted and is equal approximately to 0.83 which is way above the significance level (alpha=0.05). This shows that the two result distributions are very closely correlated. This can be seen intuitively by looking at the mean and variance under the two columns: "Variable 1" and "Variable 2" in the graph above. We ran the trials multiple times and ran t-tests on each pair. The P-values returned were very close to the value shown in the table above. This shows that the confidence in the experimental trials is high and therefore one of the explanations described above may be causing this fluctuation. A more thorough investigation may be needed to determine the cause of this.

We also compared the estimated unscaled depthmap produced by the four algorithms on real haze images. We used the five images taken at half an hour intervals (4 of these images can be seen in Figure 35) and fed them into each dehazing algorithm. The CO-CDC algorithm's recovered unscaled depthmap can be seen in the figure below. As you can see it is again, consistent with the actual image shown in figure 42.

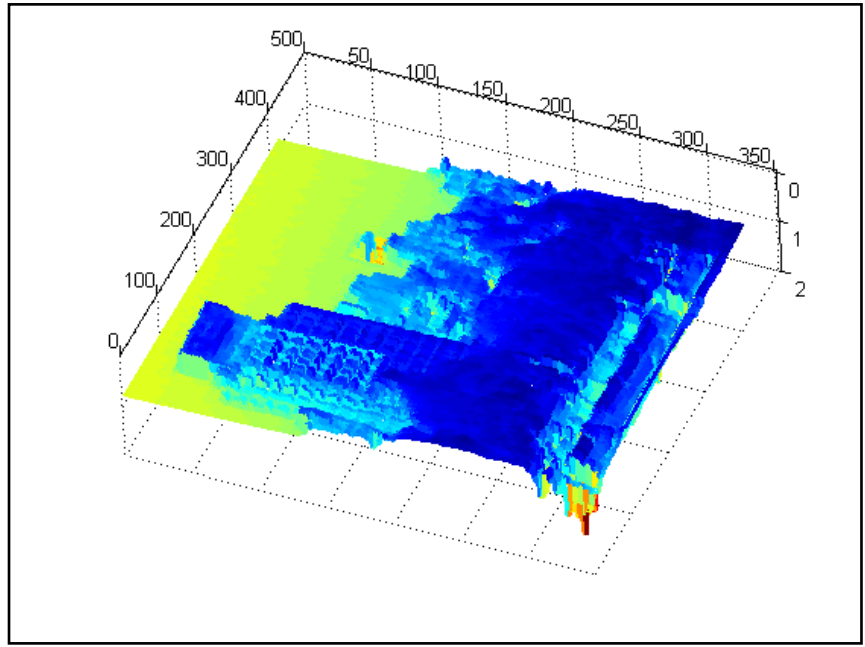

**Figure 46: The unscaled depthmap recovered by the CO-CDC algorithm. It is consistent with the imaged scene (figure 42) as can be seen by the building and vegetation that are slightly elevated above the flat sky plane and other objects.**

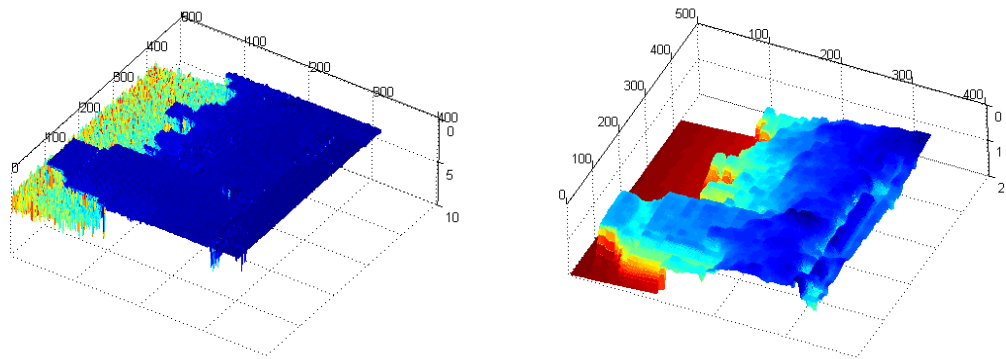

**Figure 47: The unscaled dethmaps recovered by the CDC algorithm applied to the polarization dehazing algorithm and the dark channel dehazing. The result of POL-CDC is shown on the left and the result of the DC-CDC method is shown on the right. You can see that these are consistent with the CO-CDC's depthmap shown in figure 47 and the actual image shown in figure 42.**

As we do not know the ground-truth of these we are unable to compare the actual depth values of these depthmaps. We can visually verify that these depthmaps are consistent with each other and consistent with the original image shown in figure 42. The POL-CDC algorithm's depthmap shown above has noise in the sky region which is not present in both the color optimization and the dark channel depthmaps. The accuracy of the retrieved depthmaps has been compared by simulation.

The dichromatic dehazing was unable to produce a depthmap from the sequence of five real images captured over time. The reason behind this is that there are illumination changes in the images due to the change in the position of the sun. The sun's illumination changes color slightly as it gets lower in the sky and its rays has to pass through more and more atmosphere. This is described by the Rayleigh scattering law [35]. In addition to this, as the sun changes position, the objects in the scene take on different color shades according to Lambert's law of diffuse reflection [34]. To make matters worse on the dichromatic algorithm, there were hardly noticeable variations in the haze levels from image to image.

In order to compare the accuracy of the unscaled depthmap produced by the color optimization algorithm and the depthmap produced by each of the dehazing algorithms, we use Google Maps [40] to measure ground-truth distance ratios and see how close the depthmaps conform to this.

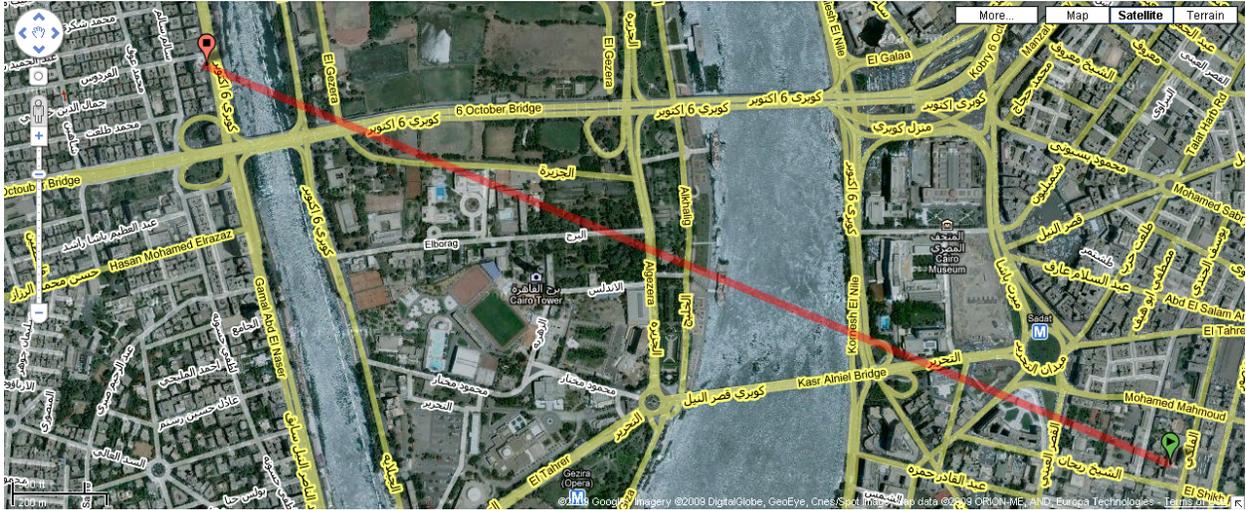

**Figure 48: This figure shows the distance between the point where the image was captured to the billboard on top of the building marked in the figure below. The distance of the line shown above is approximately 2.22303 km.**

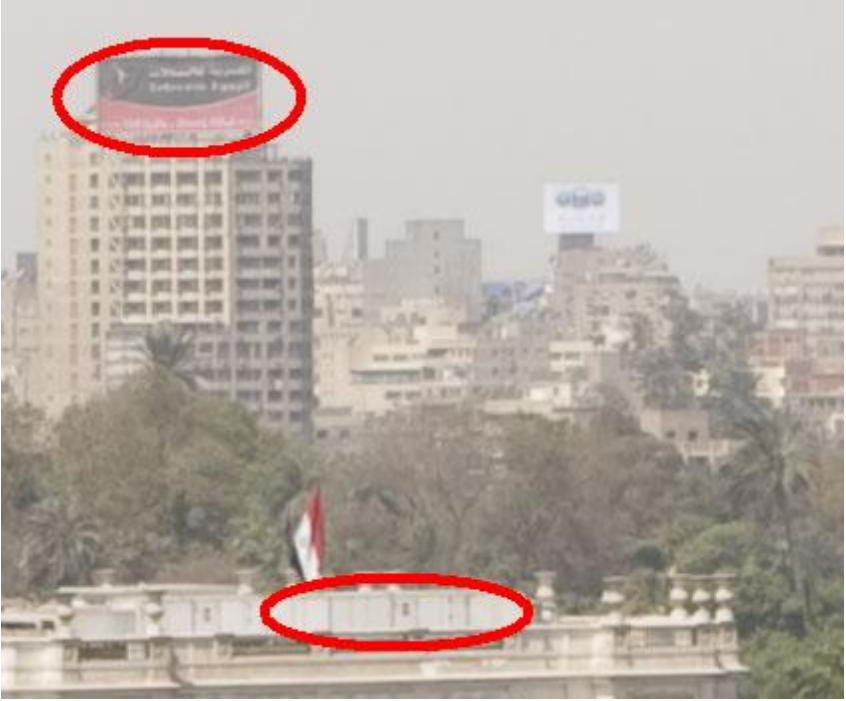

**Figure 49: This shows the two reference regions of the scene used to compute the ratio of the distances from Google Maps [40].**

The distance to the billboard in the scene above is approximately 2.22303 km and was measured as shown in the previous figure using Google Maps [40]. The distance to the flat surface of the building in the foreground is approximately 493.923 m, again measured by Google Maps [40]. The ratio between the distances of the two marked regions of the figure above is 4.5008 (= 2223.03 / 493.923). To compare the depthmap recovery of all the algorithms described in this research we used the ground-truth distance ratio as the benchmark. We took the average distance over patches of the image in the areas marked in **Figure 49** (the billboard and the foreground building).

| Algorithm | Distance Ratio |
| --- | --- |
| Ground truth (using Google Maps [40]) | 4.5008 |
| CO (scaled depthmap) | 0.1567 |
| CO-CDC (unscaled depthmap) | 0.2632 |
| POL (scaled depthmap) | 3.552 |
| POL-CDC (unscaled depthmap) | 3.6989 |
| DC (scaled depthmap) | 0.7102 |
| DC-CDC (unscaled depthmap) | 0.8318 |

**Figure 50: A comparison of the distance ratios computed by the algorithms. The closer the ratio is to the ground-truth, the more accurate the algorithm's depthmap.**

As you can see from the above table, the CDC optimization algorithm increases the accuracy of the depthmap by factoring out the atmospheric scattering scalar. For this reason all the algorithms analyzed are enhanced when applying the CDC to its output. The CDC optimization enhances the accuracy of the polarization dehazing algorithm by approximately 4% and enhances the accuracy of the dark channel by approximately 17%. From the perspective of depthmap accuracy the overall accuracy of the color optimization algorithm is not better than the polarization and dark channel dehazing algorithms. This can be seen by the recovered distance ratios being further away from the ground-truth ratio than the polarization and dark channel ratios. This could be due to unknown scales introduced in each method of dehazing. For example, the polarization method may introduce linear or non-linear scales in the process of filtering the light through the polarizer filter and using multiple polarized images of the same scene. Unknown scale factors in all of the methods can be due to ground-reflection, ambient light, sky aperture (the part of the sky which illuminates an object with its skylight) variation due to objects occluding other objects from parts of the sky. These many possible scale factors are too many to fully account for in the scope of this research. We are interested in analyzing the increase in efficiency our approach improves upon already existing methods. The polarization method has the most accurate depth estimation followed by the dark channel and then the color optimization method.

# 6.5 Conclusion

From the results described in the previous sections, we can see experimentally through the use of simulations and real live images two direct the contributions of this thesis: Color Optimization and Constant Depth Constraint (CDC). These two algorithms have provided valuable steps in the analyzing haze-filled scenes using simply the temporal factor of a sequence of images captured over time. The color optimization method does not rely on any of the limiting parameters of the previous dehazing methods to dehaze the scene and produce an accurate transmission map. The CDC optimization method quantifies the atmospheric scattering in a sequence of images of a certain scene captured over time. The goal of this is to pave the way to measuring particulate matter from digital images. We achieve this by measuring the atmospheric scattering coefficient which is strongly correlated to the levels of particulate matter (PM). In addition to this we recover an enhanced depthmap of the scene. This depthmap has been compared to the known dehazing algorithms and found to be more accurate than existing dehazing methods. When dealing with vast urban scenes a small increase in accuracy provides a large decrease in the margin of error of estimating the distances in the scene.

The polarization algorithm is one of the simplest methods of dehazing. It is an instant approach that allows the images to be taken at a point of time and the analysis done at the same time. Despite this it has a weakness in estimating atmospheric scattering. This is the instability it displays with errors in the degree of polarization. It also suffers from manual settings of the polarizer filter which needs to be automated in order to be practically usable. The dark channel and dichromatic dehazing methods show instability in other areas of the experiments, such as, image noise and errors in atmospheric airlight estimates. The dark channel shows some unstable results due to its prior which produces errors in the results and therefore prevents an accurate measurement of atmospheric scattering. The dichromatic method uses temporal knowledge but relies on too many external conditions, such as, changes in haze and overcast skies (as it does not handle changes in spectral illumination).

The color optimization presented in this research displays a promising contribution to measuring atmospheric scattering visually from images captured over time, a more accurate depthmap of the scene and dehazing haze-filled scenery. Proven by experimental verification, this thesis forms a promising framework to analyzing urban atmosphere much more accurately and therefore provides a means to measure PM levels visually.

# Chapter 7 – Contribution & Future Work

# 7.1 Contribution

This thesis has 4 main contributions:

1. **Measuring atmospheric scattering from images:** The Constant Depth Constraint optimization (CDC) takes a sequence of transmission maps produced over time by any of the dehazing algorithms in this research (or any other possible algorithm) and measures the atmospheric scattering for each time and a single depthmap of the scene. This is a novel contribution in the fact that it paves the way to visually measuring particulate matter. This is because the fluctuations in the atmospheric scattering can be correlated to levels of particulate matter.

2. **Measuring a more accurate depthmap from images taken over time:** The Color Optimization also recovers a more accurate depthmap of a scene. We compared this to existing dehazing algorithms. In addition to this the Constant Depth Constraint optimization enhances the accuracy in estimated depthmap of all the three dehazing algorithms described in this thesis. These algorithms are the polarization, temporal dichromatic and dark channel dehazing algorithms.

3. **Dehazing algorithm:** The Color Optimization algorithm recovers the scene radiance of the scene. This is a dehazed image of the scene and this method produces a dehazed image very similar to existing dehazing algorithms. In fact in addition to dehazing the scene, this algorithm also produces a depolluted image of the scene by removing the haze from the scene but leaving the natural blue sky. This makes the image appear like a clear-day scene instead of unnatural recovered images of the other dehazing algorithms which return unnatural black skies.

4. **A comprehensive comparison of dehazing algorithms from the perspective of measuring atmospheric scattering:** We compare three known dehazing algorithms with our algorithms for the purpose of measuring atmospheric scattering from digital images. This comparison, along with the other contributions of this thesis bring us one step closer to correlating this to particulate matter levels.

# 7.2 Future Work

- Correlating the atmospheric scattering quantity estimated by the methods described in this thesis to live PM measurements taken using conventional devices

- Developing a method of extracting only the haze due to PM and leaving the natural airlight in the produced images.

- Using the data gathered by this thesis to extract trends and patterns between image pixel values and transmittance in order to further dehaze the scene.

- Enhancing the optimization methods presented in this thesis by using possibly more efficient optimization methods

# Appendix: Matlab Scripts

# 1. Polarization-based Dehazing Script

```
%Polarization-based image dehazing algorithm
%input parameters: gr,gg,gb are the radiometric response curves for the camera
%output parameters: dehazedFinal,depthmap – dehazed scene and depthmap of the scene

function [dehazedFinal, depthmap] = matVTH(gr, gg, gb)

lgDeltaT = log(1/400); %natural logarithm of shutter speed

%Best and worst polarization images
imgBest = imread(<filename>);
imgWorst = imread(<filename>);

%Get dimensions of images
rmax = size(imgBest,1);
cmax = size(imgBest,2);

%convert to uint8
imgBestPolarizeduint8 = im2uint8(imgBest);
imgWorstPolarizeduint8 = im2uint8(imgWorst);

clear imgBest imgWorst;

%Get irradiance values from radiometric response curve of camera (according to Debevec and Malik's paper)
IrradianceBestR =  exp(gr(imgBestPolarizeduint8(:,:,1)+1) - lgDeltaT);
IrradianceBestG =  exp(gg(imgBestPolarizeduint8(:,:,2)+1) - lgDeltaT);
IrradianceBestB =  exp(gb(imgBestPolarizeduint8(:,:,3)+1) - lgDeltaT);
IrradianceWorstR =  exp(gr(imgWorstPolarizeduint8(:,:,1)+1) - lgDeltaT);
IrradianceWorstG =  exp(gg(imgWorstPolarizeduint8(:,:,2)+1) - lgDeltaT);
IrradianceWorstB =  exp(gb(imgWorstPolarizeduint8(:,:,3)+1) - lgDeltaT);

clear imgBestPolarizeduint8 imgWorstPolarizeduint8;

%sky patch coordinates (just above horizon)
startRow = 904;
endRow = 914;
startCol = 1083;
endCol = 1122;

%Cutout sky patch matrix from image (using coordinates above) and calculate degree of polarization (DOP)
DOP_R=(IrradianceWorstR(startRow:endRow,startCol:endCol)-IrradianceBestR(startRow:endRow,startCol:endCol))./
(IrradianceWorstR(startRow:endRow,startCol:endCol) + IrradianceBestR(startRow:endRow,startCol:endCol));
DOP_G=(IrradianceWorstG(startRow:endRow,startCol:endCol)-IrradianceBestG(startRow:endRow,startCol:endCol))./
(IrradianceWorstG(startRow:endRow,startCol:endCol) + IrradianceBestG(startRow:endRow,startCol:endCol));
DOP_B=(IrradianceWorstB(startRow:endRow,startCol:endCol)-IrradianceBestB(startRow:endRow,startCol:endCol))./
(IrradianceWorstB(startRow:endRow,startCol:endCol) + IrradianceBestB(startRow:endRow,startCol:endCol));

skyPatchArea = size(DOP_R, 1) * size(DOP_R, 2); % Sky patch area computation

%Sum up the total DOP and airlight at infinity over sky patch
sumPr = sum(sum(DOP_R));
sumPg = sum(sum(DOP_G));
sumPb = sum(sum(DOP_B));
AinfinityR = sum(sum(IrradianceWorstR)) + sum(sum(IrradianceBestR));
AinfinityG = sum(sum(IrradianceWorstG)) + sum(sum(IrradianceBestG));
AinfinityB = sum(sum(IrradianceWorstB)) + sum(sum(IrradianceBestB));

%Get average DOP and bias it up to prevent noise
Pr = ((sumPr) / skyPatchArea)*1.09
Pg = ((sumPg) / skyPatchArea)*1.09
```

```
Pb = ((sumPb) / skyPatchArea)*1.09

%Get average Airlight component at infinity
AinfinityR =  AinfinityR / skyPatchArea;
AinfinityG =  AinfinityG / skyPatchArea;
AinfinityB =  AinfinityB / skyPatchArea;

clear DOP_R DOP_G DOP_B skyPatchTotalIrrR skyPatchTotalIrrG skyPatchTotalIrrB sumPr sumPg sumPb imgBestPolarized imgWorstPolarized;
clear skyPatchArea;

%Pre-allocate matrices for operations
dehazed = zeros(rmax, cmax, 3, 'double');
transmittanceR = zeros(rmax, cmax, 1, 'double');
transmittanceG = zeros(rmax, cmax, 1, 'double');
transmittanceB = zeros(rmax, cmax, 1, 'double');
airlight = zeros(rmax, cmax, 3, 'double');

%A^ = (delta Irradiance / DOP)
airlight(:,:,1) = (IrradianceWorstR - IrradianceBestR) ./ Pr;
airlight(:,:,2) = (IrradianceWorstG - IrradianceBestG) ./ Pg;
airlight(:,:,3) = (IrradianceWorstB - IrradianceBestB) ./ Pb;

%remove airlight component from total irradiance
dehazed(:,:,1) = ((IrradianceWorstR + IrradianceBestR) - (airlight(:,:,1)));%Itotal - A;
dehazed(:,:,2) = ((IrradianceWorstG + IrradianceBestG) - (airlight(:,:,2)));%Itotal - A;
dehazed(:,:,3) = ((IrradianceWorstB + IrradianceBestB) - (airlight(:,:,3)));%Itotal - A;

clear IrradianceWorstR IrradianceWorstG IrradianceWorstB IrradianceBestR IrradianceBestG IrradianceBestB

airlightremoved = zeros(rmax, cmax, 3, 'uint8');
airlightremoved = dehazed;

%Transmittance = 1 - (airlight component / A at infinity)
%L (scene radiance) = (total irradiance - airlight) / T;
transmittanceR = 1 - (airlight(:,:,1) ./ AinfinityR);
dehazed(:,:,1) = dehazed(:,:,1) ./ transmittanceR;
transmittanceG = 1 - (airlight(:,:,2) ./ AinfinityG);
dehazed(:,:,2) = dehazed(:,:,2) ./ transmittanceG;
transmittanceB = 1 - (airlight(:,:,3) ./ AinfinityB);
dehazed(:,:,3) = dehazed(:,:,3) ./ transmittanceB;

depthmap = zeros(rmax, cmax, 1, 'double');
depthmap = double(((-log(transmittanceR)) + (-log(transmittanceG)) + (-log(transmittanceB))) ./ 3);

clear transmittanceR transmittanceG transmittanceB airlight

%Get the exposure value for the radiance value (dehazed). We do this to be able to look up the pixel brightness value in the radiometric response curve
%(the lookup is in the nested loop below) and obtain a pixel brightness value. Pixel value is Zij. g is the radiometric response curve.
%Zij = ln(irradiance) + ln(deltaT)     {according to Debevec and Malik}
dehazed = log(dehazed) + lgDeltaT;
dehazedFinal = zeros(rmax, cmax, 3, 'uint8');

for i=1:rmax
    for j=1:cmax
        %lookup g(Zij) (exposure value) in the radiometric response function to obtain the pixel
        %brightness value (0-255). This will form the image as a uint8 RGB matrix
        dehazedFinal(i,j,1) = getIndex(gr, (dehazed(i,j,1)));
        dehazedFinal(i,j,2) = getIndex(gg, (dehazed(i,j,2)));
        dehazedFinal(i,j,3) = getIndex(gb, (dehazed(i,j,3)));
    end
end
```

# 2. Temporal Dehazing Script

```
%Dichromatic framework: temporal dehazing algorithm
%input parameters: gr,gg,gb are the radiometric response curves for the camera
%output parameters: dehazedFinal,DOTz – dehazed scene and the scaled depth map (scaled by the DOT – difference in optical thickness)

function [dehazedFinal,DOTz] = dichrom(gr,gg,gb)

lgDeltaT = log(1/400); %natural logarithm of shutter speed

%read the two images taken under two different haze conditions
img1 = imread('d:\scene1.tif');
img2 = imread('d:\scene2.tif');

%get image dimensions
w = size(img1,1);
h = size(img1,2);

%get image irradiance for uint8 image pixel brightness values
E1r = exp(gr(img1(:,:,1)+1) - lgDeltaT);
E1g = exp(gg(img1(:,:,2)+1) - lgDeltaT);
E1b = exp(gb(img1(:,:,3)+1) - lgDeltaT);

E2r = exp(gr(img2(:,:,1)+1) - lgDeltaT);
E2g = exp(gg(img2(:,:,2)+1) - lgDeltaT);
E2b = exp(gb(img2(:,:,3)+1) - lgDeltaT);

E1(:,:,1)=E1r;
E1(:,:,2)=E1g;
E1(:,:,3)=E1b;

E2(:,:,1)=E2r;
E2(:,:,2)=E2g;
E2(:,:,3)=E2b;
%E1 and E2 represent image irradiance

%create dichromatic planes (Ni = E1 x E2)
Ni = cross(E1,E2);

%get matrix values: sum of (Nr^2,Ng^2,Nb^2,NrNg,NrNb,NgNb) over i (multiplied by 2)
M = zeros(3, 3, 'double');
M(1,1)=2*sum(sum((Ni(:,:,1).^2)));
M(2,2)=2*sum(sum((Ni(:,:,2).^2)));
M(3,3)=2*sum(sum((Ni(:,:,3).^2)));
M(1,2)=2*sum(sum((Ni(:,:,1).*Ni(:,:,2))));
M(2,1)=M(1,2);
M(1,3)=2*sum(sum((Ni(:,:,1).*Ni(:,:,3))));
M(3,1)=M(1,3);
M(3,2)=2*sum(sum((Ni(:,:,2).*Ni(:,:,3))));
M(2,3)=M(3,2);

%use SVD or Eigen decomposition to solve for the airlight unit vector (airlight color direction)
[U,S,V]=svd(M);
A=V(:,end);

%airlight unit vector in terms of image irradiance (derived from E1 and E2)
A = [A(1),A(2),A(3)]

% find tA - solve ((E2-tA)xE1)^2=0
for i=1:w
    for j=1:h
        e1=[E1(i,j,1) E1(i,j,2) E1(i,j,3)];
```

```matlab
        e2=[E2(i,j,1) E2(i,j,2) E2(i,j,3)];
        e1n = e1 ./ norm(e1);
        e2n = e2 ./ norm(e2);

        AxE1  = cross(A,e1);
        E2xE1 = cross(e2,e1);

        %point on unit vector A that makes E1 // E2
        dot(AxE1, E2xE1);
        dot(AxE1, AxE1);
        %calculate t
        t = dot(AxE1, E2xE1) / dot(AxE1, AxE1);
        %tA vector along A unit vector
        tA = t* A;
        %p2/p1 - direct transmission ratio
        dt_ratio(i,j) = norm(e2 - tA) / norm(e1);
        %||tA||
        mag_tA(i,j) = norm(tA);
    end
end

x = dt_ratio; y = mag_tA; p = polyfit(x,y,1);
E_infty_1 = -p(1)%slope, because equation is y=c-mx, so slope must be negated
E_infty_2 = p(2)%intersect

%scaled depth map (B2-B1)z - DOT*z
DOTz = zeros(w, h, 'double');
DOTz = log(E_infty_2/E_infty_1)-log(dt_ratio);
top = max(max(DOTz));
bottom = min(min(DOTz));
fig=figure, imshow(DOTz, [bottom top]);
saveas(fig,'d:\after.jpg');

%define cube of dimensions 10 for finding q estimate and hence minimum
%alpha value corresponding to scene point dominated by airlight
cubenorms = zeros(6, 4, 'double');
cubenorms(1,:)=[100,0,0,0];
cubenorms(2,:)=[0,100,0,0];
cubenorms(3,:)=[0,0,100,0];
cubenorms(4,:)=[-100,0,0,1000];
cubenorms(5,:)=[0,-100,0,1000];
cubenorms(6,:)=[0,0,-100,1000];

negA = -A; %negative A unit vector (direction of negative A)
q_estimate = zeros(w, h, 'double');

%check intersection with any of the cubes faces
for i=1:w
    for j=1:h
        E = [E1(i,j,1),E1(i,j,2),E1(i,j,3)]; %scene 1 point color
        %check intersection with cube
        for m=1:6
            normalofcubeface = cubenorms(m,1:3);
            dp2 = dot(normalofcubeface,negA);
            if(dp2<0)
                dp = dot(normalofcubeface,E);
                D = cubenorms(m,4);
                q_estimate(i,j)=-(dp + D) / dp2;% this is the distance along vector negA from point E to cube face
            end
        end
    end
end

Bd_estimate = -log(1-(q_estimate ./ E_infty_1));
```

```
alpha = (Bd_estimate ./ DOTz);

% get k least alpha values
 [val, col] = min(min(alpha));
[val, row] = min(alpha(:,col));
[rows,cols] = find(alpha==val);

%use k least alphas to dehazed whole scene
k=size(rows,1)

dehazed_double = zeros(w, h, 3, 'double');

%for each scene point with minimum alpha, compute real q for that point and compute dehazed color for all scene points relative to this take the average dehazed
%color of all the k alphas
for index=1:k
    e1 = [E1(rows(index),cols(index),1) E1(rows(index),cols(index),2) E1(rows(index),cols(index),3)];
    realQ = norm(e1);%magnitude of scene point color (E=qA : q is real q)
    Bz1 = -log(1-(realQ / E_infty_1));
    %get relative q for all other scene points based on realQ of scene point u,v
    rel_depth_ratio = DOTz ./ DOTz(rows(index),cols(index));
    Bz = Bz1 .* rel_depth_ratio;
    q = E_infty_1 .* (1 - exp(-Bz));
    p = (E_infty_1 .* exp(-Bz))./ (Bd_estimate^2);

    dehazed_double(:,:,1) = dehazed_double(:,:,1) + (E1(:,:,1) - (q * A(1)));
    dehazed_double(:,:,2) = dehazed_double(:,:,2) + (E1(:,:,2) - (q * A(2)));
    dehazed_double(:,:,3) = dehazed_double(:,:,3) + (E1(:,:,3) - (q * A(3)));
end

dehazed_double = dehazed_double ./ k;%get average color over all k

%convert irradiance back to color
dehazed_double = log(dehazed_double) + lgDeltaT;
dehazedFinal = zeros(w, h, 3, 'uint8');
for m=1:w
  for n=1:h
    dehazedFinal(m,n,1) = getIndex(gr, (dehazed_double(m,n,1)));
    dehazedFinal(m,n,2) = getIndex(gg, (dehazed_double(m,n,2)));
    dehazedFinal(m,n,3) = getIndex(gb, (dehazed_double(m,n,3)));
  end
end

imwrite(dehazedFinal, 'd:\dehazed.tif');
```

# 3. Dark Channel Dehazing script

```
function [Bd,tm,Dehazed,D,Best] = DC ()
time=6;
lgDeltaT = log(1/400);

%reading image files
I(:,:,:,1) = imread('D:\reg\fully registered\Ireg1_3 _tinydichrom.tif');
I(:,:,:,2) = imread('D:\reg\fully registered\Ireg2_3 _tinydichrom.tif');
I(:,:,:,3) = imread('D:\reg\fully registered\Ireg3_3 _tinydichrom.tif');
I(:,:,:,4) = imread('D:\reg\fully registered\Ireg4_3 _tinydichrom.tif');
I(:,:,:,5) = imread('D:\reg\fully registered\Ireg5_3 _tinydichrom.tif');
I(:,:,:,6) = imread('D:\reg\fully registered\Ireg6_3 _tinydichrom.tif');

I = im2double(I);

%patch size must be a factor of width and height
```

```
Imin = zeros(size(I,1), size(I,2), 3);
Inorm = zeros(size(I,1), size(I,2), 3);
Ac = zeros(time,1);
Jdark = zeros(size(I,1), size(I,2), 5);
tm = zeros(size(Jdark));
for t=1:time
   %get minimum in each color channel over patches
   Imin(:,:,1) = minfilt2(I(:,:,1,t), [13 9]);
   Imin(:,:,2) = minfilt2(I(:,:,2,t), [13 9]);
   Imin(:,:,3) = minfilt2(I(:,:,3,t), [13 9]);

   %get dark channel
   Jdarkchannel = min(min(Imin(:,:,1), Imin(:,:,2)), Imin(:,:,3));
   Jdark_t = Jdarkchannel(:,:);
   %get highest intensity pixel as atmospheric light
   %find maximum Jdark pixels and get x,y of the maximum I of these pixels
   [r,c] = find(Jdark_t==max((Jdark_t(:))));
   maxy=1;
   for u=1:size(r,1)
      for v=1:size(c,1)
         if I(r(u),c(v),1) > maxy
            maxy = I(r(u),c(v),1);
         end
      end
   end
   Ac(t) = maxy;
   %normalize by atmospheric light
   Inorm(:,:,:) = I(:,:,:,t) / Ac(t);

%get dark channel of image normalized to the atmospheric airlight

   Imin(:,:,1) = minfilt2(Inorm(:,:,1), [13 9]);
   Imin(:,:,2) = minfilt2(Inorm(:,:,2), [13 9]);
   Imin(:,:,3) = minfilt2(Inorm(:,:,3), [13 9]);

   Jdark(:,:,t) = min(min(Imin(:,:,1), Imin(:,:,2)), Imin(:,:,3));
end
clear Imin Inorm Jdark_t

%transmission map
tm = 1 - Jdark;
%scaled depthmap
Bd = -log(tm);
```

# 4. Simulated Haze Scenes

```
%Synthesize two scenes with different haze conditions
%input parameters: gr,gg,gb are the radiometric response curves for the camera
%output parameters: E1,E2 – two scenes with different haze conditions

function [E1, E2] = synthesize2(gr,gg,gb)

lgDeltaT = log(1/400); %natural logarithm of shutter speed
scene = imread('d:\synthetic_scene.tif');

w=size(scene,1);
h=size(scene,2);
```

```
radiance=zeros(w,h,3,'double');

%get image irradiance of colors
radiance(:,:,1) =  exp(gr(scene(:,:,1)+1) - lgDeltaT);
radiance(:,:,2) =  exp(gg(scene(:,:,2)+1) - lgDeltaT);
radiance(:,:,3) =  exp(gb(scene(:,:,3)+1) - lgDeltaT);

E1=zeros(w,h,3,'double');
E2=zeros(w,h,3,'double');

%image irradiance at infinity
E_infty_1=200;
E_infty_2=400;%white color at infinity

B1=1.0;
B2=1.5;
B1/B2;

%16 color patches (defining coordinates for each)
coordinate = zeros(16, 4);
coordinate(1,:)=[1,50,1,50];
coordinate(2,:)=[1,50,51,100];
coordinate(3,:)=[1,50,101,150];
coordinate(4,:)=[1,50,151,200];

coordinate(5,:)=[51,100,1,50];
coordinate(6,:)=[51,100,51,100];
coordinate(7,:)=[51,100,101,150];
coordinate(8,:)=[51,100,151,200];

coordinate(9,:)=[101,150,1,50];
coordinate(10,:)=[101,150,51,100];
coordinate(11,:)=[101,150,101,150];
coordinate(12,:)=[101,150,151,200];

coordinate(13,:)=[151,200,1,50];
coordinate(14,:)=[151,200,51,100];
coordinate(15,:)=[151,200,101,150];
coordinate(16,:)=[151,200,151,200];

%relative depths: z1, z2, ... z16
zmin=3;
zmax=4.5;
z = zeros(16, 1, 'double');
for n=1:16
    z(n)=(zmin + (zmax-zmin)*rand(1,1));
end

%airlight unit vector
r=20;
g=30;
b=50;
A=[r g b];
Airr(1) =  exp(gr(A(1)+1) - lgDeltaT);
Airr(2) =  exp(gg(A(2)+1) - lgDeltaT);
Airr(3) =  exp(gb(A(3)+1) - lgDeltaT);
A = Airr ./ norm(Airr)

%attenuation due to scattering over distance
for i=1:16
    p1=(E_infty_1*(exp(-B1*z(i)))) / (z(i)^2);
    p2=(E_infty_2*(exp(-B2*z(i)))) / (z(i)^2);

    E1(coordinate(i,1):coordinate(i,2),coordinate(i,3):coordinate(i,4),1)=radiance(coordinate(i,1):coordinate(i,2),coordinate(i,3):coordinate(i,4),1).*(p1);
```

```
  E1(coordinate(i,1):coordinate(i,2),coordinate(i,3):coordinate(i,4),2)=radiance(coordinate(i,1):coordinate(i,2),coordinate(i,3):coordinate(i,4),2).*(p1);
  E1(coordinate(i,1):coordinate(i,2),coordinate(i,3):coordinate(i,4),3)=radiance(coordinate(i,1):coordinate(i,2),coordinate(i,3):coordinate(i,4),3).*(p1);

  E2(coordinate(i,1):coordinate(i,2),coordinate(i,3):coordinate(i,4),1)=radiance(coordinate(i,1):coordinate(i,2),coordinate(i,3):coordinate(i,4),1).*(p2);
  E2(coordinate(i,1):coordinate(i,2),coordinate(i,3):coordinate(i,4),2)=radiance(coordinate(i,1):coordinate(i,2),coordinate(i,3):coordinate(i,4),2).*(p2);
  E2(coordinate(i,1):coordinate(i,2),coordinate(i,3):coordinate(i,4),3)=radiance(coordinate(i,1):coordinate(i,2),coordinate(i,3):coordinate(i,4),3).*(p2);
end

%add airlight
for j=1:16
  q1=E_infty_1*(1-(exp(-B1*z(j))));
  q2=E_infty_2*(1-(exp(-B2*z(j))));

  E1(coordinate(j,1):coordinate(j,2),coordinate(j,3):coordinate(j,4),1)=E1(coordinate(j,1):coordinate(j,2),coordinate(j,3):coordinate(j,4),1)+(q1*A(1));
  E1(coordinate(j,1):coordinate(j,2),coordinate(j,3):coordinate(j,4),2)=E1(coordinate(j,1):coordinate(j,2),coordinate(j,3):coordinate(j,4),2)+(q1*A(2));
  E1(coordinate(j,1):coordinate(j,2),coordinate(j,3):coordinate(j,4),3)=E1(coordinate(j,1):coordinate(j,2),coordinate(j,3):coordinate(j,4),3)+(q1*A(3));

  E2(coordinate(j,1):coordinate(j,2),coordinate(j,3):coordinate(j,4),1)=E2(coordinate(j,1):coordinate(j,2),coordinate(j,3):coordinate(j,4),1)+(q2*A(1));
  E2(coordinate(j,1):coordinate(j,2),coordinate(j,3):coordinate(j,4),2)=E2(coordinate(j,1):coordinate(j,2),coordinate(j,3):coordinate(j,4),2)+(q2*A(2));
  E2(coordinate(j,1):coordinate(j,2),coordinate(j,3):coordinate(j,4),3)=E2(coordinate(j,1):coordinate(j,2),coordinate(j,3):coordinate(j,4),3)+(q2*A(3));
end

E11 = log(E1) + lgDeltaT;
E22 = log(E2) + lgDeltaT;

scene1 = zeros(w, h, 3, 'uint8');
scene2 = zeros(w, h, 3, 'uint8');

%convert image irradiance back to RGB brightness value
for m=1:w
  for n=1:h
    scene1(m,n,1) = getIndex(gr, (E11(m,n,1)));
    scene1(m,n,2) = getIndex(gg, (E11(m,n,2)));
    scene1(m,n,3) = getIndex(gb, (E11(m,n,3)));

    scene2(m,n,1) = getIndex(gr, (E22(m,n,1)));
    scene2(m,n,2) = getIndex(gg, (E22(m,n,2)));
    scene2(m,n,3) = getIndex(gb, (E22(m,n,3)));
  end
end

imwrite(scene1,'d:\scene1.tif');
imwrite(scene2,'d:\scene2.tif');
toc
```

# 5. Image Registration

```
base = imread(<path>);
unregistered = imread(<path>)
tform = cp2tform(cpstruct, 'projective');
registered = imtransform(unregistered, tform);
imwrite(base, 'c:\base.tif');
imwrite(registered, 'c:\registered.tif');
```

# 6. Color Optimization

```
function [tm, Bd, Bscaled, realtmap, tmap] = CO(noiseVar)
```

```
w=40;
h=40;
patchsz = 10;
coord = getCoordinates(w,h,patchsz);
Ai = ones(5,1);%Ai's are measured using sky region or some technique (the best method)
J = rand(w,h,3);%scene radiance
Z = [20 10 9.5 9 8 7 6 5 4 3 2 1.7 1.3 1.2 1.1 1];%0.1 + (4-0.1).*rand(size(coord,1),1);
B = [0.1 0.15 0.2 0.25 0.3];

for i=1:size(coord,1)
    T(i,:) = exp(-B(:)*Z(i));
End

I = zeros(size(J,1), size(J,2), 3, size(T,2));
Test = zeros(size(T,1), size(T,2));

%create patches
for t=1:size(T,2)
    for d=1:size(coord,1)
        noise = noiseVar*randn(patchsz,patchsz,3);
        I(coord(d,1):coord(d,2),coord(d,3):coord(d,4),:,t) = (J(coord(d,1):coord(d,2),coord(d,3):coord(d,4),:) * T(d,t)) + (Ai(t) * (1-T(d,t))) + noise;
    end
    imwrite(I(:,:,:,t), strcat('d:\color optimization simulations\I', int2str(t), '.tif'));
end
display('done computing Is');

colsum = sum(sum(sum((I(:,:,:,:))))));
[n,m]=min(colsum);
indexToClampT = m
[n,m]=max(colsum);
indexToClampB = m;
Bscaled = B / B(indexToClampB);

% Ai(i)=1;%max(max(max((I(:,:,:,i))))));

for d=1:size(coord,1)
    
    %initial values
    Jx = zeros(patchsz,patchsz,3);%+0.0001;
    Ti = zeros(size(T,2),1);%+0.0001;
    %Ti(indexToClampT) = 1;
    deltaTi = 100;
    stopper = 1;
    
    %patch parameters
    Ti_num = zeros(patchsz,patchsz,3,size(T,2));
    Ti_denom = zeros(patchsz,patchsz,3,size(T,2));
    Jx_num_r = zeros(patchsz,patchsz,size(T,2));
    Jx_num_g = zeros(patchsz,patchsz,size(T,2));
    Jx_num_b = zeros(patchsz,patchsz,size(T,2));
    Jx_denom = zeros(patchsz,patchsz,size(T,2));

tic

%iterate
while deltaTi > 1e-5 && stopper < 500
    
    for u=1:size(Jx,1)
        for v=1:size(Jx,2)
            if Jx(u,v,1) < 0
                Jx(u,v,1) = 0;
            end
            if Jx(u,v,2) < 0
                Jx(u,v,2) = 0;
```

```matlab
        end
        if Jx(u,v,3) < 0
            Jx(u,v,3) = 0;
        end
        if Jx(u,v,1) > 1
            Jx(u,v,1) = 0;
        end
        if Jx(u,v,2) > 1
            Jx(u,v,2) = 0;
        end
        if Jx(u,v,3) > 1
            Jx(u,v,3) = 0;
        end
    end
  end

%compute Ti
for c=1:3
    for i=1:size(T,2)
        Ti_num(:,:,c,i) = (I(coord(d,1):coord(d,2),coord(d,3):coord(d,4),c,i) - Ai(i)).*(Ai(i) - Jx(:,:,c));
    end
    Ti_denom(:,:,c) = (Ai(i) - Jx(:,:,c)).^2;
end

  Ti_numerator(1:size(T,2)) = -sum(sum(sum(Ti_num(:,:,:,1:size(T,2)))));
  Ti_denominator = sum(sum(sum(Ti_denom(:,:,:))));
  TiOld = Ti;
  %TiOld(indexToClampT) = 1;
  Ti(1:size(T,2)) = Ti_numerator(1:size(T,2)) / Ti_denominator;
  %Ti(indexToClampT) = 1;

  %compute maximum delta of Ti's
  diff = abs(TiOld - Ti);
  deltaTi = max(diff);

  stopper = stopper + 1;

   %clamp positive
  for u=1:size(Ti,1)
    Ti(u) = min(1,Ti(u));
    Ti(u) = max(0,Ti(u));
  End

  %compute Jx
  for t=1:size(T,2)
    Jx_num_r(:,:,t) = (Ti(t) * I(coord(d,1):coord(d,2),coord(d,3):coord(d,4),1,t)) - (Ai(t)*(1-Ti(t))*Ti(t));
    Jx_num_g(:,:,t) = (Ti(t) * I(coord(d,1):coord(d,2),coord(d,3):coord(d,4),2,t)) - (Ai(t)*(1-Ti(t))*Ti(t));
    Jx_num_b(:,:,t) = (Ti(t) * I(coord(d,1):coord(d,2),coord(d,3):coord(d,4),3,t)) - (Ai(t)*(1-Ti(t))*Ti(t));
    Jx_denom(:,:,t) = Ti(t)^2;
  end

  Jx_numerator_r = Jx_num_r(:,:,1)+Jx_num_r(:,:,2)+Jx_num_r(:,:,3)+Jx_num_r(:,:,4)+Jx_num_r(:,:,5);
  Jx_numerator_g = Jx_num_g(:,:,1)+Jx_num_g(:,:,2)+Jx_num_g(:,:,3)+Jx_num_g(:,:,4)+Jx_num_g(:,:,5);
  Jx_numerator_b = Jx_num_b(:,:,1)+Jx_num_b(:,:,2)+Jx_num_b(:,:,3)+Jx_num_b(:,:,4)+Jx_num_b(:,:,5);
  Jx_denomenator = Jx_denom(:,:,1)+Jx_denom(:,:,2)+Jx_denom(:,:,3)+Jx_denom(:,:,4)+Jx_denom(:,:,5);
  Jx(:,:,1) = Jx_numerator_r ./ Jx_denomenator;
  Jx(:,:,2) = Jx_numerator_g ./ Jx_denomenator;
  Jx(:,:,3) = Jx_numerator_b ./ Jx_denomenator;

end
Test(d,:)=Ti(:);
end

for d=1:size(Test,1)
```

```
    for t=1:size(T,2)
        realtmap(coord(d,1):coord(d,2),coord(d,3):coord(d,4),t) = T(d,t);
        tmap(coord(d,1):coord(d,2),coord(d,3):coord(d,4),t) = Test(d,t);
    end
end

for t=1:5
    [n,m]=find(tmap(:,:,t)==0)
    for i=1:size(n)
        for j=1:size(m)
            if tmap(n(i),m(j),t) == 0
                tmap(n(i),m(j),t)=1e-20;
            end
        end
    end
end

for d=1:size(coord,1)
    realZ(coord(d,1):coord(d,2),coord(d,3):coord(d,4)) = Z(d);
end
realZ = realZ(1:40,1:40);

tm = tmap;
Bd = -log(tm);
```

# 7. Constant Depth Constraint Optimization

```
function [D, B] = optBd_polarizationAlt(T, indexToClamp)

Tti = zeros(size(T,1),size(T,2),size(T,3));
Tti(:,:,:) = T(:,:,1:size(T,3));
Tti(:,:,:) = -log(Tti(:,:,:));

%initial values
B = zeros(size(T,3),1)+0.0001;
D = zeros(size(T,1), size(T,2)) + 0.0001;
B(indexToClamp) = 1;
deltaB = 100;
stopper = 1;

%iterate - compute depth (D), compute Betas (B)
while deltaB > 1e-5 && stopper < 500

    %compute B(t)
    for t=1:size(T,3)
        B_num(:,:,t) = Tti(:,:,t).*D(:,:);
        B_denom(:,:,t) = D(:,:).^2;
    end

    B_numerator(1:size(T,3)) = sum(sum(B_num(:,:,1:size(T,3))));
    B_denominator(1:size(T,3)) = sum(sum(B_denom(:,:,1:size(T,3))));
    BOld = [B(1:size(T,3))];
    BOld(indexToClamp) = 1;
    B(1:size(T,3)) = B_numerator(1:size(T,3)) ./ B_denominator(1:size(T,3));
    B(indexToClamp) = 1;

    %compute maximum delta of B's
    diff = abs(BOld - B);
    deltaB = max(diff)

    stopper = stopper + 1;
```

```
%keep B positive and between 0 and 1
for u=1:size(B,1)
   B(u)=min(1,B(u));
   B(u)=max(0,B(u));
end

%compute D(i)
sumDnum = zeros(size(T,1), size(T,2));
sumDdenom = 0;
for t=1:size(T,3)
   sumDnum = sumDnum + Tti(:,:,t)*B(t);
   sumDdenom = sumDdenom + B(t)^2;
end

D = sumDnum / sumDdenom;

  %keep D positive
for u=1:size(D,1)
   for v=1:size(D,2)
       D(u,v) = max(0,D(u,v));
   end
end

end

deltatBeta = deltaB
```

# 8. Coordinates of Depth Patches (getCoordinates)

```
function [coord] = getCoordinates(w,h,ps)

index=1;
for i=1:(w/ps)
   startrow = ps*(i-1) + 1;
   endrow   = ps*i;

   for j=1:(h/ps)
      startcol = ps*(j-1) + 1;
      endcol   = ps*j;

      coord(index,:) = [startrow endrow startcol endcol];
      index = index + 1;
   end
end
```